\documentclass[10pt]{article}

\usepackage{tgpagella}      
\usepackage{mathpazo}       
\usepackage{inconsolata}    

\usepackage[letterpaper, margin=0.75in, top=0.6in, bottom=0.7in]{geometry}

\usepackage[T1]{fontenc}
\usepackage{microtype}
\usepackage{parskip}        
\usepackage{graphicx}
\usepackage{subcaption}
\usepackage[dvipsnames,table]{xcolor}
\usepackage{array}
\usepackage{hyperref}
\usepackage{url}
\usepackage{natbib}
\setcitestyle{authoryear,round,citesep={;},aysep={,},yysep={;}}
\usepackage{caption}
\usepackage{enumitem}
\setlist{leftmargin=*}

\usepackage{booktabs}
\usepackage{longtable}
\usepackage{amsmath,amssymb}
\usepackage{amsthm}
\theoremstyle{remark}
\newtheorem{remark}{Remark}
\usepackage{autonum}          
\usepackage{enumitem}
\usepackage{float}
\usepackage{wrapfig}
\usepackage{placeins}        
\usepackage[most]{tcolorbox}
\usepackage{multirow}
\usepackage{pifont}
\usepackage{colortbl}
\usepackage{titletoc}        
\usepackage[capitalise,noabbrev]{cleveref}  
\usepackage{algorithm}
\usepackage{algpseudocode}

\definecolor{bpBase03}{HTML}{002B36}
\definecolor{bpBase01}{HTML}{586E75}
\definecolor{bpBase1}{HTML}{93A1A1}
\definecolor{bpBase2}{HTML}{EEE8D5}
\definecolor{bpBase3}{HTML}{FDF6E3}
\definecolor{bpBlueBg}{HTML}{E7F4FF}
\definecolor{bpGreenBg}{HTML}{F3F8E8}
\definecolor{bpBlue}{HTML}{268BD2}
\definecolor{bpSky}{HTML}{4FA3D9}
\definecolor{bpCyan}{HTML}{2AA198}
\definecolor{bpMagenta}{HTML}{D33682}
\definecolor{bpRose}{HTML}{F06AA6}
\definecolor{bpViolet}{HTML}{6C71C4}
\definecolor{bpGreen}{HTML}{859900}
\definecolor{bpYellow}{HTML}{B58900}
\definecolor{bpOrange}{HTML}{CB4B16}
\definecolor{bpRed}{HTML}{DC322F}
\hypersetup{
  colorlinks = true,
  linkcolor  = bpBlue,
  citecolor  = bpBlue,
  urlcolor   = bpBlue,
}

\newcommand{\cmark}{\ding{51}}
\newcommand{\xmark}{\ding{55}}

\newcommand{\benchpress}{\textsc{BenchPress}}

\newtcolorbox{titlebox}[1][]{
  colback=bpBlueBg, colframe=bpBlue,
  fonttitle=\bfseries\small, sharp corners=south, boxrule=1pt,
  left=6pt, right=6pt, top=5pt, bottom=5pt, fontupper=\small,
  #1,
}
\newcounter{finding}
\crefname{finding}{Finding}{Findings}
\Crefname{finding}{Finding}{Findings}

\NewDocumentEnvironment{finding}{m m +b}{%
  \refstepcounter{finding}%
  \begin{tcolorbox}[
    colback=bpMagenta!5, colframe=bpMagenta,
    boxrule=0.5pt, arc=3pt,
    left=6pt, right=6pt, top=4pt, bottom=4pt,
    fontupper=\small,
  ]\textbf{Finding~\thefinding:}\label{find:#2} #1 #3\end{tcolorbox}%
}{}

\newcommand{\cBP}{\cellcolor{bpMagenta!6}} 

\usepackage{amsmath,amsfonts,bm}

\def\eqref#1{(\ref{#1})}

\def\1{\bm{1}}

\DeclareMathAlphabet{\mathsfit}{\encodingdefault}{\sfdefault}{m}{sl}
\SetMathAlphabet{\mathsfit}{bold}{\encodingdefault}{\sfdefault}{bx}{n}

\graphicspath{{figures/}}

\usepackage{fontawesome5}

\begin{document}

\vspace*{-0.4in}
\begin{center}
{\LARGE\bf You Don't Need to Run Every Eval}\\[6pt]
{\normalsize Yuchen Zeng \& Dimitris Papailiopoulos}\\[2pt]
{\normalsize\bf Microsoft Research, AI Frontiers}\\[4pt]
{\small \faGithub\;\textbf{Code:} \href{https://github.com/microsoft/benchpress}{\texttt{https://github.com/microsoft/benchpress}}}\\[2pt]
{\small \faExternalLink*\;\textbf{Project page:} \href{https://microsoft.github.io/benchpress/}{\texttt{https://microsoft.github.io/benchpress/}}}\\[2pt]
{\small \raisebox{-0.15em}{\includegraphics[height=0.9em]{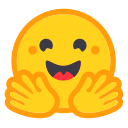}}\;\textbf{Dataset:} \href{https://huggingface.co/datasets/microsoft/benchpress-score-matrix}{\texttt{https://huggingface.co/datasets/microsoft/benchpress-score-matrix}}}
\end{center}
\vspace{-0.5cm}
\noindent\rule{\textwidth}{0.4pt}
\vspace{-0.2cm}

\renewcommand{\thefootnote}{}
\footnotetext{\small Emails: \texttt{\{zengyuchen, dimitriosp\}@microsoft.com}}
\renewcommand{\thefootnote}{\arabic{footnote}}

\renewcommand{\abstractname}{}    
\vspace{-0.5cm}
\begin{abstract}
 \noindent
A modern model release reports scores on 40+ benchmarks and the same evaluations were run many more times before it: to track training progress, compare design choices, and select the checkpoint for the release. 
But do we need to run every eval? 
We compile a public score matrix of 84 frontier models on 133 benchmarks (2{,}604 cells, 23.3\% filled) and find it is approximately rank-2: a model's scores across all benchmarks are largely determined by just two numbers. 
We confirm this in two ways: (i) scores hidden from the matrix are best recovered using two factors, and (ii) two factors already explain over 90\% of the variation among models on the benchmarks they share. 
Building on this, we design \textsc{BenchPress}: a logit-space rank-2 matrix completion method that recovers held-out scores to within $4.6$ points. 
Using \textsc{BenchPress}, we find a subset of five benchmarks $\{$GPQA-D, HLE, Codeforces, MMLU-Pro, ARC-AGI-1$\}$ that can recover the rest of a model's public scorecard to within $3.93$ points. 
For a tighter evaluation budget, a cheaper set $\{$GPQA-D, MMLU-Pro, Aider Polyglot, MATH-500, AIME 2026$\}$ can predict a model's evals to within $4.55$. 
Finally, we identify what affects prediction reliability and combine these factors with predictor disagreement to quantify when predictions can be trusted.
We release the score matrix, the code for reproducing all of our experiments on Github.
\end{abstract}

\vspace{-0.1in}

\vspace{0.7cm}
\begin{figure}[H]
\centering
\begin{subfigure}[t]{0.48\linewidth}
    \centering
    \vspace{0pt}
    \includegraphics[width=\linewidth]{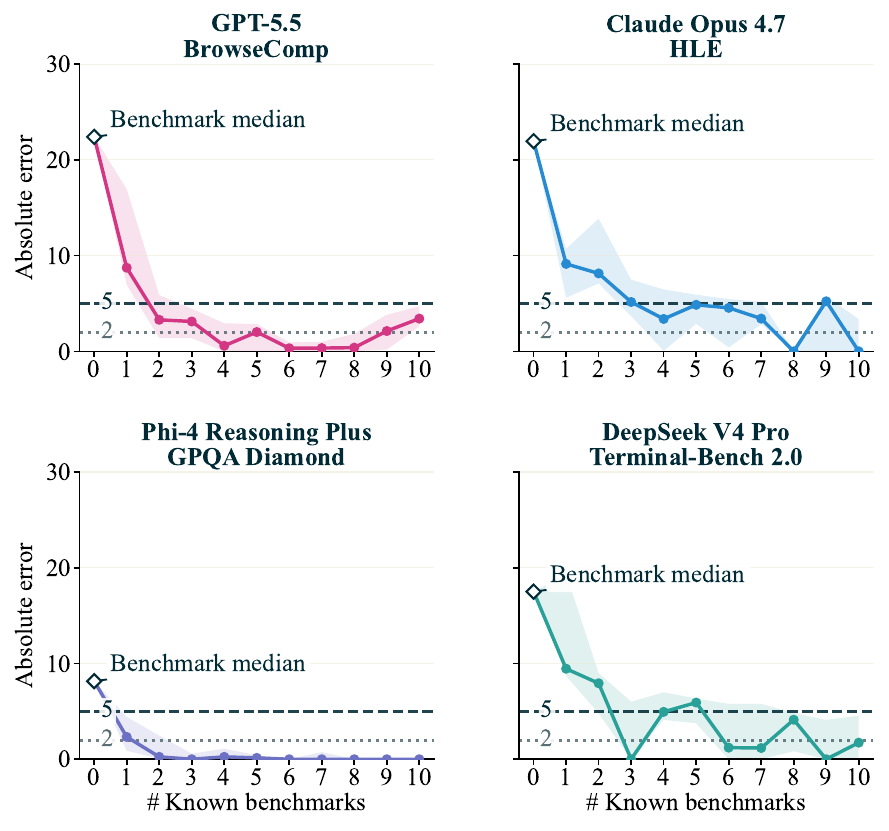}
    \caption{Accuracy of predicting one (model, benchmark) score.}
    \label{fig:hero_examples}
\end{subfigure}%
\hfill
\begin{subfigure}[t]{0.51\linewidth}
    \centering
    \vspace{0pt}
    \includegraphics[width=\linewidth]{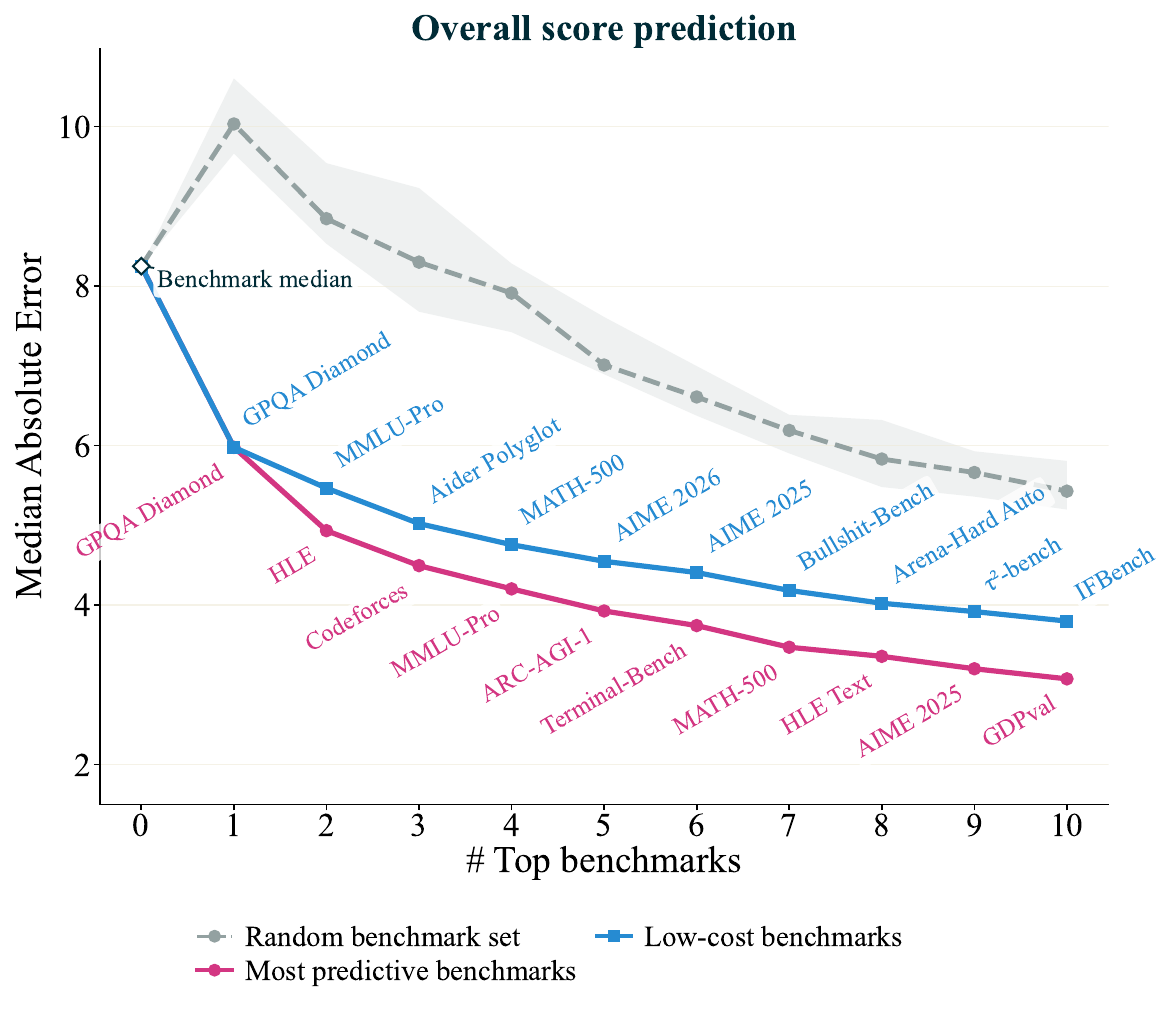}
    \caption{Reporting overall score-prediction error.}
    \label{fig:hero_overall}
\end{subfigure}
\caption{\textbf{\benchpress{} predicts unseen benchmark scores from known benchmark scores.}
\textbf{Left:} For each of four model--benchmark examples, we randomly reveal $k$ scores from the same model and hide the rest.
The y-axis shows the absolute error for the highlighted target score.
Error drops sharply after only a few scores from the same model are revealed.
\textbf{Right:} A complementary setting that mimics how a practitioner would run \benchpress{} in practice. A fixed set of $k$ benchmarks is chosen as the probe set, and every model is evaluated on whichever probe scores it has observed; \benchpress{} predicts the rest of each model's scores, and we report pooled error across all evaluated cells. With only five benchmark probes selected on the current matrix, pooled MedAE drops to $3.93$ score points ($4.55$ when restricted to a lower inference cost list; see \Cref{sec:probe_selection}).
See \Cref{app:hero_figure} for the detailed experiment setting.}
\label{fig:hero}
\end{figure}

\newpage
\begingroup
\setlength{\parskip}{0.35em}
\setcounter{tocdepth}{2}
\tableofcontents
\setcounter{tocdepth}{2}
\endgroup
\newpage

\newpage
\section{Introduction}
\label{sec:intro}

LLM evaluation is expensive, and its cost is continuing to rise.
A frontier model release now routinely reports scores on dozens of benchmarks: Qwen3.5 reports 40 language benchmark rows \citep{qwen35}, and Kimi K2.5 reports 43 benchmark rows \citep{kimik25}.
But the scores reported at release are only a small fraction of the evaluations run during model development.
During model development, researchers repeatedly run the same benchmarks to compare checkpoints and model variants trained under different settings.
After release, users run additional evaluations to compare models for their own applications. These evaluations can cost thousands of dollars and take several days.
This raises a question: 
\begin{center}
    {\it Do we always need to run every evaluation, or can we use scores we already have to predict those we have not measured?}
\end{center}

Benchmark scores are clearly not independent measurements. 
Strong performance on coding and agentic benchmarks often co-occurs with strong performance on competition-math benchmarks: for example, SWE-bench Verified~\citep{jimenez2024swebench, openai2024swebenchverified} is strongly correlated with AIME~\citep{aime} and MATH-500~\citep{lightman2023math500}, and Terminal-Bench variants~\citep{terminalbench} show similar but noisier trends.
Prior work has found low-dimensional structure in benchmark scores~\citep{burnell2023structure,ilic2024,ruan2024,burnham2025}.
Some studies have also used fixed sets of evaluations to predict broader performance~\citep{aitchison2023atari5,ruan2024}.
\citet{aitchison2023atari5} select five games that approximate an agent's median performance across all 57 Atari games.
\citet{ruan2024} derive capability measures from a fixed, densely observed set of standard benchmarks and fit a separate predictor for each selected downstream target.
We ask a more general question: for a new model, can any subset of its known benchmark scores predict all remaining scores, without fixing the input benchmarks or prediction targets in advance?
If a few observed scores can predict the rest of a model's benchmark profile to useful accuracy, practitioners have a new option for evals: run a small set of probes and infer the rest, instead of running every evaluation independently.
To answer this question, we make the contributions summarized below. 

\begin{enumerate}[nosep,leftmargin=*]
  \item \textbf{We compile a public score matrix and show that it is effectively rank-2.}
  We note that broad evaluation collections such as HELM and Every Eval Ever already aggregate scores across many models and benchmarks~\citep{liang2023helm,batzner2026everyeval}, while a separate line of work has documented low-dimensional structure in model--benchmark score matrices~\citep{burnell2023structure,ilic2024,ruan2024,burnham2025}.
  Our contribution connects these two lines in the general prediction setting: given any new model and any subset of its observed benchmark scores, can we predict its remaining scores without further assumptions?
  We collect scores from public sources, canonicalize near-duplicate model variants and benchmark configurations, and filter out models and benchmarks with insufficient observations to obtain an $84 \times 133$ matrix with $2{,}604$ observed entries (23.3\% of all model--benchmark cells).
  Two independent diagnostics on this matrix show that it is effectively rank-2: rank-sweeping Soft-Impute matrix completion minimizes held-out prediction error at rank 2, and SVDs of the largest fully-observed submatrices show that two factors explain more than 90\% of variance (\Cref{sec:predictability}).
  \item \textbf{We build \benchpress{}, a benchmark score predictor.}
  We evaluate seven feature transforms and twelve prediction methods, finding that the best full-coverage score predictor is a rank-2 alternating least squares (ALS) matrix-completion method in logit space~\citep{koren2009}. 
  It predicts every missing model--benchmark cell, reaching $4.6$ score-point median absolute error on held-out entries at $100\%$ coverage (\Cref{sec:bp_methods}).
  \item \textbf{We show what \benchpress{} enables for model evaluation.}
  (i) We \emph{select compact probe sets} that recover a model's scorecard under an evaluation budget: even when restricted to a low-cost benchmark allowlist, five probes lead to pooled MedAE of $4.55$ score points (\Cref{sec:probe_selection}).
  (ii) We \emph{verify ranking preservation}: allowing a five-point margin on the true scores, completed scores by \benchpress{} preserve $92.1\%$ of pairwise model rankings on the same benchmark (\Cref{sec:ranking_preservation}).
  (iii) We \emph{stress-test predictions on newly released models}: across 27 target models, each predicted from a training matrix that predates its release, five seed scores lead to median absolute error of $4.83$ points (\Cref{sec:temporal_deployment}).
  \item \textbf{We characterize when predictions should be trusted.}
  We first identify what makes a score easier or harder to predict: target-model and target-benchmark coverage, the availability of similar peer models and neighboring benchmarks, and recency of training anchors. 
    We combine these factors with disagreement among several score predictors to estimate the reliability of each \benchpress{} prediction.
    This produces a trust probability and a conformally calibrated $90\%$ prediction interval (\Cref{sec:trust}).
\end{enumerate}

\Cref{fig:hero} previews \benchpress{} on four individual model--benchmark examples and across all models in the score matrix. 
In both cases, as more benchmark scores are observed, \benchpress{} quickly reduces prediction error.

\vspace{-0.3cm}
\paragraph{Scope and caveats.}
Our claims should be read within three limits.
\emph{(i) Public-score heterogeneity:} the matrix mixes vendor-reported and third-party scores that may differ in reasoning effort, sampling temperature, prompting, judge model, evaluation harness, and benchmark version.
\benchpress{} therefore predicts what this public matrix would extrapolate to rather than what a controlled re-evaluation under a single configuration would yield.
\emph{(ii) Snapshot dependence:} the main experiments use the fixed May 2026 score matrix.
We continue to update the released matrix, and \benchpress{} also supports score matrices from other sources; \Cref{app:other_matrices} reports results on the August 2026 update, HELM~\citep{liang2023helm}, and Every Eval Ever~\citep{batzner2026everyeval}.
\emph{(iii) Score inferability:} our analysis identifies benchmark \emph{scores} that are currently inferable from others, not benchmarks whose \emph{existence} is unnecessary.
Running a benchmark can still show where a model fails, whether it may have seen the test data before, and how it performs on new kinds of data. Benchmark results also influence what model developers choose to improve.
\section{Related Work}
\label{sec:related}


\paragraph{Low-rank structure in evaluation.}
A recurring finding is that benchmark reporting is highly redundant~\citep{burnell2023}: scores are governed by a few latent capability factors that recur across very different model and benchmark collections.
On 27 HELM~\citep{liang2023helm} tasks over 29 models, \citet{burnell2023structure} recover three factors (reasoning, comprehension, core language modeling); scaling to 591 Open LLM Leaderboard models across 12 benchmarks, \citet{ilic2024} find a single dominant $g$-factor explaining 85\% of the variance; and on the Epoch AI Capabilities Index, \citet{burnham2025} reduce scores to a general-capability axis plus a provider-specific residual via PCA.
This redundancy invites prediction: \citet{ruan2024} build a low-dimensional capability space from standard benchmarks and use it to estimate selected downstream capabilities.
\benchpress{} builds on the same premise but works from a heterogeneous, frontier-era public matrix, where we likewise recover an effectively rank-2 geometry (\Cref{sec:bp_svd}).

\paragraph{Benchmark compression and design.}
\citet{perlitz2024} showed HELM evaluation can be compressed $100\times$ with minimal ranking reliability loss; their follow-up \citep{perlitz2024bat} formalized best practices for evaluating whether benchmarks agree with one another.
\citet{ni2024mixeval} constructed MixEval, a single compact benchmark from web-query-matched items that achieves 0.96 Chatbot Arena correlation.
These approaches select or design a fixed evaluation suite \emph{a priori}.
\benchpress{} instead predicts missing scores from whatever benchmarks happen to be available, requiring no fixed probe set: a practitioner can feed in MMLU \citep{hendrycks2021mmlu} and GPQA \citep{rein2024gpqa} today, or LiveCodeBench and AIME tomorrow, without reconfiguration.

\paragraph{Item-level subset selection.}
A complementary line of work reduces cost \emph{within} individual benchmarks by selecting which test items to run.
\emph{IRT-based} methods include MetaBench \citep{kipnis2024}, which uses item response theory to keep 3\% of items across six benchmarks while preserving aggregate conclusions, and tinyBenchmarks \citep{polo2024}, which builds on Anchor Points using IRT to pick informative items.
\emph{Correlation- or embedding-based} methods include Anchor Points \citep{vivek2024anchorpoints}, which selects items via cross-model correlations; Scales++ \citep{bean2025scalespp}, which uses cognitive-scale embeddings to reduce cost $18\times$ at 2.9\% MAE without prior model evaluations; DISCO \citep{rubinstein2025}, which condenses sets by selecting items where models disagree most; SubLIME \citep{saranathan2025sublime}, which trains a correlation predictor for compact subsets; EssenceBench \citep{wang2026essencebench}, which applies genetic algorithms for up to $200\times$ compression; and \citet{zhou2025}, who exploit low-rank structure at the example level for up to $20\times$ speedups.
Most of these methods require instance-level pass/fail data across many models to calibrate item selection; Scales++ is a notable exception.
\benchpress{} requires only aggregate scores and predicts \emph{across} benchmarks, a complementary approach that could be combined with item-level methods for end-to-end savings.

\paragraph{Score prediction.}
Closest to our work are methods that predict aggregate benchmark scores directly.
The nearest methodological predecessor is \citet{schram2023}, who applied Bayesian matrix factorisation to cross-lingual NLP performance, though in a different domain (languages $\times$ tasks, not LLMs $\times$ benchmarks); on LLM scores specifically, \citet{zhang2024cpp} applied collaborative filtering, \citet{polo2024sloth} used latent skill models for cross-benchmark prediction, and \citet{ye2023} showed BIG-bench is 95\%+ predictable.
A second group predicts performance from a fixed panel of densely observed evaluations: \citet{aitchison2023atari5} select five games from the fully observed Arcade Learning Environment to estimate median performance across the 57-game suite, and \citet{ruan2024} construct capability measures from a fixed block of standard benchmarks to fit target-specific predictors.
Both assume a pre-specified input panel observed across historical models or agents; we compare directly against the latter in \Cref{app:osl_comparison}, while Atari-5 does not transfer because such a panel is essentially never fully observed in our matrix.
Others predict from outside the score matrix entirely: \citet{park2025precog} use LLMs to predict scores from text descriptions alone, with no execution (we revisit this comparison empirically in \Cref{sec:additional_llm_baseline}), and \citet{koh2026rbridge} train a small proxy model to transfer reasoning performance to larger models, whereas \benchpress{} needs no model access at all.
\benchpress{} instead performs general score prediction on an uncontrolled public matrix, where any subset of a new model's scores may be observed and the rest are targets.
We differ from these methods in three ways: (1)~we operate at substantially larger scale and on a frontier-era snapshot ($84$ models $\times$ $133$ benchmarks, including post-2024 reasoning, coding, and agentic suites), (2)~we compare 84 transform--method configurations head-to-head on the same data, and (3)~we provide explicit failure analysis, showing where and why prediction breaks.

\section{The Score Matrix and Its Geometry}
\label{sec:predictability}

In this section we ask: given a collection of existing LLM benchmark scores, can we predict the missing ones from a small subset?
Our starting point is a \emph{score matrix} with models on one axis and benchmarks on the other, populated from publicly available evaluations.
\Cref{sec:bp_collection} describes how each cell is sourced and audited.
\Cref{sec:bp_data} introduces the resulting matrix and discusses its data quality limitations.
\Cref{sec:bp_svd} reveals that the score matrix is effectively rank-2.
Appendix details for data collection and the released score matrix appear in \Cref{app:data_collection,app:benchmark_score_matrix}.
Throughout, $M$ denotes the number of models, $B$ denotes the number of benchmarks, $s_{mb}$ denotes the observed score of model $m$ on benchmark $b$, and $\hat{s}_{mb}$ denotes its prediction.

\subsection{Data Collection}
\label{sec:bp_collection}

Our data collection proceeds in four steps.
\emph{First}, we seed a queue with a small initial set of models (GPT-5.5~\citep{openai2026gpt55}, Claude Opus 4.7~\citep{anthropic2026claude47}, Gemini 3.1 Pro~\citep{google2026gemini31}, DeepSeek-V4-Pro~\citep{deepseek2026v4pro}, and a few other widely-discussed recent releases) and crawl every official source attached to each: the release blog, the system card, the technical report, and the Hugging Face model card.
\emph{Second}, we recurse: each source typically reports the model's own scores together with a handful of competitor baselines, so any newly mentioned model is added to the queue and crawled in turn, until no further round introduces an unvisited model.
\emph{Third}, we sweep a fixed list of primary leaderboards (MathArena~\citep{matharena}, ARC-Prize~\citep{arcprize}, Terminal-Bench~\citep{terminalbench}, LMArena~\citep{lmarena}, Epoch AI~\citep{epochai_frontiermath}, LiveBench~\citep{white2024livebench}) to fill remaining gaps for models and benchmarks that vendors do not directly cover.
\emph{Fourth}, we filter the resulting raw matrix to a dense subset that supports the analyses in the rest of the paper.

When the same model-benchmark cell is reported by multiple sources we resolve conflicts as best we can: we keep the highest-priority value, with priority order release blog, then system card, then technical report, then HuggingFace model card, then primary leaderboards, breaking ties by recency.
We also fix one canonical configuration per model (typically the setting the vendor itself foregrounds, such as a specific reasoning effort level) to avoid inflating coverage with effectively duplicate rows from near-duplicate variants.
Every retained value carries the URL it was sourced from (\Cref{app:data_collection} shows the released record format), and roughly half come from a source outside the model's own provider.
Alongside the score itself, each entry records the model's release date, provider, and canonical evaluation setting (mode, reasoning effort, sampling, judge, harness, prompt style, temperature, tool use), and each benchmark records its category, metric, problem count, and reference link, so downstream analyses can condition on release timing, evaluation regime, or benchmark type without re-crawling the sources.

\begin{figure}[!htbp]
\centering
\newsavebox{\bpfigbox}%
\savebox{\bpfigbox}{%
  \begin{minipage}{0.56\textwidth}
    \centering
    \includegraphics[width=\textwidth]{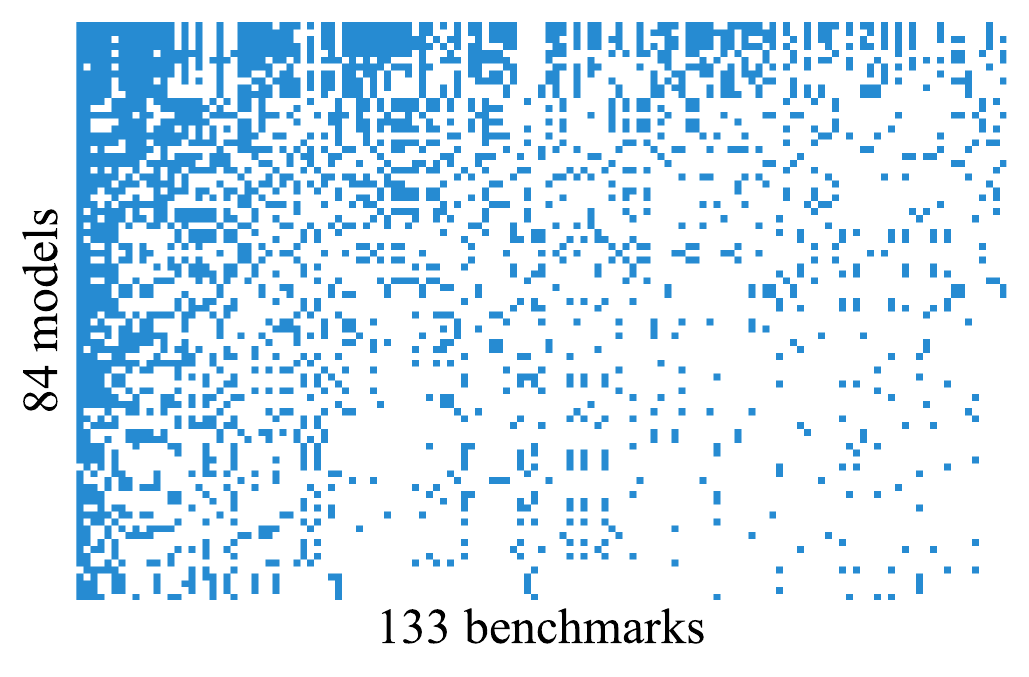}
    \captionof{figure}{Observation pattern of the $84 \times 133$ score matrix, sorted by coverage. Each row is a model and each column is a benchmark; dark cells are observed scores. Only 23.3\% of entries are filled.}
    \label{fig:bp_matrix}
  \end{minipage}%
}%
\begin{minipage}[c][\dimexpr\ht\bpfigbox+\dp\bpfigbox\relax]{0.42\textwidth}
    \centering
    \captionof{table}{\textbf{The threshold filter trades coverage for density.} Each row applies a minimum-observation requirement to model rows and benchmark columns, iterated to a fixed point. We use the bolded setting throughout.}
    \label{tab:bp_threshold_sweep}
    \vspace{0.3em}
    \small
    \setlength{\tabcolsep}{3pt}
    \resizebox{\textwidth}{!}{%
    \begin{tabular}{cc cccr}
    \toprule
    \multicolumn{2}{c}{Min.\ obs.\ per} & \multicolumn{4}{c}{Resulting matrix} \\
    \cmidrule(lr){1-2}\cmidrule(lr){3-6}
    Model & Bench. & \#Models & \#Bench. & \#Obs. & Fill \\
    \midrule
    \multicolumn{2}{c}{Raw audit pool} & 188 & 316 & 4{,}493 & 7.6\% \\
    \cmidrule(lr){1-6}
    10 & 8  & 130 & 141 & 3{,}201 & 17.5\% \\
    10 & 12 & 124 & 104 & 2{,}795 & 21.7\% \\
    10 & 16 & 112 &  70 & 2{,}246 & 28.6\% \\
    \cmidrule(lr){1-6}
    \cBP\textbf{15} & \cBP\textbf{8}  & \cBP\textbf{84} & \cBP\textbf{133} & \cBP\textbf{2{,}604} & \cBP\textbf{23.3\%} \\
    15 & 12 &  61 &  81 & 1{,}811 & 36.7\% \\
    15 & 16 &  53 &  53 & 1{,}337 & 47.6\% \\
    \cmidrule(lr){1-6}
    20 & 8  &  49 & 110 & 1{,}885 & 35.0\% \\
    20 & 12 &  41 &  69 & 1{,}353 & 47.8\% \\
    \bottomrule
    \end{tabular}}
    \vfill
\end{minipage}%
\hfill
\usebox{\bpfigbox}%
\end{figure}

However, the raw score matrix suffer from the following problems. 
\begin{itemize}[nosep,leftmargin=*]
  \item \textbf{Evaluation heterogeneity.}
    Scores come from vendor reports and independent third parties and may differ in prompting strategy, reasoning budget, sampling temperature, evaluation harness, benchmark version, and evaluation date.
    Vendor-reported scores may be optimistically biased relative to independent reproductions.
    Even with identical prompts, non-deterministic decoding can change scores by 1 to 3 points across runs, while different harnesses can shift the same benchmark score by more than 5 points~\citep{biderman2024}.

  \item \textbf{Structured missingness.}
    Popular models $\times$ popular benchmarks are over-represented, violating the uniform-sampling assumption underlying standard matrix-completion guarantees.
\end{itemize}

We allow these effects to remain for two reasons.
First, our goal is to predict roughly where a model lands on a benchmark, not to reproduce an exact score.
This is enough for a practitioner deciding whether to run the evaluation.
Second, the public record is heterogeneous by nature; if we required a single controlled setting, too few scores would remain to work with.
Reducing these effects still improves accuracy, so we apply filtering procedure below keeps one canonical setting per model and one canonical column per benchmark family, then removes rows and columns with too few observations.

\paragraph{Filtering.}
The raw matrix at this point (May, 2026) contains 188 models and 316 benchmarks but only 4{,}493 of the 59{,}408 cells are filled (7.6\%), with the long tail dominated by barely-observed rows and columns.
This raw audit pool is useful for provenance, but it is not yet the analysis matrix: some rows or columns are alternate views of the same underlying signal.
We first canonicalize these cases.
For model setting variants, we keep one representative row rather than making one mode trivially predictable from another.
For benchmark variants, we keep one canonical column per task family; same-scale versions may fill missing canonical cells as non-canonical measurements, while different-scale variants are excluded.
This yields a canonicalized pool with 181 models, 304 benchmarks, and 4{,}177 observed cells.
The canonicalized pool is still very sparse, so we then filter to a dense subset by requiring every retained model to be observed on at least a minimum number of benchmarks, and every retained benchmark to be observed on at least a minimum number of models.
\Cref{tab:bp_threshold_sweep} shows how the number of retained entries and the fill rate change across candidate thresholds.
But there is no single definition of the best threshold: a looser threshold keeps more data but leaves the matrix sparser, while a stricter threshold creates a denser matrix by discarding more data.
We therefore use 15 observations per model and 8 per benchmark as a practical balance for all analyses in this paper, but other choices are equally reasonable.
Prediction results remain similar across the full grid, with several stricter thresholds giving lower error (\Cref{app:data_collection}).
The resulting matrix has 84 models $\times$ 133 benchmarks with 2{,}604 observed cells (23.3\% fill).

\begin{remark}
Large collections of public benchmark scores already exist, for example the HELM leaderboard \citep{liang2023helm} and Every Eval Ever (EEE)~\citep{batzner2026everyeval}, and \benchpress{} is not tied to our particular matrix: it works on independent matrices rebuilt from these sources as well (\Cref{app:other_matrices}). We nonetheless run the main study on our own curated matrix, because the structural questions we ask require control over how each score was produced. We keep one canonical setting per model and one canonical column per benchmark family, rather than letting the same model or benchmark recur under different settings. Otherwise near-duplicate rows or columns would make the prediction task trivially easy, since one variant could be read off almost directly from another, and the low-rank geometry would reflect measurement redundancy rather than shared capability.
\end{remark}

\subsection{The Final Score Matrix}
\label{sec:bp_data}

\begin{figure*}[!t]
\centering
\begin{subfigure}[t]{0.49\linewidth}
    \centering
    \includegraphics[width=\linewidth]{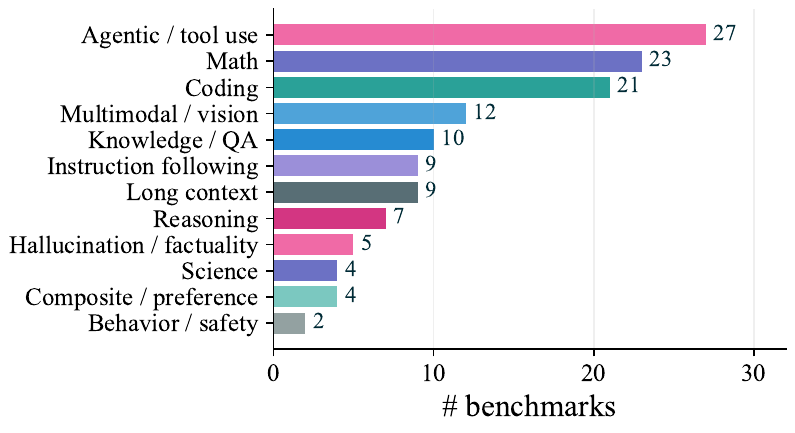}
    \caption{Benchmark mix by category.}
    \label{fig:bp_bench_mix}
\end{subfigure}\hfill
\begin{subfigure}[t]{0.49\linewidth}
    \centering
    \includegraphics[width=\linewidth]{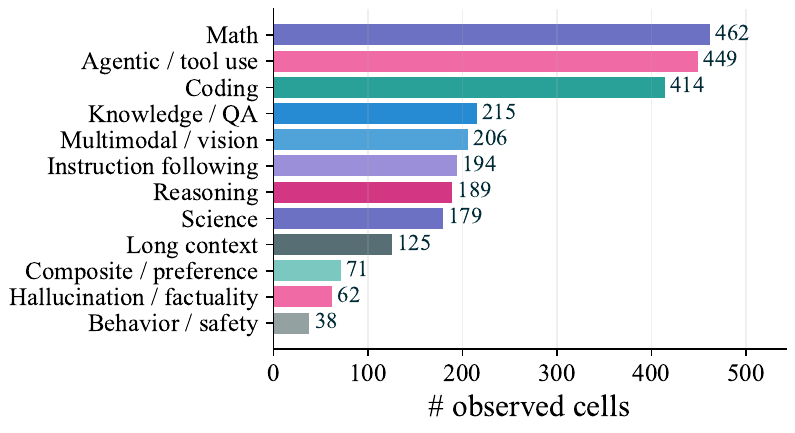}
    \caption{Where observed scores concentrate.}
    \label{fig:bp_obs_concentrate}
\end{subfigure}

\vspace{0.6em}

\begin{subfigure}[t]{0.32\linewidth}
    \centering
    \includegraphics[width=\linewidth]{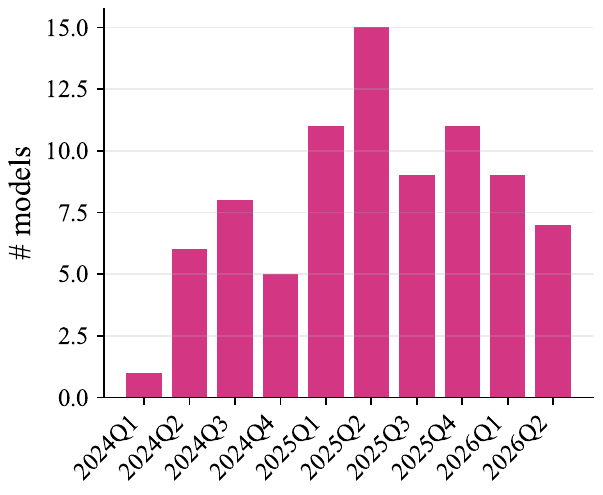}
    \caption{Model releases over time.}
    \label{fig:bp_releases}
\end{subfigure}\hfill
\begin{subfigure}[t]{0.32\linewidth}
    \centering
    \includegraphics[width=\linewidth]{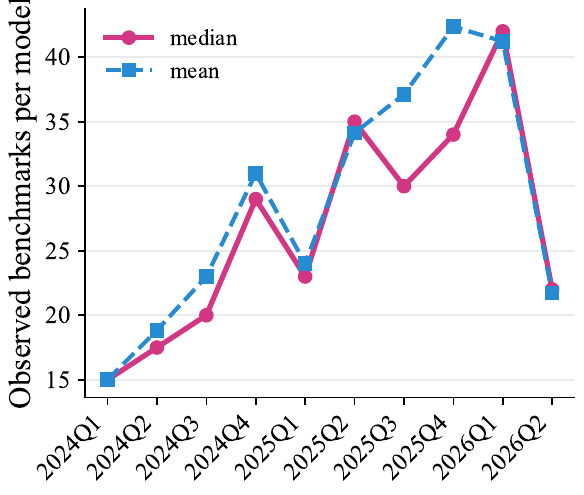}
    \caption{Coverage by model release time.}
    \label{fig:bp_coverage}
\end{subfigure}\hfill
\begin{subfigure}[t]{0.32\linewidth}
    \centering
    \includegraphics[width=\linewidth]{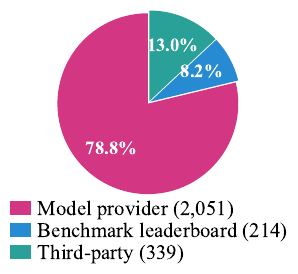}
    \caption{Source provenance of observed cells.}
    \label{fig:source_provenance}
\end{subfigure}
\caption{\textbf{Composition and coverage of the adopted score matrix} (84 models $\times$ 133 benchmarks, 2{,}604 observed cells, 23.3\% fill). The matrix spans a broad mix of benchmark categories (a), with most observed cells in math, coding, agentic/tool-use, and knowledge-oriented evaluations (b). Models are concentrated in recent releases (c), and coverage varies by release time because newer models are often reported on different benchmark suites than older baselines (d). Roughly four in five scores come from the model provider's own materials, with the remainder split between benchmark leaderboards and third-party aggregators (e).}
\label{fig:data_distribution}
\end{figure*}

Throughout, $s_{mb}$ denotes the observed score of model $m$ on benchmark $b$ and $\hat{s}_{mb}$ its prediction.
After the curation pipeline of \Cref{sec:bp_collection}, the score matrix contains 2{,}604 observed entries out of 11{,}172 cells (23.3\% fill rate).
\Cref{fig:bp_matrix} shows the sparsity pattern: popular models and benchmarks are well-covered, but the lower-right corner is almost entirely empty.
\Cref{fig:data_distribution} summarizes what this adopted matrix contains: a broad benchmark mix, with observed cells concentrated in math, coding, agentic/tool-use, and knowledge-oriented evaluations; a model set concentrated in recent releases, with coverage varying by release time; and score provenance dominated by model-provider materials, with the remainder split between benchmark leaderboards and third-party aggregators.

\paragraph{Benchmarks.}
The 133 benchmarks span all major LLM evaluation axes: agentic tasks and tool use, math, coding, multimodal and vision, long context, instruction following, knowledge and QA, reasoning, hallucination and factuality, science, composite indices, human preference, safety, and other specialized categories.
The full benchmark inventory with metrics, item counts, and source links is provided in \Cref{tab:benchmarks}.

\paragraph{Models.}
The 84 models span 13 providers: OpenAI (20), Google (12), Anthropic (11), Alibaba/Qwen (11), DeepSeek (9), Meta (6), Zhipu AI (4), Moonshot AI (3), xAI (3), MiniMax (2), Cohere (1), ByteDance (1), and Mistral (1).
Among models with annotated type, 51 are reasoning models (chain-of-thought) and 31 are non-reasoning.
Among models with annotated release status, 35 are open-weight and 47 are closed.
Where parameter counts are disclosed, they range from 1B (e.g., Gemma 3 1B~\citep{google2025gemma3}) to 1.6T (e.g., DeepSeek-V4-Pro \citep{deepseek2026v4pro}).

Full benchmark and model details are provided in \Cref{app:data} (\Cref{tab:benchmarks,tab:models}).

\paragraph{Research snapshot and practical use.}
We answer the paper's research questions using the fixed May 2026 snapshot, which after filtering contains 84 models and 133 benchmarks (\Cref{fig:bp_matrix}); the unfiltered audit pool covers 188 models across 316 benchmarks.
All reported experiments remain tied to this snapshot for reproducibility.
For practical use, we have since updated the curated matrix to an August 2026 snapshot and extended \benchpress{} to support HELM~\citep{liang2023helm}, Every Eval Ever (EEE)~\citep{batzner2026everyeval}, and user-provided score matrices.
\Cref{app:other_matrices} confirms that the same predictor remains effective on independently built HELM and EEE matrices.

\subsection{Rank-2 Geometry}
\label{sec:bp_svd}

\begin{figure*}[!t]
\centering
\begin{minipage}[t]{0.44\textwidth}
    \centering
    \vspace{0pt}
    \includegraphics[width=0.9\linewidth]{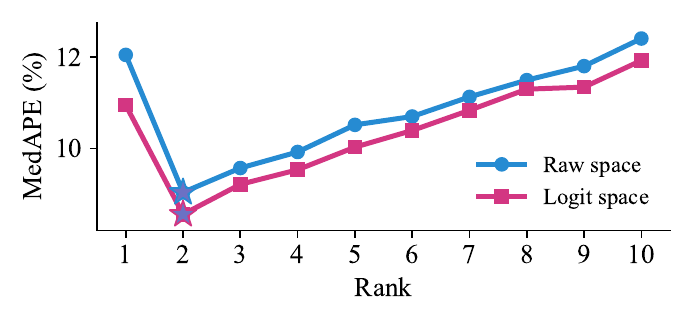}
    \vspace{-.2in}
    \captionof{figure}{Held-out MedAPE vs.\ rank for raw- and logit-space Soft-Impute matrix completion. Both curves bottom out at rank~2.}
    \label{fig:rank_ucurve}
\end{minipage}%
\hfill
\begin{minipage}[t]{0.53\textwidth}
    \centering
    \vspace{0pt}
    \captionof{table}{Complete submatrix SVD analysis across different trade-off points between benchmark coverage and model count. ``Var (top $k$)'' denotes the cumulative variance explained by the first $k$ singular components.}
    \label{tab:submatrix}
    \vspace{-.05in}
    \small
    \setlength{\tabcolsep}{2.5pt}
    \resizebox{\linewidth}{!}{%
    \begin{tabular}{ccccc}
    \toprule
    \textbf{\# Bench.} & \textbf{\# Models} & \cBP\textbf{Stable rank} & \textbf{Var (top 1)} & \textbf{Var (top 2)} \\
    \midrule
4 & 42 & \cBP 1.11 & 90.4\% & 97.6\% \\
7 & 11 & \cBP 1.28 & 78.2\% & 91.2\% \\
10 & 7 & \cBP 1.16 & 86.5\% & 95.0\% \\
13 & 6 & \cBP 1.12 & 89.6\% & 95.9\% \\
    \bottomrule
    \end{tabular}
    }
\end{minipage}%
\end{figure*}

\paragraph{Evidence for effective rank-2.}
If model capabilities lie in a low-dimensional space, the score matrix should be predictable from a low-rank completion.
The operational question is therefore not only whether observed scores can be compressed, but which rank best predicts held-out benchmark scores.
We establish rank 2 through two lines of evidence:
\emph{(i)}~rank-2 matrix completion minimizes held-out prediction error in raw-score and logit-score spaces (\Cref{fig:rank_ucurve}), and
\emph{(ii)}~\emph{Singular Value Decomposition} (SVD) of fully-observed submatrices shows matching rank-2 geometry.

\emph{Evidence 1: Rank-2 completion minimizes held-out prediction error.}
We use Soft-Impute \citep{mazumder2010}, a standard matrix-completion method that alternates between filling missing entries and taking a low-rank SVD approximation.
We sweep its rank in raw-score and logit-transformed score spaces; the latter linearizes percentage scores before standardization.
We evaluate on held-out entries using Median Absolute Percentage Error (MedAPE), the median of absolute percentage errors $|\mathrm{predicted}-\mathrm{true}|/|\mathrm{true}|\times100\%$.
\Cref{fig:rank_ucurve} shows the same pattern in both score spaces: held-out error is minimized at rank~2 and rises for higher ranks.

We also test rank selection in a harder setting that separates newer models from the older models used to learn the shared structure.
We remove a group of newer models from the training matrix and predict their remaining scores using the older models and the scores observed for each target.
Rank 2 remains the best-performing choice in logit space, consistent with the main rank sweep (\Cref{app:rank_geometry}).

\emph{Evidence 2: Complete submatrices show matching rank-2 geometry.}
Recall that the best rank-$r$ approximation retains the top $r$ singular values, and the \emph{stable rank} $\|M\|_F^2 / \sigma_1^2$ measures effective dimensionality: values near 1 mean that one component dominates.
We mean-center each benchmark column before computing SVD, so that the leading component reflects directions of model variation rather than the shared average score level (equivalent to PCA on the submatrix).
We then take the largest fully-observed model subsets available at several benchmark-coverage levels and compute the SVD of each submatrix.
\Cref{tab:submatrix} shows the same pattern across these different shapes: the spectra are dominated by a single direction, and two components explain more than 90\% of the variance in every submatrix.

Taken together, the held-out completion sweep gives the operational reason to use rank~2, and the fully-observed SVDs show the matching geometry behind that choice:

\begin{finding}{The $84 \times 133$ model--benchmark score matrix behaves as an effectively rank-2 prediction problem.}{rank2}
\end{finding}

\paragraph{Interpretation of the top two factors.}
We complete the matrix at rank~2 and take the SVD of the completed, column-standardized scores; \Cref{tab:pc_factors} lists the top and bottom models and benchmarks on each of the two factors.
The first component tracks general capability: frontier models sit at the top and small or older models at the bottom.
All $133$ benchmarks load on it with the same sign and with similar magnitude (loadings run from $0.042$ to $0.097$, against $0.087$ for a perfectly uniform direction), so no single benchmark or family drives it, and its per-model scores match the naive capability ranking (each model's mean $z$-scored observed score) at Spearman $0.893$.
Since this component is computed on the completed matrix, we also check it against the observed entries alone: using only the scores actually reported, $131$ of $133$ benchmarks correlate positively with the first-component model score.
The second component instead separates domains.
Its benchmark loadings take both signs, with long-context and coding suites on one side and agentic-search and competition-math on the other, and it does not order models by strength (for example, Claude~3.5 and 3.7 Sonnet rank above GPT-5.5).
Its coding-heavy side matches the task-specialized components in \citet[Figure~2]{ruan2024}, where PC2 emphasizes mathematical and coding benchmarks and PC3 primarily reflects programming capability.
\citet{zhang2025train} show that this second factor can also be related to task preparation: how much a model was trained on the kind of task a benchmark measures.

\begin{table*}[!htbp]
\centering
\caption{\textbf{Top and bottom models and benchmarks on each of the top two factors}, from the SVD of the rank-2 completed, column-standardized score matrix. PC1 orders models by overall capability, with frontier models at the top and small or older models at the bottom, and loads every benchmark with the same sign; PC2 does not order models by capability and instead splits benchmarks by domain.}
\label{tab:pc_factors}
\small
\setlength{\tabcolsep}{4pt}
\renewcommand{\arraystretch}{1.25}
\begin{tabular}{@{}l>{\raggedright\arraybackslash}p{0.40\textwidth}>{\raggedright\arraybackslash}p{0.40\textwidth}@{}}
\toprule
 & PC1 (general capability) & PC2 (domain lean) \\
\midrule
Top-5 models & GPT-5.5, Claude Opus 4.7, Gemini 3.1 Pro, GPT-5.4, DeepSeek-V4-Pro & Claude 3.7 Sonnet, Claude Opus 4.1, Llama 4 Maverick, Claude Opus 4, Claude 3.5 Sonnet \\
Bottom-5 models & Llama 3.2 1B, Gemma 3 1B, GPT-3.5 Turbo, Gemma 2 9B, Llama 3.1 8B & Ministral 8B, Gemma 3 1B, GPT-4.1 nano, Llama 3.2 1B, GPT-5.4 \\
\midrule
Top-5 benchmarks & MultiChallenge, BFCL v3, Vibe-Eval, SWE-Lancer Freelance, Multi-IF & MRCR v1, BigCodeBench, RepoQA, MBPP+, AA Intelligence Index \\
Bottom-5 benchmarks & OmniDocBench 1.5, FACTS Grounding, Safety, MathArena Apex, MRCR v1 & xbench-DeepSearch, HMMT Feb 2026, BrowseComp-ZH, CyberGym, AIME 2026 \\
\bottomrule
\end{tabular}
\end{table*}

\section{\benchpress{}: A Low-rank Benchmark Score Predictor}
\label{sec:bp_methods}

Because the held-out rank sweep in \Cref{sec:bp_svd} selects rank 2, we now ask: can we turn this structure into a predictor for missing benchmark scores?
\Cref{sec:method} introduces the candidate prediction methods; \Cref{sec:method_comparison} moves from these candidates to \benchpress{} by comparing all transform--method combinations on a common experiment setting and selecting the default score predictor; and \Cref{sec:additional_llm_baseline} compares \benchpress{} against LLMs as benchmark score predictors.

\subsection{Candidate Methods}
\label{sec:method}

Suppose a new model arrives with scores on $k$ benchmarks.
How do we predict its scores on the remaining benchmarks?
We decompose the problem into two design choices: (i)~a \emph{feature transform} that reshapes raw scores into a space amenable to linear methods, and (ii)~a \emph{prediction method} that exploits correlations across benchmarks or models.

\paragraph{Feature transforms.}
Let $s\in[0,100]$ denote a benchmark score.
Scores are bounded percentages, and models cluster near the ceiling on easy benchmarks and near the floor on hard ones.
We evaluate seven transforms that address this nonlinearity in different ways:
\begin{itemize}[nosep,leftmargin=*]
  \item \emph{Identity.} Use scores $s$ as-is, with no transformation.
  \item \emph{Log.} Apply $\log(s + 1)$, compressing high scores and stretching low ones.
  \item \emph{Logit.} Apply $\log\!\bigl(s/(100-s)\bigr)$, mapping the bounded score range to an unbounded scale that symmetrically spreads apart scores near both 0 and 100.
  \item \emph{Arcsinh.} Apply $\operatorname{arcsinh}(s/50)$, a smooth approximation to log that is defined at zero.
  \item \emph{Square root.} Apply $\sqrt{s}$, a mild compression that reduces the influence of high scores.
  \item \emph{Probit.} Apply $\Phi^{-1}(s/100)$, where $\Phi$ is the standard normal CDF. Similar to logit but with heavier tails.
  \item \emph{Quantile.} Replace each score with its within-benchmark rank divided by $n+1$, producing uniform marginals. Non-parametric but discards magnitude information.
\end{itemize}
Transforms that assume a $[0,100]$ range (logit, probit, arcsinh, square root) are applied only to percentage-scale benchmarks; the few non-percentage benchmarks (Codeforces rating \citep{codeforces}, Chatbot Arena Elo \citep{chiang2024}, GDPval (Artificial Analysis ELO) \citep{gdpval}) do not suffer from ceiling or floor effects and are left untransformed.
For logit and probit, scores are clipped slightly away from the endpoints before transformation to avoid infinite values.
After applying the chosen transform, we standardize each benchmark column to zero mean and unit variance, so every prediction method operates entirely in the transformed, standardized space.
After prediction, we invert the pipeline in reverse order: first undo the standardization (restoring each column's stored mean and standard deviation), then apply the inverse feature transform (e.g., sigmoid for logit) to map predictions back to the original score scale.
For percentage-scale benchmarks we clip the final predictions to $[0,100]$; non-percentage benchmarks are left unconstrained.

\paragraph{Prediction methods.}
We compare the following methods, each evaluated across multiple feature transforms:
\begin{itemize}[nosep,leftmargin=*]
  \item \emph{Benchmark mean.}
    Predict each missing score as the column average. No tunable parameters.

  \item \emph{Model mean.}
    Adjust the benchmark mean by each model's overall strength percentile. No tunable parameters.

  \item \emph{Benchmark-KNN} (\emph{Bench-KNN}).
    For each missing entry, find the $k$ benchmarks most correlated\footnote{Throughout the paper, ``correlation'' refers to the Pearson correlation: for two columns $a, b$ of length $n$, $\rho(a, b) = \frac{\sum_i (a_i - \bar a)(b_i - \bar b)}{\sqrt{\sum_i (a_i - \bar a)^2 \cdot \sum_i (b_i - \bar b)^2}} \in [-1, 1]$, where $\bar a, \bar b$ are the column means.} with the target benchmark and predict from the model's observed scores on those neighbors, using correlation-based weights. Hyperparameter: $k$ (number of neighbors).

  \item \emph{Model-KNN.}
    Find the $k$ models closest to the target model by root-mean-square distance over shared observed benchmarks, then average their scores on the target benchmark. Hyperparameter: $k$.

  \item \emph{Per-benchmark regression} (\emph{BenchReg}).
    For each target benchmark, BenchReg selects the $k$ most correlated predictor benchmarks, fits one univariate regression per predictor benchmark, and combines the available predictions with $R^2$ weights\footnote{We use the coefficient of determination $R^2 = 1 - \mathrm{SSE}/\mathrm{SST}$, where $\mathrm{SST} = \sum_i (y_i - \bar y)^2$ is the total variance of the target around its mean $\bar y$ and $\mathrm{SSE} = \sum_i (y_i - \hat y_i)^2$ is the residual variance left by the fit, with $y_i$ the values we are trying to predict (the target benchmark's observed scores across shared models in the univariate-regression case) and $\hat y_i$ the corresponding predicted values. $R^2 = 1$ means a perfect fit ($\mathrm{SSE} = 0$), $R^2 = 0$ means the fit does no better than predicting the constant mean $\bar y$, and $R^2 < 0$ means it does worse than the constant mean.}.
    Targets and predictor pairs with fewer than five observations are skipped.
    When a model lacks observations on some predictors, BenchReg uses only the observed predictors; if none are observed, the cell is left unpredicted (coverage $< 100\%$).
    We use an ensemble of univariate regressions rather than a single multivariate model because the number of shared observations per benchmark pair is often very small (5--12 models), making joint estimation of $k$ coefficients prone to overfitting.
    Hyperparameters: $k \in \{3, 5, 7\}$, $R^2_{\min} \in \{0.1, 0.2, 0.3\}$.

  \item \emph{Per-model regression} (\emph{ModelReg}).
    ModelReg is the row-wise counterpart to BenchReg.
    For each target model, it selects the $k$ most correlated predictor models over shared benchmarks, fits univariate regressions from each predictor model's benchmark scores to the target model's scores, and combines the resulting predictions with $R^2$ weights.
    Like BenchReg, it skips targets and predictor pairs with fewer than five observations and can leave a cell unpredicted when no usable predictor has enough shared observations.
    Hyperparameters: $k \in \{3, 5, 7\}$, $R^2_{\min} \in \{0.1, 0.2, 0.3\}$.

  \item \emph{Soft-Impute}.
    Soft-Impute \citep{mazumder2010} iterates between SVD truncation at a chosen rank and re-imputation of missing entries until convergence. We fix the rank to 2 following the held-out rank sweep in \Cref{sec:bp_svd} (no tunable hyperparameters).

  \item \emph{NMF.}
    Non-negative matrix factorization \citep{lee1999nmf}, constraining both factors to be non-negative. Hyperparameter: rank $r$.

  \item \emph{PMF.}
    Probabilistic matrix factorization \citep{salakhutdinov2008pmf} with Gaussian priors on both factors. Hyperparameter: rank $r$.

  \item \emph{Nuclear norm minimization.}
    Convex relaxation of rank minimization \citep{candes2009}: minimize the nuclear norm of the completed matrix plus a squared-error fit on observed entries, with $\lambda$ trading off low rank against data fidelity. Hyperparameter: $\lambda$.

  \item \emph{Bias-decomposed alternating least squares (ALS)} \citep{koren2009}.
    Bias ALS is the full-coverage method that we later adopt.
    After the feature transform and column standardization, let $X\in\mathbb{R}^{M\times B}$ be the transformed score matrix over $M$ models and $B$ benchmarks, let $x_{mb}$ be the observed entry for model $m$ and benchmark $b$, and let $\Omega$ be the observed cells.
    Let $\bar{x}$, $\bar{x}_{m\cdot}$, and $\bar{x}_{\cdot b}$ be the observed global, model, and benchmark means in this transformed space.
    For rank $R$ and regularization $\lambda$, Bias ALS fits $U\in\mathbb{R}^{M\times R}$ and $V\in\mathbb{R}^{B\times R}$ by
    \[
    \begin{aligned}
    (U,V)=\arg\min_{U,V}\;&
    \sum_{(m,b)\in\Omega}
    \Bigl[
      x_{mb}
      -\bigl(\bar{x}+(\bar{x}_{m\cdot}-\bar{x})+(\bar{x}_{\cdot b}-\bar{x})\bigr)
      -(UV^\top)_{mb}
    \Bigr]^2 \\
    &+\lambda\left(\|U\|_F^2+\|V\|_F^2\right).
    \end{aligned}
    \]
    Its prediction is the sum of a global level, a model offset, a benchmark offset, and a rank-$R$ residual correction:
    \[
      \hat{x}_{mb}
      =
      \underbrace{\bar{x}}_{\text{global level}}
      +\underbrace{(\bar{x}_{m\cdot}-\bar{x})}_{\text{model }m\text{ offset}}
      +\underbrace{(\bar{x}_{\cdot b}-\bar{x})}_{\text{benchmark }b\text{ offset}}
      +\underbrace{(UV^\top)_{mb}}_{\text{rank-}R\text{ residual correction}}.
    \]
    The biases absorb row and column offsets, so the low-rank term only has to model residual model--benchmark interaction structure.
    ALS updates each block in closed form via ridge regression on observed entries; we ensemble-average over multiple random initializations to reduce sensitivity to local minima.
    We fix the rank to 2 following the held-out rank sweep in \Cref{sec:bp_svd}; the only tunable hyperparameter is the regularization $\lambda$.

  \item \emph{Neural baseline} (\emph{MLP}).
    A 2-layer MLP (hidden dimension 32) with binary mask for missing entries, trained for 500 epochs. Hyperparameter: learning rate.
\end{itemize}
Full definitions of all prediction methods, including equations and fallback rules, are in \Cref{app:method_details}.

\subsection{From Candidate Methods to \benchpress{}}
\label{sec:method_comparison}

We evaluate all combinations of the seven feature transforms and twelve prediction methods, selecting hyperparameters independently for each combination.
For each model, we randomly hide half of its known benchmark scores, train on the remaining half (plus all other models' data), predict the hidden scores, and measure error.
We use 3 folds per seed and 10 seeds (${\sim}20{,}000$ test predictions per pair).
Each (transform, method) pair follows a four-stage pipeline: (i)~apply the feature transform, (ii)~standardize each column, (iii)~run the prediction method, (iv)~invert both transforms to recover original-scale predictions.
Predictions on percentage-scale benchmarks are clipped to $[0, 100]$; omitting standardization degrades most methods, especially NMF and PMF.

\paragraph{Metrics.}
Using the notation from \Cref{sec:bp_data}, we measure prediction quality with two score-error metrics:
\emph{(i)}~\emph{median absolute percentage error} ($\mathsf{MedAPE}$\,$\downarrow$), computed within each held-out fold as the median of ${|\hat{s} - s|}/{|s|} \times 100\%$ and then summarized by the median over folds, and
\emph{(ii)}~\emph{median absolute error} ($\mathsf{MedAE}$\,$\downarrow$), computed analogously from $|\hat{s} - s|$ in raw score points.
We use median-based score-error metrics because the error distribution is heavy-tailed: a handful of near-zero scores inflate percentage errors and a few large misses would dominate the mean, so the median better reflects typical accuracy.
In the analyses below, the figures and reported numbers summarize these per-record errors by their median rather than their mean.
We also report \emph{coverage}, the fraction of held-out entries for which a method produces a finite prediction, because some methods (e.g., regression) cannot predict when insufficient correlated data exists.

\paragraph{Hyperparameter selection.}
For each (transform, method) pair we grid-search over the hyperparameters listed below, selecting the configuration with the lowest MedAPE on validation cells split from each fold's training data:
\begin{itemize}[nosep,leftmargin=*]
  \item \emph{Benchmark Mean, Model Mean.} No tunable parameters.
  \item \emph{Bench-KNN, Model-KNN.} Number of neighbors $k \in \{3, 5, 7, 10\}$.
  \item \emph{BenchReg, ModelReg.} Number of predictors $k \in \{3, 5, 7\}$, minimum correlation $R^2_{\min} \in \{0.1, 0.2, 0.3\}$ (9 configurations each).
  \item \emph{Soft-Impute.} No tunable hyperparameters; rank fixed at 2.
  \item \emph{Bias-decomposed ALS.} Regularization $\lambda \in \{0.01, 0.1, 1.0\}$; rank fixed at 2.
  \item \emph{NMF.} Rank $r$ in $\{1, 2, 3, 5\}$.
  \item \emph{PMF.} Rank $r$ in $\{1, 2, 3, 5\}$.
  \item \emph{Nuclear Norm.} Regularization $\lambda \in \{0.1, 0.5, 1.0, 5.0\}$.
  \item \emph{MLP.} Learning rate $\in \{10^{-4}, 10^{-3}, 10^{-2}\}$; architecture fixed at 2 layers with hidden dimension 32 and 500 training epochs.
\end{itemize}
All reported errors, including the leaderboard below, are measured on the held-out folds, which take no part in selection; \Cref{tab:model_selection} gives the corresponding validation-error leaderboard.

\paragraph{Results.}
\Cref{tab:top15} ranks the best-performing configurations by the two score-error metrics.
Several patterns emerge:
(i)~BenchReg and ModelReg dominate the top score-error entries, but their coverage is not always complete: some cells are left blank rather than predicted.
(ii)~Among methods that predict every missing cell, Logit Bias ALS is the strongest and remains very close to the best regression entries.
(iii)~The leading configurations are not separated by a large qualitative gap: related logit/probit variants and regularization choices give similar performance.
We therefore use Logit Bias ALS with $\lambda=0.1$ and rank 2 as \benchpress{}'s default score predictor because it sits near the top of the leaderboard, has full coverage, and keeps the downstream error and reliability analyses tied to a single simple configuration.
The full $7 \times 12$ transform--method grid is in \Cref{app:method_comparison}.

\begin{finding}{Logit-transformed, bias-decomposed alternating least squares (ALS) matrix completion \citep{koren2009}, with rank 2 and regularization 0.1, gives near-best score-prediction accuracy while predicting every missing model--benchmark score.}{method_comparison}
\end{finding}

\begin{table*}[!htbp]
\centering
\caption{Top-15 transform--method configurations ranked independently by score-error metric. \# is per-metric rank; values are shown as metric value followed by coverage in parentheses, e.g.\ 7.8 (100\%).
All results use standardization and report the median over 10 seeds $\times$ 3 folds.
Pink highlights mark the Logit Bias ALS configuration adopted as \benchpress{}'s main score predictor.
}
\label{tab:top15}
\resizebox{\textwidth}{!}{%
\begin{tabular}{@{}rlllr@{\hspace{12pt}}rlllr@{}}
\toprule
\multicolumn{5}{c}{\textbf{MedAPE (\%) $\downarrow$}} &
\multicolumn{5}{c}{\textbf{MedAE $\downarrow$}} \\
\cmidrule(r){1-5} \cmidrule(l){6-10}
\# & Transform & Method & Hyperparameter & Value &
\# & Transform & Method & Hyperparameter & Value \\
\midrule
1 & Probit & ModelReg & $R^2_{\min}{=}0.2$, $k{=}7$ & 7.7 (82\%) & 1 & Logit & Bias ALS & $\lambda{=}0.01$, $r{=}2$ & 4.62 (100\%) \\
2 & Probit & ModelReg & $R^2_{\min}{=}0.1$, $k{=}5$ & 7.7 (74\%) & 2 & Probit & Bias ALS & $\lambda{=}0.1$, $r{=}2$ & 4.62 (100\%) \\
3 & Probit & ModelReg & $R^2_{\min}{=}0.3$, $k{=}7$ & 7.7 (82\%) & 3 & \cBP\textbf{Logit} & \cBP\textbf{Bias ALS} & \cBP\textbf{$\lambda{=}0.1$, $r{=}2$} & \cBP\textbf{4.63 (100\%)} \\
4 & Probit & BenchReg & $R^2_{\min}{=}0.2$, $k{=}7$ & 7.7 (85\%) & 4 & Probit & BenchReg & $R^2_{\min}{=}0.3$, $k{=}7$ & 4.64 (84\%) \\
5 & Probit & ModelReg & $R^2_{\min}{=}0.1$, $k{=}7$ & 7.7 (83\%) & 5 & Logit & BenchReg & $R^2_{\min}{=}0.3$, $k{=}7$ & 4.66 (84\%) \\
6 & Probit & ModelReg & $R^2_{\min}{=}0.3$, $k{=}5$ & 7.7 (74\%) & 6 & Quantile & Bias ALS & $\lambda{=}0.1$, $r{=}2$ & 4.66 (100\%) \\
7 & Probit & ModelReg & $R^2_{\min}{=}0.2$, $k{=}5$ & 7.7 (74\%) & 7 & Probit & ModelReg & $R^2_{\min}{=}0.3$, $k{=}5$ & 4.66 (74\%) \\
8 & Probit & BenchReg & $R^2_{\min}{=}0.3$, $k{=}7$ & 7.8 (84\%) & 8 & Probit & BenchReg & $R^2_{\min}{=}0.2$, $k{=}7$ & 4.66 (85\%) \\
9 & Logit & ModelReg & $R^2_{\min}{=}0.3$, $k{=}5$ & 7.8 (74\%) & 9 & Probit & ModelReg & $R^2_{\min}{=}0.2$, $k{=}5$ & 4.67 (74\%) \\
10 & Probit & BenchReg & $R^2_{\min}{=}0.1$, $k{=}7$ & 7.8 (86\%) & 10 & Probit & ModelReg & $R^2_{\min}{=}0.1$, $k{=}5$ & 4.67 (74\%) \\
11 & Logit & BenchReg & $R^2_{\min}{=}0.1$, $k{=}7$ & 7.8 (86\%) & 11 & Logit & BenchReg & $R^2_{\min}{=}0.1$, $k{=}7$ & 4.67 (86\%) \\
12 & \cBP\textbf{Logit} & \cBP\textbf{Bias ALS} & \cBP\textbf{$\lambda{=}0.1$, $r{=}2$} & \cBP\textbf{7.8 (100\%)} & 12 & Probit & ModelReg & $R^2_{\min}{=}0.2$, $k{=}3$ & 4.67 (60\%) \\
13 & Logit & BenchReg & $R^2_{\min}{=}0.3$, $k{=}7$ & 7.8 (84\%) & 13 & Probit & ModelReg & $R^2_{\min}{=}0.3$, $k{=}3$ & 4.67 (60\%) \\
14 & Logit & BenchReg & $R^2_{\min}{=}0.2$, $k{=}7$ & 7.8 (85\%) & 14 & Probit & ModelReg & $R^2_{\min}{=}0.1$, $k{=}3$ & 4.67 (60\%) \\
15 & Probit & Bias ALS & $\lambda{=}0.1$, $r{=}2$ & 7.8 (100\%) & 15 & Logit & BenchReg & $R^2_{\min}{=}0.2$, $k{=}7$ & 4.67 (85\%) \\
\bottomrule
\end{tabular}

}
\end{table*}

\paragraph{The \benchpress{} predictor.}
Because Logit Bias ALS is the strongest full-coverage configuration in the comparison above, we adopt this configuration as \benchpress{}'s default score predictor for the rest of the paper.

\begin{titlebox}[title={\textsc{BenchPress}: Point Prediction Recipe}]
Given a partially observed model--benchmark score matrix, \benchpress{} predicts every missing cell as follows:
\begin{enumerate}[nosep,leftmargin=*]
  \item Transform percentage scores with the logit transform; leave non-percentage scores on their native scale.
  \item Standardize each benchmark column using the observed training entries.
  \item Fit bias-decomposed alternating least squares (ALS) matrix completion \citep{koren2009} with a global level, model offsets, benchmark offsets, and a rank-2 residual interaction ($\lambda=0.1$).
  \item Invert the standardization and feature transform to return predictions on the original benchmark scale.
\end{enumerate}
\end{titlebox}

\subsection{\benchpress{} vs. LLMs as Benchmark Score Predictors}
\label{sec:additional_llm_baseline}

\benchpress{} predicts from the observed score matrix alone.
Another natural question is whether a frontier LLM can predict a benchmark score directly from the target model, the target benchmark, and a few nearest-peer examples.
This is a per-cell LLM predictor: each target score is queried separately, and the prompt may expose public model and benchmark identities.

\begin{wraptable}{r}{0.3\textwidth}
\vspace{-0.3in}
\centering
\small
\setlength{\tabcolsep}{2.5pt}
\caption{\textbf{LLM score prediction from peer examples.}
The informed prompt sees real model and benchmark names; the blind prompt does not. Lower is better.}
\label{tab:llm_five_shot}
\vspace{-.12in}
\resizebox{\linewidth}{!}{%
\begin{tabular}{@{}l c r r@{}}
\toprule
Predictor & Names & MedAPE & MedAE \\
\midrule
GPT-5.5 & \cmark & 5.86 & 3.50 \\
GPT-5.5 & \xmark & 7.89 & 4.70 \\
\midrule
\cBP\textbf{\benchpress{}} & \cBP\textbf{\xmark} & \cBP\textbf{7.77} & \cBP\textbf{4.63} \\
\bottomrule
\end{tabular}}
\vspace{-0.12in}
\end{wraptable}
\paragraph{Experiment setting.}
We use the same held-out cells as the method-comparison experiment in \Cref{sec:method_comparison}.
For each target cell (model, benchmark), we select peer examples using only the training matrix for that fold.
A candidate peer model must have an observed score on the target benchmark and share at least five visible benchmarks with the target model.
Among eligible peers, we choose the five models with the highest Pearson correlation to the target model over shared visible scores.
The prompt then asks GPT-5.5 to predict the target model's score on the target benchmark from these five peer examples.
We consider two scenarios.
In the \emph{informed} condition, the prompt keeps the real model and benchmark names.
In the \emph{blind} condition, model and benchmark identifiers are anonymized within the prompt, while the scores and peer-example structure are preserved.
The informed condition tests whether a frontier LLM can exploit public model and benchmark semantics; the blind condition tests whether the numerical peer structure alone is enough.
We score both conditions on the same held-out cells as \benchpress{} using MedAPE and MedAE.
The exact prompt template is given in \Cref{app:llm_five_shot_prompt}.

\paragraph{Results.}

\Cref{tab:llm_five_shot} shows that the informed prompt is a strong baseline and confirms that a frontier LLM can often predict held-out benchmark scores from nearest-peer examples.
But this is not the same capability as score prediction from the matrix alone: names give the LLM access to public model reputations, benchmark semantics, and possibly memorized leaderboard facts.
The blind condition removes that channel and is therefore the diagnostic comparison.
There, the LLM is close to \benchpress{} but not a cheaper or more reliable replacement: it still requires paid generation over many target cells, while \benchpress{} fits the score matrix once and predicts every missing cell deterministically.

\begin{finding}{A frontier LLM can predict benchmark scores when real names are visible, but that advantage depends on model and benchmark memory. In the blind setting, \benchpress{} is more accurate and more scalable because it fits the score matrix once rather than querying an LLM over target cells.}{llm_completer}
\end{finding}

\section{What \benchpress{} Enables for Model Evaluation}
\label{sec:findings}

With the default score predictor fixed, we now evaluate what \benchpress{} enables across realistic model-evaluation tasks.
Three questions guide this section.
\textbf{First}, under a fixed evaluation budget, which probe benchmarks should a practitioner run so that \benchpress{} best recovers the model's scorecard (\Cref{sec:probe_selection})?
\textbf{Second}, do \benchpress{}'s predicted scores preserve same-benchmark model rankings well enough that practitioners can use them to compare models (\Cref{sec:ranking_preservation})?
\textbf{Third}, when a brand-new model is released after the matrix was assembled, can \benchpress{} still produce useful predictions from a small seed evaluation (\Cref{sec:temporal_deployment})?

Unless otherwise stated, all analyses in this section use the default \benchpress{} score predictor from \Cref{sec:method_comparison}, and final metrics are reported after mapping predictions back to the original raw-score scale.

\subsection{Budgeted Scorecard Recovery}
\label{sec:probe_selection}

The rank-2 structure identified in \Cref{sec:bp_svd} suggests that benchmark scores contain substantial shared information, rather than $133$ independent measurements. Direct evidence supports this: most benchmarks can be predicted from the rest with low error (\Cref{app:per_benchmark_predictability}), and almost every benchmark column has at least one strongly correlated peer (\Cref{app:probe_selection}). Many scores can therefore plausibly be inferred from a small amount of carefully chosen evidence.
We therefore ask the operational version of the same question: if a practitioner can run only a few benchmarks on a new model, which scores should be measured, and which scores can \benchpress{} infer from them?

\paragraph{Experiment setting.}
We simulate a practitioner who evaluates each target model on a fixed probe set and then asks \benchpress{} to complete the rest of that model's public scorecard.
For a target model, only the probe columns remain visible in its row; all other models keep their observed rows.
We evaluate every observed model-benchmark cell at every budget, so the denominator is fixed across probe-set sizes.
Observed probe cells are counted as exact predictions with zero error, and all remaining observed cells for the target model are predicted from the masked matrix.
Thus the curves measure how much of the current score matrix can be reconstructed from a small number of selected probes.
They should not be read as held-out transfer estimates for a fixed universal probe set, because the probe identities are themselves selected on this current score matrix.

Starting from an empty probe set, we build a ten-benchmark set greedily.
At each step, we try every remaining candidate benchmark, temporarily add it to the current probe set, evaluate pooled error on the fixed universe, and keep the candidate with the lowest error.
We compare two greedy probe-set methods against a random baseline:
\begin{itemize}[leftmargin=*]
    \item \textbf{Cost-unaware greedy:} any benchmark can be selected as the next probe.
    \item \textbf{Cost-aware greedy:} candidates are restricted by the low-cost allowlist.
    \item \textbf{Random baseline:} we run 10 seeds. Each seed draws one global random benchmark ordering, and each budget uses the corresponding prefix for every target model. The plotted line is the mean across seeds; the shaded band is the 25th--75th percentile range.
\end{itemize}

\begin{table}[!t]
\centering
\caption{\textbf{Top-10 probe sets selected by each objective.} Each row uses the same set of observed model--benchmark cells and the same greedy procedure; only the objective and candidate pool change.}
\label{tab:probe_sets}
\scriptsize
\setlength{\tabcolsep}{4pt}
\resizebox{\textwidth}{!}{%
\begin{tabular}{@{}l l p{0.78\textwidth}@{}}
\toprule
Objective & Candidate pool & Top-10 prefix \\
\midrule
\multirow{2}{*}{\raisebox{-14pt}{MedAPE}} & Any benchmark & GPQA Diamond; HLE; AIME 2024; MMLU-Pro; ARC-AGI-1; ARC-AGI-2; Aider Polyglot (diff mode); LiveCodeBench; Terminal-Bench 2.0; SWE-bench Verified \\
\cmidrule(lr){2-3}
 & Low-cost benchmarks & GPQA Diamond; MMLU-Pro; Aider Polyglot (diff mode); MathArena Apex 2025; HMMT Nov 2025; Bullshit-Bench (Clear Pushback); MATH-500; AIME 2025; Arena-Hard Auto; IFBench \\
\midrule
\multirow{2}{*}{\raisebox{-14pt}{MedAE}} & Any benchmark & GPQA Diamond; HLE; Codeforces Rating; MMLU-Pro; ARC-AGI-1; Terminal-Bench 2.0; MATH-500; HLE Text; AIME 2025; GDPval (Artificial Analysis ELO) \\
\cmidrule(lr){2-3}
 & Low-cost benchmarks & GPQA Diamond; MMLU-Pro; Aider Polyglot (diff mode); MATH-500; AIME 2026; AIME 2025; Bullshit-Bench (Clear Pushback); Arena-Hard Auto; $\tau^2$-bench Airline; IFBench \\
\bottomrule
\end{tabular}}
\end{table}

\begin{figure*}[!t]
\centering
\begin{minipage}[t]{0.30\textwidth}
  \centering
  \includegraphics[width=\linewidth]{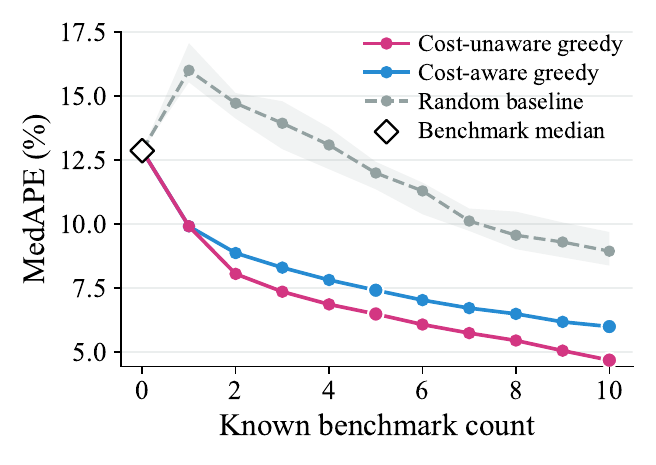}
  \vspace{-0.8em}
  \captionof{figure}{\textbf{MedAPE during probe-set construction.} Pooled MedAPE decreases as selected benchmark scores are revealed; every budget is evaluated on the same observed cells.}
  \label{fig:probe_eval}
\end{minipage}\hfill
\begin{minipage}[t]{0.23\textwidth}
  \centering
  \includegraphics[width=\linewidth]{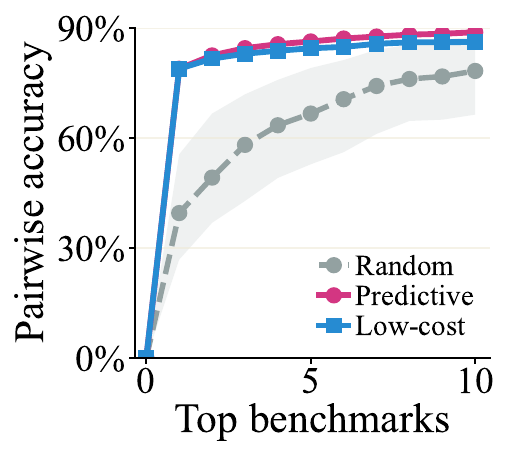}
  \vspace{-0.6em}
  \captionof{figure}{\textbf{Overall ranking preservation.} Pairwise ranking accuracy as the probe budget grows.}
  \label{fig:ranking_preservation_overall}
\end{minipage}\hfill
\begin{minipage}[t]{0.39\textwidth}
  \centering
  \includegraphics[width=\linewidth]{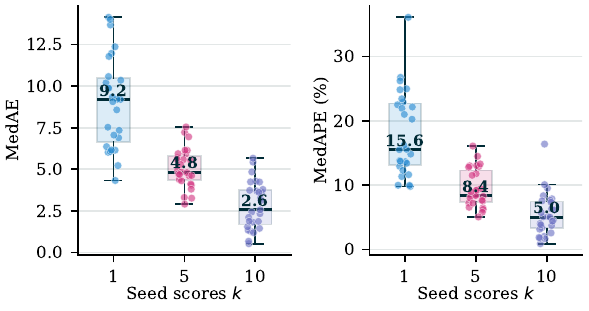}
  \vspace{-0.6em}
  \captionof{figure}{\textbf{Predicting newly released models under a pre-specified temporal window.} Each dot is one target model; boxes show the median and interquartile range across 27 targets.}
  \label{fig:temporal_deployment}
\end{minipage}
\end{figure*}

\paragraph{Results.}
The greedy procedure yields four ten-probe sets (\Cref{tab:probe_sets}), one for each combination of \emph{objective} (MedAPE or MedAE) and \emph{candidate pool} (any benchmark, or the low-cost allowlist).
All four sets lean heavily on reasoning- and math-oriented benchmarks (GPQA Diamond, ARC-AGI, MATH-500, multiple AIME and HMMT contests, and similar): reasoning is a dominant axis of variation in the score matrix, so these benchmarks supply the cleanest signal for \benchpress{} to triangulate the rest of a model's profile.
\Cref{fig:probe_eval} (MedAPE) and \Cref{fig:hero_overall} (MedAE) trace the corresponding error curves as the probe budget grows from one to ten.
Two qualitative trends emerge.
First, cost-aware greedy tracks the cost-unaware curve closely at every budget; restricting probes to the low-cost allowlist costs surprisingly little, because the cleanest reasoning-axis probes are already low-cost.
Second, both greedy curves clearly separate from the random baseline at every budget: \emph{which} benchmarks are chosen matters far more than how many.
A more exhaustive probe-selection analysis in \Cref{app:probe_selection} shows that greedy probe selection is cost-effective: its performance matches or differs only marginally from more exhaustive search.

\begin{remark}[Near-duplicate probes from the same benchmark family]
Greedy occasionally selects two versions of the same benchmark family (e.g., two AIME contests in the low-cost MedAE run, ARC-AGI-1 and ARC-AGI-2 in the unrestricted MedAPE run). This is not a sign that the second version adds a new axis; it is noise averaging on an axis with no good substitute. In the low-cost setting, once the highest-signal reasoning probes (GPQA Diamond, MMLU-Pro, Aider Polyglot, MATH-500) are exhausted, the remaining low-cost candidates are mostly near-duplicate math contests. In the unrestricted MedAPE setting, ARC-AGI captures a frontier-reasoning axis with very low absolute scores, where small errors translate into large percentage errors, so a second ARC-AGI instance is the highest-marginal way to reduce error on that axis.
\end{remark}

\begin{remark}[Comparability with \Cref{sec:method_comparison}]
The setting differs from the 50\% per-model holdout: only the target model's non-probe cells are masked, so \benchpress{} sees richer context and the endpoints are not directly comparable to \Cref{sec:method_comparison}.
\end{remark}

\begin{remark}[Snapshot-specific probe sets]
For a different matrix or snapshot, this construction should be rerun: the current probe set is a snapshot-specific recommendation, not a permanent benchmark list.
\end{remark}

\begin{finding}{With only five probe benchmarks, \benchpress{} predicts the remaining score profile to a pooled MedAE of $3.93$ score points; restricting probes to the low-cost allowlist still reaches $4.55$ score points.}{probe_recipe}
\end{finding}

\subsection{Preserving Model Rankings}
\label{sec:ranking_preservation}

The practical value of score prediction is not only numerical accuracy, but whether the predictions support the same evaluation decisions.
Here the operational question is simple: when two models differ meaningfully on the same benchmark, does \benchpress{} preserve which model is better?

\paragraph{Experiment setting.}
We reuse the holdout setting of \Cref{sec:method_comparison} (10 seeds, three folds per model) and complete each benchmark leaderboard with true scores on seen cells and \benchpress{} predictions on held-out cells.
Because adjacent leaderboard slots are often separated by tiny gaps that a small prediction error can flip, we evaluate margin-aware pairwise ordering rather than exact ranks.
An auxiliary evaluation of whether the predicted leaderboard recovers the true top-ranked models, together with ranking-aware probe selection, is reported in \Cref{app:ranking_preservation}.
For each benchmark, we form all same-benchmark model pairs with at least one held-out cell, keep those whose true score gap is at least the row's margin, and report the median across benchmarks of
\[
\text{pairwise accuracy}
=
\frac{\#\{\text{comparable pairs whose completed order matches the true order}\}}
{\#\{\text{comparable pairs}\}}.
\]

The margin-0 row includes every non-tied pair and is most sensitive to near-ties; larger margins focus on clearer model differences.

\paragraph{Results.}
\begin{wraptable}[7]{r}{0.22\textwidth}
\vspace{-1.8em}
\centering
\scriptsize
\setlength{\tabcolsep}{3.0pt}
\caption{\textbf{Pairwise ranking preservation.}}
\label{tab:ranking_preservation}
\vspace{-0.4em}
\begin{tabular}{@{}crr@{}}
\toprule
Margin & Accuracy & \# Pairs \\
\midrule
0 & $83.8\%$ & $589{,}830$ \\
1 & $86.3\%$ & $561{,}283$ \\
2 & $88.0\%$ & $531{,}498$ \\
5 & $92.1\%$ & $454{,}090$ \\
\bottomrule
\end{tabular}
\vspace{-0.8em}
\end{wraptable}

The margin-aware results in \Cref{tab:ranking_preservation} show that score-prediction errors rarely overturn meaningful ordering decisions.
The margin-0 row is lower because it includes many near-tied model pairs where either ordering is fragile.
At a two-point score margin, \benchpress{} achieves $88.0\%$ pairwise ranking accuracy across $531{,}498$ comparable pairs.
When the true score gap is at least five points, pairwise ranking accuracy rises to $92.1\%$.
We additionally plot pairwise ranking accuracy as a probe budget grows from one to ten benchmarks (\Cref{fig:ranking_preservation_overall}); both informativeness-greedy and cost-aware probe sets stay well above the random baseline, and the underlying greedy probe-set selection is detailed in \Cref{app:ranking_preservation}.

\vspace{0.4em}
\begin{finding}{For same-benchmark model pairs separated by at least five score points, \benchpress{}'s predicted scores yield the correct ordering $92.1\%$ of the time.}{ranking_preservation}
\end{finding}

\subsection{Predicting Newly Released Models}
\label{sec:temporal_deployment}

The evaluations so far hide cells from the same matrix used to fit \benchpress{}, so every model in the test set has already contributed training signal elsewhere in the matrix.
Real deployment is stricter: when a new model is released, the matrix was assembled before that release and contains no information about the new model.
We therefore ask whether \benchpress{} can still produce useful scores for a brand-new model from the historical matrix plus a small seed evaluation on that model.

\paragraph{Experiment setting.}
We evaluate an intermediate segment of the release timeline, chosen using only release metadata and matrix coverage before inspecting prediction errors.
This choice avoids two uninformative extremes: very early targets leave too few older models in the training matrix, while the latest releases would be predicted from almost the full snapshot rather than a meaningfully historical matrix.
Concretely, we use models from the post-DeepSeek-R1 reasoning era through GPT-5.1, and keep only models with more than 20 observed benchmark scores.
The coverage threshold ensures that, after revealing up to ten seed scores, each target still has enough hidden cells for a meaningful per-model error estimate.
This yields 27 target models across the recent reasoning-era release window.
For each target model, we train \benchpress{} on only the models released before the target's release date, so the predictor sees no information about the new release beyond what we explicitly reveal.
We then reveal $k \in \{1, 5, 10\}$ of that target model's observed benchmark scores and predict the rest, repeating each setting over $10$ random seeds and reporting the median.
Revealed cells contribute zero error to the pooled metric, hidden cells with finite predictions enter MedAPE and MedAE, and hidden cells without a finite deployment prediction are dropped.

\paragraph{Results.}
Two patterns stand out across the 27 targets in \Cref{fig:temporal_deployment}.
First, even with strict time cutoffs, revealing a small seed set sharply reduces prediction error: the median target drops from $9.20$ MedAE at $k{=}1$ to $4.83$ at $k{=}5$ and $2.57$ at $k{=}10$.
The corresponding MedAPE drops from $15.57\%$ to $8.40\%$ and then $5.02\%$.
Second, the distribution narrows as more seed scores are revealed, showing that the gain is not driven by only a few easy releases.
A small seed evaluation on the new release contributes more than additional historical models.

\begin{finding}{Across 27 target models in a pre-specified temporal window, five seed scores bring \benchpress{}'s predictions within $4.83$ points of the true value, and ten seeds tighten this to $2.57$ points.}{temporal_deployment}
\end{finding}

\section{When to Trust \benchpress{}'s Predictions}
\label{sec:trust}

The practical question is not only whether \benchpress{} can fill in missing scores, but when those filled-in scores are safe to use.
This section explains when to trust the default \benchpress{} score predictor selected in \Cref{sec:method_comparison}.
We first identify benchmark- and model-side factors associated with prediction quality, then use those signals together with predictor disagreement to estimate prediction reliability.

\subsection{What Affects Prediction Reliability}
\label{sec:reliability_analysis}

The natural next question is where prediction error comes from.
Some sources of error may be benchmark-side: sparse observations, weak benchmark neighbors, or score distributions that are intrinsically hard to interpolate.
Others may be model-side: sparse model rows, weak peers, provider effects, scale, or recency.

\paragraph{Methodology.}
In what follows, we propose a set of hypotheses about why \benchpress{} mispredicts certain cells: seven targeting benchmark-side factors and nine targeting model-side factors.
Each hypothesis is phrased as a no-effect claim: a candidate factor is not associated with \benchpress{}'s prediction quality, measured by MedAPE and MedAE.

We assess each one with a standard statistical hypothesis test that returns a $p$-value: assuming the no-effect hypothesis were true, this is the probability that random chance alone would produce data deviating from ``no effect'' by at least as much as ours.
A small $p$ means such data would be extremely unlikely under the hypothesis, so the observation is inconsistent with the hypothesis and we reject it; a large $p$ means our data is well within what random chance could produce, so we have no grounds to reject.

Throughout this section we treat $p<0.01$ as our rejection threshold: when $p<0.01$ we reject the hypothesis, i.e., the data provide strong evidence that the factor \emph{does} matter; otherwise we fail to reject (which may either mean the factor truly has no effect, or that we lack the sample size to detect one).
Depending on the shape of the hypothesis, we use one of two tests:

\begin{itemize}[nosep,leftmargin=*]
    \item \emph{Spearman rank correlation test.} We use this for observational hypotheses, where each benchmark or model contributes a measured feature and its \benchpress{} prediction error. Spearman tests whether higher feature values are monotonically associated with higher or lower error, while being less sensitive to outliers than raw-value correlation.
    \item \emph{Paired Wilcoxon signed-rank test.} We use this for intervention-style hypotheses, where the same benchmark or model is evaluated under a baseline setting and an ablated setting. The paired design controls for inherent target difficulty, and the rank-based test is more reliable than a paired $t$-test for our heavy-tailed error shifts.
\end{itemize}

\Cref{app:reliability_analysis} gives the full test definitions, approximations, and $p$-value calculations.

\paragraph{Benchmark analysis.}
\label{sec:bench_predict}
What determines whether a benchmark is easy or hard for \benchpress{} to predict?
We test seven hypotheses split into two families.
H1--H3 probe benchmark-intrinsic features (low-rank fit, score level, score spread), each evaluated by univariate Spearman correlation between the feature and \benchpress{}'s per-benchmark error across all 133 targets.
H4--H7 probe data availability and structural overlap with other benchmarks, each evaluated by a paired hide-half ablation: we intervene on the training matrix and compare prediction quality against an unintervened baseline (paired Wilcoxon over benchmarks).

\begin{itemize}[nosep,leftmargin=*]
\item \emph{H1 Low-rank fit.} \emph{Intuition:} \benchpress{} fills scores from a simple two-factor pattern, so a benchmark whose scores already follow that pattern is easier to reconstruct. \emph{Hypothesis:} a benchmark's column $R^2$ under the rank-2 SVD reconstruction is not associated with how well it can be predicted. \emph{Feature:} column $R^2$ under the rank-2 reconstruction of the standardized, zero-imputed score matrix. \emph{Test:} Spearman.
\item \emph{H2 Score level.} \emph{Intuition:} on a very easy (or very hard) benchmark almost every model scores near the top (or bottom), so the scores bunch together and are easy to guess. \emph{Hypothesis:} the overall score level (i.e., difficulty) of a benchmark is not associated with how well it can be predicted. \emph{Feature:} median observed score per benchmark. \emph{Test:} Spearman.
\item \emph{H3 Score spread.} \emph{Intuition:} if all models score about the same on a benchmark, its scores are nearly flat and easy to guess; when scores are spread out, there is more room to be wrong. \emph{Hypothesis:} the spread of scores across models on a benchmark is not associated with how well it can be predicted. \emph{Feature:} standard deviation of observed scores per benchmark. \emph{Test:} Spearman.
\item \emph{H4 Target coverage.} \emph{Intuition:} the more of a benchmark's own scores we have already observed, the more the predictor has to work from when filling the rest. \emph{Hypothesis:} reducing the amount of training evidence for a target benchmark does not change its prediction error. \emph{Intervention:} for each target benchmark we first split its observed cells in half: one half is held out for evaluation and the other half remains available for training; we then compare the full training half against a version where three quarters of those training cells are removed. \emph{Test:} paired Wilcoxon.
\item \emph{H5 Strong-neighbor presence.} \emph{Intuition:} if another benchmark's scores track the target's closely, we can read the target off that neighbor; remove the neighbor and that shortcut disappears. \emph{Hypothesis:} masking strongly correlated neighbor benchmarks does not change a target benchmark's prediction error. \emph{Intervention:} for each target benchmark, mask every neighbor benchmark whose Pearson correlation with the target is at least 0.85 on shared models, then rerun \benchpress{} on the target's held-out cells. \emph{Test:} paired Wilcoxon.
\item \emph{H6 Strong-neighbor support.} \emph{Intuition:} even with such a neighbor, the prediction can only be as good as the number of models the two benchmarks were both scored on. \emph{Hypothesis:} reducing overlapping evidence from the strongest neighbor does not change a target benchmark's prediction error. \emph{Intervention:} for each target benchmark, identify its strongest neighbor, keep only the models scored by both benchmarks, and compare the full shared-evidence condition against a version where three quarters of those overlapping neighbor cells are removed. \emph{Test:} paired Wilcoxon.
\item \emph{H7 Same-category evidence.} \emph{Intuition:} if a benchmark is predictable only because it shares a topic (e.g., other math benchmarks) with its neighbors, dropping all same-topic benchmarks should break it; if it stays accurate, topic was not what mattered. \emph{Hypothesis:} masking same-category benchmarks does not change a target benchmark's prediction error. \emph{Intervention:} mask all same-category benchmarks during training (43~benchmarks $\times$ 10~seeds). \emph{Test:} paired Wilcoxon.
\end{itemize}

\begin{table*}[!htbp]
\centering
\caption{\textbf{Which features predict per-benchmark prediction quality?} H1--H3 use the Spearman rank correlation test across target benchmarks; H4--H7 use paired Wilcoxon signed-rank tests on hide-half ablations. The $p$-value is the probability of seeing an effect at least this large by chance if the listed (no-effect) hypothesis were true; smaller $p$ means stronger evidence against it. We reject a hypothesis when $p<0.01$ (bold), i.e., the data provide strong evidence that the factor does matter. Pink rows are rejected under both MedAPE and MedAE and are visualized in \Cref{fig:predictability_factors_51}.}
\label{tab:predictability_factors}
\vspace{0.3em}
\small
\renewcommand{\arraystretch}{1.22}
\setlength{\tabcolsep}{2.0pt}
\begin{tabular}{@{}>{\raggedright\arraybackslash}p{0.40\textwidth}>{\centering\arraybackslash}p{0.28\textwidth}>{\centering\arraybackslash}p{0.28\textwidth}@{}}
\toprule
Hypothesis & MedAPE $\downarrow$ ($p$-value) & MedAE $\downarrow$ ($p$-value) \\
\midrule
H1\; Low-rank fit            & $p=0.150$              & $p=0.036$ \\
H2\; Score level             & \textbf{$p<0.001$}    & $p=0.054$ \\
\cBP \textbf{H3\; Score spread}       & \cBP \textbf{$p<0.001$} & \cBP \textbf{$p<0.001$} \\
\midrule
\cBP \textbf{H4\; Target coverage}    & \cBP \textbf{$p<0.001$} & \cBP \textbf{$p<0.001$} \\
\cBP \textbf{H5\; Strong-neighbor presence} & \cBP \textbf{$p<0.001$} & \cBP \textbf{$p<0.001$} \\
H6\; Strong-neighbor support & $p=0.209$              & $p=0.035$ \\
H7\; Same-category evidence  & $p=0.832$              & $p=0.725$ \\
\bottomrule
\end{tabular}
\end{table*}

\begin{figure*}[!htbp]
\centering
\includegraphics[width=\linewidth]{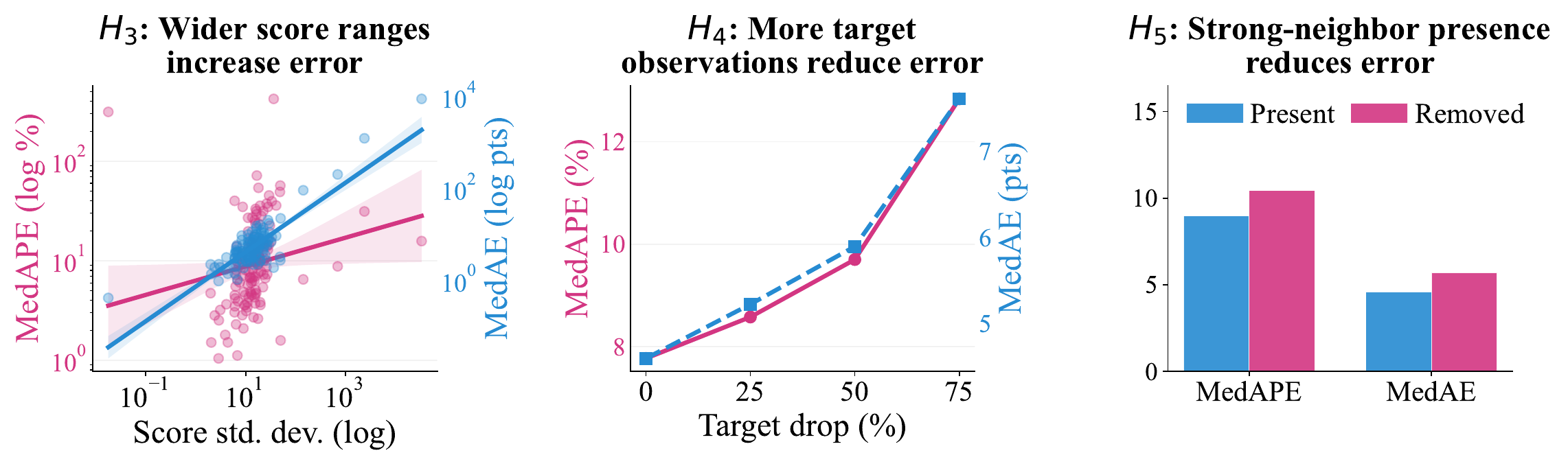}
\caption{\textbf{Benchmark-level prediction-error patterns.} Three benchmark-side factors jointly affect how hard a benchmark is to predict: a wider score spread across models makes prediction harder (H3), more observed model scores on the target benchmark makes it easier (H4), and having at least one strongly correlated neighbor benchmark in the training matrix makes it easier (H5).}
\label{fig:predictability_factors_51}
\end{figure*}

\Cref{tab:predictability_factors} reports the test results: three benchmark-side hypotheses are rejected under both error metrics (H3 score spread, H4 target coverage, and H5 strong-neighbor presence), and \Cref{fig:predictability_factors_51} visualizes the corresponding effects.

Three hypotheses are rejected under both metrics, i.e., these factors do affect prediction quality. Rejecting H3 (score spread) means benchmarks with wider score ranges across models are harder to predict. Rejecting H4 (target coverage) and H5 (strong-neighbor presence) means a benchmark is easier to predict when it has many observed model scores, and when at least one strongly correlated neighbor remains in the training matrix. The remaining hypotheses (H1 low-rank fit, H2 score level, H6 strong-neighbor support, H7 same-category evidence) are not rejected under both metrics; in particular, failing to reject H7 indicates that \benchpress{} benefits from observed correlations among benchmarks, not from category metadata.
The full $7 \times 2$ hypothesis $\times$ metric grid is reported in \Cref{app:predictability_factors_full}.

\paragraph{Model analysis.}
\label{sec:model_predict}
Symmetrically, what determines whether a model is easy or hard for \benchpress{} to predict?
Per-model prediction error varies by $\sim$40$\times$ across the 84 models in our matrix, so this matters in practice: harder-to-predict models warrant less trust in their point estimates.
For each model we run the same hide-half evaluation (10 random splits with the full Logit Bias ALS pipeline) and aggregate the held-out predictions into per-model $\mathsf{MedAPE}$ and $\mathsf{MedAE}$.

We test nine hypotheses split into three families.
H1--H4 probe model-intrinsic features (size, type, score level, low-rank fit), each evaluated by univariate Spearman correlation against per-model error (H1 uses the $n=25$ models with disclosed parameter counts; H2--H4 use all 84).
H5--H8 probe data availability and overlap with other models, each evaluated by a paired hide-half ablation that modifies the training matrix and measures the change in error.
H9 tests temporal generalization via a rolling simulation: train only on older models and predict newer ones.

\begin{figure*}[!htbp]
\centering
\includegraphics[width=\linewidth]{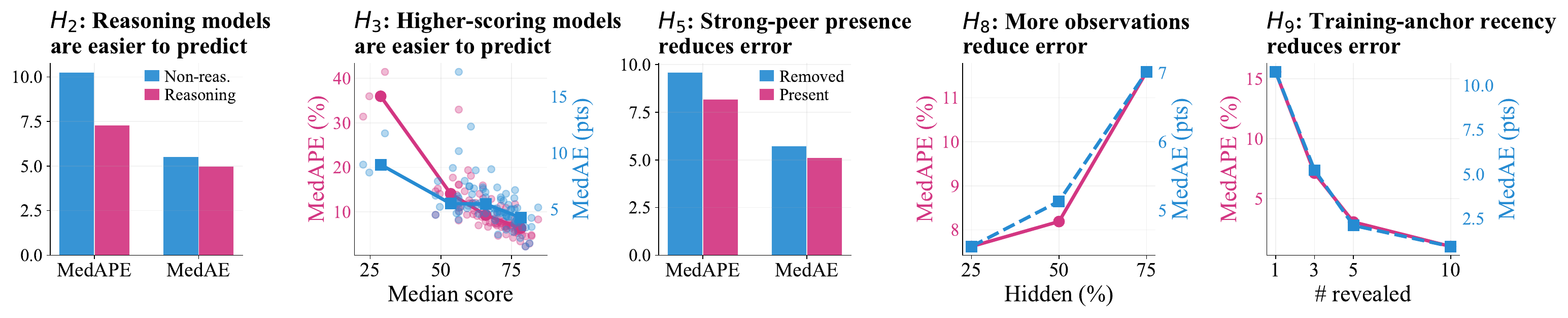}
\caption{\textbf{Representative model-level prediction-error patterns.} Five model-side factors jointly affect how hard a model is to predict: reasoning models are easier than non-reasoning ones (H2), higher-scoring models are easier than lower-scoring ones (H3), having at least one strongly correlated peer model in the training matrix makes prediction easier (H5), more observed benchmark scores on the target model makes prediction easier (H8), and a training matrix containing models recently released relative to the target makes prediction easier (H9).}
\label{fig:error_hypotheses_52}
\end{figure*}

\begin{itemize}[nosep,leftmargin=*]
\item \emph{H1 Model size.} \emph{Intuition:} larger models may behave more consistently, so they could be easier to predict from similar models. \emph{Hypothesis:} model size is not associated with how well a model can be predicted. \emph{Feature:} parameter count for the 25 models with disclosed sizes. \emph{Test:} Spearman.
\item \emph{H2 Model type.} \emph{Intuition:} reasoning and non-reasoning models score in noticeably different ways, so a model is easier to predict when others of its own kind are present. \emph{Hypothesis:} whether a model is a reasoning model is not associated with how well it can be predicted. \emph{Feature:} binary reasoning vs.\ non-reasoning indicator among models with annotated type. \emph{Test:} Spearman.
\item \emph{H3 Score level.} \emph{Intuition:} very strong or very weak models stand out from the pack, which could make them easier or harder to predict. \emph{Hypothesis:} the overall score level (i.e., capability) of a model is not associated with how well it can be predicted. \emph{Feature:} per-model median observed score. \emph{Test:} Spearman.
\item \emph{H4 Low-rank fit.} \emph{Intuition:} a model whose scores already follow \benchpress{}'s simple two-factor pattern is, by construction, one it can describe well. \emph{Hypothesis:} a model's row $R^2$ under the rank-2 SVD reconstruction is not associated with how well it can be predicted. \emph{Feature:} row-level $R^2$ under the rank-2 reconstruction of the standardized, zero-imputed score matrix. \emph{Test:} Spearman.
\item \emph{H5 Strong-peer presence.} \emph{Intuition:} if another model scores almost the same as the target across benchmarks, we can copy its scores; remove that peer and the copy is gone. \emph{Hypothesis:} masking strongly correlated peer models does not change a target model's prediction error. \emph{Intervention:} for each target model, mask all peer models whose Pearson correlation with the target is at least 0.95 on shared benchmarks, then rerun \benchpress{} on the target's hide-half cells. \emph{Test:} paired Wilcoxon.
\item \emph{H6 Strong-peer support.} \emph{Intuition:} even with such a peer, the prediction can only be as good as the number of benchmarks the two models were both scored on. \emph{Hypothesis:} reducing overlapping evidence from the strongest peer does not change a target model's prediction error. \emph{Intervention:} for each target model and hide-half split, identify the strongest peer model (highest $|r|$, requiring $|r|\ge0.95$), restrict to benchmarks observed by both the target and that peer, and drop nested prefixes $f \in \{0, 0.25, 0.5, 0.75\}$ of those overlapping peer cells before rerunning \benchpress{} on the target's held-out cells; we compare $f{=}0$ against $f{=}0.75$. \emph{Test:} paired Wilcoxon.
\item \emph{H7 Same-provider evidence.} \emph{Intuition:} models from one provider (e.g., all GPT versions) are often built alike and score alike, so removing them tests whether that family resemblance is what made the model predictable. \emph{Hypothesis:} masking same-provider variants does not change a target model's prediction error. \emph{Intervention:} mask all same-provider rows (e.g.\ all GPT variants when predicting a GPT model) and rerun \benchpress{} on the target's hide-half cells. \emph{Test:} paired Wilcoxon.
\item \emph{H8 Observation count.} \emph{Intuition:} the more of a model's own scores we have already seen, the more the predictor has to work from when filling the rest. \emph{Hypothesis:} reducing the amount of training evidence for a target model does not change its prediction error. \emph{Intervention:} compare the standard hide-half split against a more severe split that hides three quarters of each model's observed scores (\Cref{fig:error_hypotheses_52} shows the full trajectory across the evaluated hide fractions). \emph{Test:} paired Wilcoxon.
\item \emph{H9 Training-anchor recency.} \emph{Intuition:} a new model is easier to place when the models we train on include recent ones close to it in time and ability. \emph{Hypothesis:} the recency of the training matrix is not associated with how well newly released models can be predicted. \emph{Intervention:} sort all 84 models by release date and split into oldest, middle, and newest thirds; train the \benchpress{} score predictor using only the oldest third or only the middle third, reveal a small number of benchmark scores for each newest-third target model, and predict the rest. The table reports the condition with three revealed benchmarks. \emph{Test:} paired Wilcoxon.
\end{itemize}

\begin{table*}[!htbp]
\centering
\caption{\textbf{What makes a model easy or hard to predict?} H1--H4 use the Spearman rank correlation test; H5--H8 use paired Wilcoxon signed-rank tests on hide-half ablations; H9 compares older vs.\ more recent training data. The $p$-value is the probability of seeing an effect at least this large by chance if the listed (no-effect) hypothesis were true; smaller $p$ means stronger evidence against it. We reject a hypothesis when $p<0.01$ (bold), i.e., the data provide strong evidence that the factor does matter. Pink rows are rejected under both MedAPE and MedAE and are visualized in \Cref{fig:error_hypotheses_52}.}
\label{tab:model_hypotheses}
\vspace{0.3em}
\small
\renewcommand{\arraystretch}{1.18}
\setlength{\tabcolsep}{2.0pt}
\begin{tabular}{@{}>{\raggedright\arraybackslash}p{0.40\textwidth}>{\centering\arraybackslash}p{0.27\textwidth}>{\centering\arraybackslash}p{0.27\textwidth}@{}}
\toprule
Hypothesis & MedAPE $\downarrow$ ($p$-value) & MedAE $\downarrow$ ($p$-value) \\
\midrule
H1\; Model size ($\log_{10}$\,params)    & $p=0.101$             & $p=0.263$ \\
\cBP \textbf{H2\; Model type}                     & \cBP \textbf{$p<0.001$} & \cBP \textbf{$p=0.003$} \\
\cBP \textbf{H3\; Score level}                    & \cBP \textbf{$p<0.001$} & \cBP \textbf{$p<0.001$} \\
H4\; Low-rank fit                        & $p=0.389$             & $p=0.042$ \\
\midrule
\cBP \textbf{H5\; Strong-peer presence}           & \cBP \textbf{$p<0.001$} & \cBP \textbf{$p=0.004$} \\
H6\; Strong-peer support                 & $p=0.033$             & $p=0.309$ \\
H7\; Same-provider evidence              & $p=0.011$             & $p=0.094$ \\
\cBP \textbf{H8\; Observation count}              & \cBP \textbf{$p<0.001$} & \cBP \textbf{$p<0.001$} \\
\midrule
\cBP \textbf{H9\; Training-anchor recency}        & \cBP \textbf{$p=0.002$} & \cBP \textbf{$p=0.002$} \\
\bottomrule
\end{tabular}
\end{table*}

\Cref{tab:model_hypotheses} reports the full results, and \Cref{fig:error_hypotheses_52} visualizes the model-side hypotheses that are rejected under both error metrics: H2, H3, H5, H8, and H9.

Five hypotheses are rejected under both metrics, i.e., these factors do affect prediction quality. 
Reasoning models are easier to predict than non-reasoning ones (H2), and higher-scoring models are easier than lower-scoring ones (H3). Among the ablations, models with at least one strongly correlated peer in the matrix are easier to predict (H5), models with more observed benchmark scores are easier (H8), and the temporal experiment shows that prediction quality on newer models improves when the training matrix contains more recent anchors rather than only the oldest third (H9). 
The H5 result matches what \citet{zhang2025} find: predictions are only reliable when the target model performs similarly to the models the predictor was fit on.
The remaining hypotheses (H1 model size, H4 low-rank fit, H6 strong-peer support, H7 same-provider evidence) are not rejected under both metrics; in particular, failing to reject H7 indicates that \benchpress{} uses capability-profile similarity rather than provider identity.
The full $9 \times 2$ hypothesis $\times$ metric grid is reported in \Cref{app:model_predictability_factors_full}, and a per-model predictability ranking is given in \Cref{app:per_model_predictability}.

Because we run $32$ tests across both families ($16$ hypotheses under two metrics), we control the false discovery rate: under a Benjamini-Hochberg correction \citep{benjamini1995} at $\alpha=0.01$, all eight jointly-supported factors stay significant under both metrics (\Cref{app:reliability_stat_tests}).

\begin{finding}{A predicted score is most trustworthy when both sides of the matrix are well supported: the target benchmark has many observations and correlated neighbors, the target model has many observed scores and correlated peers, and the training matrix contains recent models near the target.}{prediction_error_factors}
\end{finding}

\subsection{Estimating Prediction Reliability}
\label{sec:confidence_calibration}


Point predictions alone are not enough for deciding when to skip benchmark runs.
If a predicted score will be used to decide whether to skip an expensive evaluation, the useful question is not only ``what score would this model get?'', but also ``how much should we trust this prediction?''
We therefore train reliability estimators for predictions from the default \benchpress{} score predictor in \Cref{sec:method_comparison}.
Each estimator assigns a predicted cell a risk score, where larger risk means the score prediction is less reliable.
We compare the reliability estimators by asking which one best identifies safe-to-use predictions before the benchmark is run.
The same risk score can also be calibrated into a trust probability and a conformal prediction interval; those details and interval-width results are reported in \Cref{app:confidence_calibration_details}.
Throughout this subsection we report MedAE only, since the risk-coverage curve is measured on the absolute score-point scale.

\paragraph{Reliability estimators.}
We compare three lightweight ways to compute this risk score, all sharing the same setup.
Each one is a small model trained on held-out cells from the training folds: given features about a cell, predict how far off the Logit Bias ALS point estimate will be on that cell.
At test time it sees only what is available \emph{before} running the benchmark, and outputs a larger risk for cells where the point prediction is likely to be less accurate.
The three methods differ only in which features they use.
\begin{itemize}[nosep,leftmargin=*]
\item \emph{Ensemble-spread reliability estimator} uses only disagreement among score predictors. For the same hidden cell, we collect the point predictions made by the Logit Bias ALS regularization settings around the selected one and by the strongest full-coverage methods from \Cref{sec:method_comparison}. The features are simple summaries of how much those predictions spread out: their standard deviation, median absolute deviation, central 80\% span, and distance between the selected Logit Bias ALS prediction and the median prediction. If many plausible predictors disagree, this estimator should assign higher risk.
\item \emph{Matrix-support reliability estimator} ignores predictor disagreement and instead reuses the model- and benchmark-side signals from \Cref{sec:reliability_analysis}, in the same order as the hypothesis tables. From the benchmark side: median score (H2), score spread (H3), observation count (H4), and strongest-neighbor correlation (H5). From the model side: median score (H3), strongest-peer correlation (H5), strongest-peer overlap (H6), and observation count (H8). Median score on either axis did not reach joint significance, but we keep it as a low-cost control. This estimator asks whether the cell is easy to infer from observed scores on correlated peer models and benchmarks.
\item \emph{Hybrid reliability estimator} uses both feature groups in one model. It can learn cases where predictor disagreement is enough to flag risk, cases where sparse structural support is enough to flag risk, and cases where both signals reinforce each other.
\end{itemize}

\begin{wrapfigure}{r}{0.35\textwidth}
\vspace{-0.8em}
\centering
\includegraphics[width=\linewidth]{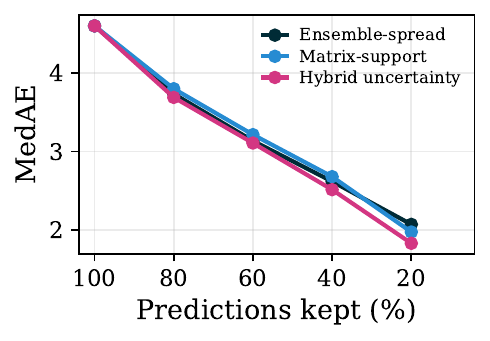}
\caption{\textbf{Reliability estimates identify safer predictions.} Lower curves mean safer subsets are identified more reliably; the hybrid estimator gives the cleanest ordering.}
\label{fig:confidence_calibration}
\vspace{-2.0em}
\end{wrapfigure}
All three reliability estimators use the same fold-internal model selection over a linear ridge baseline and small ReLU MLPs.
For each evaluated fold, we train candidate risk models on the other folds only, standardize features using that training split, select the risk-model architecture inside the training folds, and then predict risks for the held-out fold.
This keeps the reliability experiment honest: the risk score for a hidden cell is learned from other cells, not from its own error.
Full feature lists and training details for all three estimators are in \Cref{app:confidence_calibration_details}.

Once the reliability estimator outputs a risk score, the main-text evaluation treats it as a ranking signal: predictions with lower risk should have lower realized error.

\paragraph{Evaluation setting.}
All reliability estimators are evaluated on the same held-out folds as the point-prediction comparison.
For every hidden score, the reliability estimator may use the training matrix, the fixed Logit Bias ALS point prediction, and auxiliary quantities derived from the training fold, but not the held-out value.
We evaluate reliability ranking with a risk-coverage curve: sort cells from lowest to highest risk and plot MedAE after keeping the most trusted 100\%, 80\%, 60\%, 40\%, or 20\% of cells.
This asks whether the risk score can identify predictions that are safe enough to use for triage, while flagging predictions that should still trigger benchmark runs.

\paragraph{Results.}
When keeping only the most-trusted 20\% of cells, the hybrid estimator lowers selective MedAE to 1.83 score points, beating both single-feature variants (\Cref{fig:confidence_calibration}); at 40\% and 60\% kept, its MedAE is 2.51 and 3.10.
We therefore use the \textbf{hybrid reliability estimator} as the reliability layer: it takes both ensemble spread and matrix support into account, and assigns each prediction a risk score that identifies whether the prediction is safe to use.

\begin{finding}{A hybrid reliability estimator uses ensemble spread and matrix support to identify low-risk score predictions before running the benchmark.}{confidence_calibration}
\end{finding}

\section{Discussion}
\label{sec:discussion}

This paper shows that the public benchmark landscape has enough shared structure to support score prediction at scale.
Starting from an $84 \times 133$ public score matrix, we find that its dominant variation is effectively rank-2 and build \benchpress{}, a matrix-completion predictor for missing model--benchmark scores.
With this predictor fixed, we show that a small probe set can recover much of a model's scorecard, that predicted scores preserve most meaningful same-benchmark rankings, and that a few seed evaluations can anchor predictions for models in a pre-specified temporal window.
Finally, the reliability analysis identifies when those point predictions are well supported by the observed matrix and when the benchmark should still be run.

\paragraph{Limitations and future work.}
We close with six pairings of a current limitation and the most natural extension it suggests.
\emph{First}, \benchpress{} cannot reliably predict a candidate model whose capability profile lacks a close neighbor in the matrix; incorporating external signals about the model (training-data composition, architecture, model size, and other published metadata) and computing model-to-model similarity from these features could anchor outliers even before any benchmark scores are observed.
\emph{Second}, benchmark-level predictions are only as good as the benchmarks themselves: a noisy or poorly constructed benchmark is faithfully predicted as such; pushing score prediction beyond aggregated benchmark scores to instance-level outcomes would let \benchpress{} capture within-benchmark structure and improve predictions on the hardest tails.
\emph{Third}, our matrix already covers mainstream text and vision-language benchmarks, but more specialized ecosystems (audio and speech, robotics and embodied agents, scientific simulators) remain untested; whether the same low-rank treatment carries over to these settings is an open question.
\emph{Fourth}, the rank-2 geometry is a property of the current snapshot rather than a guarantee for future releases; as the matrix grows, tracking whether the rank stays at two, or whether a third latent factor emerges, will determine the long-term viability of this approach and signal when a refresh of the score-prediction recipe is warranted.
\emph{Fifth}, the factors we test for predictability are a plausible but not exhaustive set, and the reported associations are correlational rather than causal: attributes we did not encode, such as architecture family or training recipe, could also shape how predictable a model is, and a broader, more causal feature search is a natural next step.
\emph{Sixth}, our reliability layer is a practical add-on rather than a core component: we estimate per-cell risk with a small MLP, but simpler statistical estimators could fill the same role, and sharper, better-grounded uncertainty quantification remains open.

\bibliographystyle{plainnat}
\bibliography{references}

\appendix
\addcontentsline{toc}{section}{Appendix}
\addtocontents{toc}{\protect\setcounter{tocdepth}{-1}}


\newpage
\clearpage

\phantomsection

{\LARGE\bfseries Appendix}\par\medskip

\noindent The appendix is organized as supplements to the main text.
\Cref{app:intro} supplements \Cref{sec:intro} with the experiment setting for \Cref{fig:hero}.
\Cref{app:data} supplements \Cref{sec:predictability} with data-collection provenance, full benchmark and model catalogs, and additional evidence for the low-rank structure.
\Cref{app:methods} supplements \Cref{sec:bp_methods} with the full candidate-method specification, comparisons to observational scaling laws and independent score matrices, and the additional LLM baseline prompt template.
\Cref{app:evaluation_design} supplements \Cref{sec:findings} with budgeted scorecard recovery, ranking preservation, and temporal-deployment details.
\Cref{app:trust} supplements \Cref{sec:trust} with prediction-error and prediction-reliability details.

\medskip
{\noindent\bfseries Contents}\par\smallskip
\begingroup
\setlength{\parskip}{0pt}
\setlength{\parindent}{0pt}
\renewcommand{\baselinestretch}{1.1}\selectfont
\hyperref[app:intro]{\textbf{A}}\quad Supplemental to \Cref{sec:intro}: \hyperref[app:intro]{Introduction} \dotfill \pageref{app:intro}\par
\hspace{2em}\hyperref[app:hero_figure]{A.1\quad Experiment Setting for \Cref{fig:hero}} \dotfill \pageref{app:hero_figure}\par
\hspace{2em}\hyperref[app:updated_score_matrix_2026_08_26]{A.2\quad Updated Score Matrix (Version 2, August 26, 2026)} \dotfill \pageref{app:updated_score_matrix_2026_08_26}\par
\smallskip
\hyperref[app:data]{\textbf{B}}\quad Supplemental to \Cref{sec:predictability}: \hyperref[app:data]{The Score Matrix and Its Geometry} \dotfill \pageref{app:data}\par
\hspace{2em}\hyperref[app:data_collection]{B.1\quad Data Collection} \dotfill \pageref{app:data_collection}\par
\hspace{2em}\hyperref[app:benchmark_score_matrix]{B.2\quad The Final Score Matrix} \dotfill \pageref{app:benchmark_score_matrix}\par
\hspace{2em}\hyperref[app:rank_geometry]{B.3\quad Rank-2 Geometry} \dotfill \pageref{app:rank_geometry}\par
\smallskip
\hyperref[app:methods]{\textbf{C}}\quad Supplemental to \Cref{sec:bp_methods}: \hyperref[app:methods]{\benchpress{}: A Low-rank Benchmark Score Predictor} \dotfill \pageref{app:methods}\par
\hspace{2em}\hyperref[app:method_details]{C.1\quad Candidate Methods} \dotfill \pageref{app:method_details}\par
\hspace{2em}\hyperref[app:method_comparison]{C.2\quad From Candidate Methods to \benchpress{}} \dotfill \pageref{app:method_comparison}\par
\hspace{2em}\hyperref[app:llm_five_shot_prompt]{C.3\quad \benchpress{} vs. LLMs as Benchmark Score Predictors} \dotfill \pageref{app:llm_five_shot_prompt}\par
\hspace{2em}\hyperref[app:osl_comparison]{C.4\quad Comparison to Observational Scaling Laws} \dotfill \pageref{app:osl_comparison}\par
\hspace{2em}\hyperref[app:other_matrices]{C.5\quad \benchpress{} on Other Score Matrices} \dotfill \pageref{app:other_matrices}\par
\smallskip
\hyperref[app:evaluation_design]{\textbf{D}}\quad Supplemental to \Cref{sec:findings}: \hyperref[app:evaluation_design]{What \benchpress{} Enables for Model Evaluation} \dotfill \pageref{app:evaluation_design}\par
\hspace{2em}\hyperref[app:probe_selection]{D.1\quad Budgeted Scorecard Recovery} \dotfill \pageref{app:probe_selection}\par
\hspace{2em}\hyperref[app:ranking_preservation]{D.2\quad Preserving Model Rankings} \dotfill \pageref{app:ranking_preservation}\par
\hspace{2em}\hyperref[app:temporal_deployment]{D.3\quad Predicting Newly Released Models} \dotfill \pageref{app:temporal_deployment}\par
\smallskip
\hyperref[app:trust]{\textbf{E}}\quad Supplemental to \Cref{sec:trust}: \hyperref[app:trust]{When to Trust \benchpress{}'s Predictions} \dotfill \pageref{app:trust}\par
\hspace{2em}\hyperref[app:reliability_analysis]{E.1\quad What Affects Prediction Reliability} \dotfill \pageref{app:reliability_analysis}\par
\hspace{2em}\hyperref[app:confidence_calibration_details]{E.2\quad Estimating Prediction Reliability} \dotfill \pageref{app:confidence_calibration_details}\par
\endgroup

\newpage

\section{Supplemental to \texorpdfstring{\Cref{sec:intro}}{Section 1}: Introduction}
\label{app:intro}

\subsection{Experiment Setting for \texorpdfstring{\Cref{fig:hero}}{Figure 1}}
\label{app:hero_figure}

\paragraph{Left panel (per-cell error).}
We pick four highlighted cells: (Claude Opus 4.7, SWE-bench Verified), (GPT-5.5, Terminal-Bench), (Gemini 3.1 Pro, LiveCodeBench), and (DeepSeek-V4-Pro, HLE Text). For each cell and each $k \in \{1, \dots, 10\}$, we (i) hide the target cell, (ii) sample $k$ scores uniformly at random from the same model's other observed cells, (iii) feed the resulting masked matrix to \benchpress{}, and (iv) record the absolute error on the held-out target cell. We repeat over $10$ seeds (base seed $42$); the line is the per-cell median and the shaded band is the 25--75 percentile range. Whenever the target cell itself appears in the revealed prefix the error drops to zero. The diamond at $k{=}0$ marks the benchmark-median baseline (no \benchpress{}).

\paragraph{Right panel (overall pooled error).}
The right panel reuses the global probe-set setting of \Cref{sec:probe_selection,app:probe_selection}: a fixed probe set of $k$ benchmarks is chosen, every model is evaluated on whichever probe scores it has observed, and \benchpress{} predicts the rest of each model's observed cells. Pooled MedAE is reported across all evaluated cells. The greedy curves use the cost-aware and cost-unaware MedAE orderings from \Cref{tab:probe_sets}; the gray random baseline draws one global random benchmark ordering per seed ($10$ seeds, base seed $42$) and uses the corresponding prefix for every target model. The shaded band is the 25--75 percentile range across seeds.


\subsection{Updated Score Matrix (Version 2, August 26, 2026)}
\label{app:updated_score_matrix_2026_08_26}

We identify each released version of our living score matrix by its snapshot date. The August 26, 2026 score matrix contains 283 models, 712 benchmarks, and 8{,}713 primary scores, together with 2{,}756 additional source-specific score candidates retained alongside the primary records. Requiring at least 15 observed benchmarks per model and eight observed models per benchmark yields a $129 \times 253$ matrix with 4{,}905 observed cells (15.0\% filled), compared with the $84 \times 133$ matrix with 2{,}604 observed cells used for the paper's May 2026 experiments. We rerun the complete method comparison on this updated matrix in \Cref{app:other_matrices}.

\section{Supplemental to \texorpdfstring{\Cref{sec:predictability}}{Section 3}: The Score Matrix and Its Geometry}
\label{app:data}

\subsection{Data Collection}
\label{app:data_collection}

\Cref{sec:bp_collection} describes how we crawl public model releases, technical reports, model cards, and primary leaderboards, then canonicalize and filter the resulting raw matrix.
Here we make explicit what information is preserved in the released data, because later analyses reuse these fields without re-crawling the original sources.
The released data keeps three linked record types: one record per model, one record per benchmark, and one record per observed model--benchmark score.
\begin{titlebox}[title={Example model record}, fontupper=\footnotesize\ttfamily]
\{\\
\quad "id": "gpt-5.2",\\
\quad "name": "GPT-5.2",\\
\quad "provider": "OpenAI",\\
\quad "release\_date": "2025-12-11",\\
\quad "is\_reasoning": true,\\
\quad "open\_weights": false,\\
\quad "canonical\_setting": \{\\
\quad\quad "mode": "thinking",\\
\quad\quad "effort": "xhigh",\\
\quad\quad "tools": "none",\\
\quad\quad "sampling": "pass@1",\\
\quad\quad "judge": "rule-based",\\
\quad\quad "harness": "official",\\
\quad\quad "prompt\_style": "default"\\
\quad \}\\
\}
\end{titlebox}

\begin{titlebox}[title={Example benchmark record}, fontupper=\footnotesize\ttfamily]
\{\\
\quad "id": "aime\_2025",\\
\quad "name": "AIME 2025",\\
\quad "category": "Math",\\
\quad "metric": "\% correct (pass@1)",\\
\quad "num\_problems": 30,\\
\quad "source\_url": "https://artofproblemsolving.com/wiki/index.php/2025\_AIME",\\
\quad "canonical\_setting": \{\\
\quad\quad "version": "AIME-2025-I+II",\\
\quad\quad "metric\_type": "pct",\\
\quad\quad "range": [0, 100],\\
\quad\quad "higher\_is\_better": true,\\
\quad\quad "multimodal\_input": false,\\
\quad\quad "tools": "none"\\
\quad \}\\
\}
\end{titlebox}

\begin{titlebox}[title={Example observed cell record}, fontupper=\footnotesize\ttfamily]
\{\\
\quad "model\_id": "gpt-5.2",\\
\quad "benchmark\_id": "aime\_2025",\\
\quad "score": 100.0,\\
\quad "reference\_url": "https://openai.com/index/introducing-gpt-5-2/",\\
\quad "source\_type": "official\_blog",\\
\quad "audit\_status": "verified",\\
\quad "reported\_setting": \{\\
\quad\quad "mode": "thinking",\\
\quad\quad "effort": "xhigh",\\
\quad\quad "tools": "none",\\
\quad\quad "sampling": "pass@1",\\
\quad\quad "judge": "rule-based",\\
\quad\quad "harness": "official"\\
\quad \},\\
\quad "matches\_canonical": true,\\
\quad "candidates": [\{"score": 99.0, "source\_type": "model\_card"\}, ...]\\
\}
\end{titlebox}

This structure separates entity metadata from score provenance.
Benchmark-level fields, such as item count and modality, support analyses that reason about the columns of the matrix.
Model-level fields, such as provider, release date, and reasoning capability, support analyses that reason about the rows.
Cell-level fields keep the audit trail for the actual number used in the matrix: where it came from, how the model was run, whether that setting matches the canonical setting, and which alternative values were seen but not selected as the primary score.

\paragraph{Filtering-threshold robustness.}
The density filter trades matrix coverage for a more fully observed analysis block.
To test whether the adopted minimum of 15 benchmark observations per model and 8 model observations per benchmark was selected because it yields favorable prediction error, we apply the fixed \benchpress{} predictor to the complete threshold grid under the same 10-seed, three-fold per-model holdout protocol.
Each row in \Cref{tab:threshold_error_sweep} is evaluated on the observed cells retained by that threshold setting.
Prediction error remains comparable across the grid, while several stricter filters yield lower error by discarding more sparsely observed rows and columns.
The adopted setting therefore balances matrix coverage and density rather than minimizing reported error.

\begin{table}[!htbp]
\centering
\caption{\textbf{Prediction error remains comparable across filtering thresholds.}
Each row applies the fixed logit-space rank-2 Bias ALS predictor ($\lambda=0.1$) and reports median error under 10 seeds and three per-model folds on that row's retained observed cells.
``Canonicalized, no density filter'' is evaluated after model and benchmark variants are consolidated but before minimum-observation filtering; bold marks the adopted setting.}
\label{tab:threshold_error_sweep}
\small
\setlength{\tabcolsep}{3pt}
\begin{tabular}{ccrrrr}
\toprule
\multicolumn{2}{c}{Min.\ obs.\ per} & \#Obs. & Fill & MedAE & MedAPE \\
\cmidrule(lr){1-2}
Model & Bench. & & & $\downarrow$ & $\downarrow$ \\
\midrule
\multicolumn{2}{c}{Canonicalized, no density filter} & 4{,}177 & 7.6\% & 4.92 & 8.44\% \\
\midrule
10 & 8 & 3{,}201 & 17.5\% & 4.64 & 7.67\% \\
10 & 12 & 2{,}795 & 21.7\% & 4.52 & 7.52\% \\
10 & 16 & 2{,}246 & 28.6\% & 4.79 & 7.79\% \\
\cBP\textbf{15} & \cBP\textbf{8} & \cBP\textbf{2{,}604} & \cBP\textbf{23.3\%} & \cBP\textbf{4.63} & \cBP\textbf{7.77\%} \\
15 & 12 & 1{,}811 & 36.7\% & 3.92 & 6.59\% \\
15 & 16 & 1{,}337 & 47.6\% & 3.83 & 6.23\% \\
20 & 8 & 1{,}885 & 35.0\% & 4.05 & 7.16\% \\
20 & 12 & 1{,}353 & 47.8\% & 3.91 & 6.68\% \\
20 & 16 & \multicolumn{4}{c}{Empty after filtering} \\
\bottomrule
\end{tabular}
\end{table}
\FloatBarrier

\subsection{The Final Score Matrix}
\label{app:benchmark_score_matrix}

\Cref{sec:bp_data} introduced the benchmark score matrix as an $84 \times 133$ table with $s_{mb}$ the score of model $m$ on benchmark $b$, populated from publicly available evaluations and 23.3\% filled.
This appendix provides the complete benchmark and model inventories underlying that matrix.
\Cref{tab:benchmarks} lists all 133 benchmarks with their categories, metrics, item counts, and source links.
\Cref{tab:models} enumerates all 84 retained models with parameter counts, reasoning capability, open-weight status, release dates, and source links.
Every score in the matrix is attributed to one of these sources; the full (model, benchmark) $\to$ URL mapping is released with the accompanying repository.

{\scriptsize
\setlength{\tabcolsep}{1.2pt}
\begin{longtable}{p{0.13\textwidth} p{0.40\textwidth} p{0.19\textwidth} r c}
\caption{\textbf{Benchmark inventory.} All 133 benchmarks in the adopted score matrix. Categories are grouped to match the main-text summary.}
\label{tab:benchmarks} \\
\toprule
\textbf{Category} & \textbf{Benchmark} & \textbf{Metric} & \textbf{Items} & \textbf{Link} \\
\midrule
\endfirsthead
\toprule
\textbf{Category} & \textbf{Benchmark} & \textbf{Metric} & \textbf{Items} & \textbf{Link} \\
\midrule
\endhead
\bottomrule
\endfoot
\multirow{26}{*}{\parbox[t]{0.13\textwidth}{Agentic \& tool use (26)}} & BFCL & --- & --- & \href{https://cohere.com/research/papers/command-a-technical-report.pdf}{\faExternalLink*} \\
 & BFCL v3 & --- & --- & \href{https://gorilla.cs.berkeley.edu/leaderboard.html}{\faExternalLink*} \\
 & BrowseComp & \% correct & 1{,}266 & \href{https://openai.com/index/browsecomp/}{\faExternalLink*} \\
 & BrowseComp-ZH & --- & 1{,}156 & \href{https://github.com/PALIN2018/BrowseComp-ZH}{\faExternalLink*} \\
 & ComplexFuncBench & \% & 1{,}000 & \href{https://github.com/THUDM/ComplexFuncBench}{\faExternalLink*} \\
 & CyberGym & \% solved & 1{,}507 & \href{https://www.cybergym.io/}{\faExternalLink*} \\
 & DeepSearchQA (Accuracy) & \% & 900 & \href{https://huggingface.co/datasets/google/deepsearchqa}{\faExternalLink*} \\
 & Finance Agent v1.1 & \% solved & 537 & \href{https://arxiv.org/abs/2508.00828}{\faExternalLink*} \\
 & FinSearchComp-Global & \% & 317 & \href{https://arxiv.org/abs/2509.13160}{\faExternalLink*} \\
 & Frames & \% & 824 & \href{https://arxiv.org/abs/2409.12941}{\faExternalLink*} \\
 & GAIA (text only) & \% & 103 & \href{https://arxiv.org/abs/2509.06501}{\faExternalLink*} \\
 & MCPAtlas Public & \% correct (pass@1) & 500 & \href{https://huggingface.co/datasets/ScaleAI/MCP-Atlas}{\faExternalLink*} \\
 & MCPMark & \% success (pass@1) & 127 & \href{https://github.com/eval-sys/mcpmark}{\faExternalLink*} \\
 & OSWorld & \% success & 369 & \href{https://os-world.github.io/}{\faExternalLink*} \\
 & tau-bench Airline & \% success & 50 & \href{https://arxiv.org/abs/2406.12045}{\faExternalLink*} \\
 & Tau-Bench Retail & \% success & 115 & \href{https://arxiv.org/abs/2406.12045}{\faExternalLink*} \\
 & Terminal-Bench 1.0 & \% solved & --- & \href{https://terminal-bench.com/}{\faExternalLink*} \\
 & Terminal-Bench 2.0 & \% solved & --- & \href{https://www.tbench.ai/leaderboard/terminal-bench/2.0}{\faExternalLink*} \\
 & Toolathlon & \% correct (pass@1) & 108 & \href{https://toolathlon.github.io/}{\faExternalLink*} \\
 & Vending-Bench 2 & --- & 15{,}000 & \href{https://andonlabs.com/evals/vending-bench-2}{\faExternalLink*} \\
 & WideSearch (item-F1) & \% & 200 & \href{https://huggingface.co/datasets/ByteDance-Seed/WideSearch}{\faExternalLink*} \\
 & xbench-DeepSearch & \% & 100 & \href{https://huggingface.co/datasets/xbench/DeepSearch}{\faExternalLink*} \\
 & $\tau^2$-bench Airline & \% success & 50 & \href{https://arxiv.org/abs/2506.07982}{\faExternalLink*} \\
 & $\tau^2$-bench Retail & \% success & 115 & \href{https://arxiv.org/abs/2506.07982}{\faExternalLink*} \\
 & $\tau^2$-bench Telecom & \% success & 114 & \href{https://arxiv.org/abs/2506.07982}{\faExternalLink*} \\
 & $\tau^3$-Bench & \% & 1{,}500 & \href{https://z.ai/blog/glm-5.1}{\faExternalLink*} \\
\cmidrule(lr){1-5}
\multirow{23}{*}{\parbox[t]{0.13\textwidth}{Math (23)}} & AIME 2024 & \% correct (pass@1) & 30 & \href{https://artofproblemsolving.com/wiki/index.php/2024_AIME}{\faExternalLink*} \\
 & AIME 2025 & \% correct (pass@1) & 30 & \href{https://artofproblemsolving.com/wiki/index.php/2025_AIME}{\faExternalLink*} \\
 & AIME 2026 & \% correct (pass@1) & 30 & \href{https://huggingface.co/datasets/MathArena/aime_2026_I}{\faExternalLink*} \\
 & Beyond AIME & \% & 100 & \href{https://huggingface.co/datasets/ByteDance-Seed/BeyondAIME}{\faExternalLink*} \\
 & BRUMO 2025 & \% correct (pass@1) & 30 & \href{https://huggingface.co/datasets/MathArena/brumo_2025}{\faExternalLink*} \\
 & CMIMC 2025 & \% correct (pass@1) & 40 & \href{https://huggingface.co/datasets/MathArena/cmimc_2025}{\faExternalLink*} \\
 & CNMO 2024 & \% & 6 & \href{https://www.cms.org.cn/Home/comp/comp_details/id/1253.html}{\faExternalLink*} \\
 & FrontierMath & \% correct T1-3 & 300 & \href{https://epoch.ai/benchmarks/frontiermath}{\faExternalLink*} \\
 & FrontierMath Tier 4 & \% & 48 & \href{https://epoch.ai/benchmarks/frontiermath}{\faExternalLink*} \\
 & GSM8K & \% correct & 1{,}319 & \href{https://arxiv.org/abs/2110.14168}{\faExternalLink*} \\
 & HMMT Feb 2025 & \% & 30 & \href{https://huggingface.co/datasets/MathArena/hmmt_feb_2025}{\faExternalLink*} \\
 & HMMT Feb 2026 & \% correct (pass@1) & 33 & \href{https://huggingface.co/datasets/MathArena/hmmt_feb_2026}{\faExternalLink*} \\
 & HMMT Nov 2025 & \% correct & 30 & \href{https://huggingface.co/datasets/MathArena/hmmt_nov_2025}{\faExternalLink*} \\
 & IMO-AnswerBench & --- & 400 & \href{https://imobench.github.io/}{\faExternalLink*} \\
 & MATH & --- & 12{,}500 & \href{https://github.com/hendrycks/math}{\faExternalLink*} \\
 & MATH-500 & \% correct & 500 & \href{https://arxiv.org/abs/2103.03874}{\faExternalLink*} \\
 & MathArena Apex 2025 & \% correct & 12 & \href{https://matharena.ai/apex/}{\faExternalLink*} \\
 & MathVision & \% correct & 3{,}040 & \href{https://huggingface.co/datasets/MathLLMs/MathVision}{\faExternalLink*} \\
 & MathVista & \% & 1{,}000 & \href{https://huggingface.co/datasets/AI4Math/MathVista}{\faExternalLink*} \\
 & MGSM & exact match (\%) & 2{,}500 & \href{https://github.com/google-research/url-nlp/tree/main/mgsm}{\faExternalLink*} \\
 & MT-AIME2024 & \% & 1{,}650 & \href{https://huggingface.co/datasets/amphora/MCLM}{\faExternalLink*} \\
 & SMT 2025 & \% correct (pass@1) & 53 & \href{https://huggingface.co/datasets/MathArena/smt_2025}{\faExternalLink*} \\
 & USAMO 2025 & \% of 42 points & 6 & \href{https://huggingface.co/datasets/MathArena/usamo_2025}{\faExternalLink*} \\
\cmidrule(lr){1-5}
\multirow{21}{*}{\parbox[t]{0.13\textwidth}{Coding (21)}} & Aider Polyglot (diff mode) & \% & 450 & \href{https://aider.chat/2024/12/21/polyglot.html}{\faExternalLink*} \\
 & Aider Polyglot (whole mode) & \% & 450 & \href{https://aider.chat/2024/12/21/polyglot.html}{\faExternalLink*} \\
 & ArtifactsBench & \% & 5{,}475 & \href{https://github.com/Tencent-Hunyuan/ArtifactsBenchmark}{\faExternalLink*} \\
 & BigCodeBench & pass@1 \% & 1{,}140 & \href{https://bigcode-bench.github.io/}{\faExternalLink*} \\
 & Bird-SQL (Dev) & --- & --- & \href{https://bird-bench.github.io/}{\faExternalLink*} \\
 & Codeforces Rating & Elo rating & --- & \href{https://codeforces.com/}{\faExternalLink*} \\
 & HumanEval & pass@1 \% & 164 & \href{https://github.com/openai/human-eval}{\faExternalLink*} \\
 & LiveCodeBench & pass@1 \% & 1{,}055 & \href{https://livecodebench.github.io/}{\faExternalLink*} \\
 & MBPP+ & --- & --- & \href{https://cohere.com/research/papers/command-a-technical-report.pdf}{\faExternalLink*} \\
 & Multi-SWE-bench & \% & 1{,}632 & \href{https://huggingface.co/datasets/ByteDance-Seed/Multi-SWE-bench}{\faExternalLink*} \\
 & MultiPL-E (average) & \% & 12{,}667 & \href{https://huggingface.co/datasets/nuprl/MultiPL-E}{\faExternalLink*} \\
 & NL2Repo-Bench & \% & 104 & \href{https://arxiv.org/abs/2512.12730}{\faExternalLink*} \\
 & OJBench & \% & 232 & \href{https://arxiv.org/abs/2506.16395}{\faExternalLink*} \\
 & RepoQA & --- & 500 & \href{https://arxiv.org/abs/2406.06025}{\faExternalLink*} \\
 & SciCode & \% correct & 338 & \href{https://scicode-bench.github.io/}{\faExternalLink*} \\
 & SWE-bench Multilingual & \% resolved & --- & \href{https://www.swebench.com/}{\faExternalLink*} \\
 & SWE-bench Pro & \% resolved & 731 & \href{https://scale.com/leaderboard/swe_bench_pro_public}{\faExternalLink*} \\
 & SWE-bench Verified & \% resolved & 500 & \href{https://www.swebench.com/}{\faExternalLink*} \\
 & SWE-Lancer IC Diamond & \% & 198 & \href{https://github.com/openai/frontier-evals/tree/main/project/swelancer}{\faExternalLink*} \\
 & SWE-Lancer IC SWE Diamond Freelance (\$) & dollars & 198 & \href{https://github.com/openai/frontier-evals/tree/main/project/swelancer}{\faExternalLink*} \\
 & Terminal-Bench Hard & \% & --- & \href{https://z.ai/blog/glm-4.7}{\faExternalLink*} \\
\cmidrule(lr){1-5}
\multirow{12}{*}{\parbox[t]{0.13\textwidth}{Multimodal \& vision (12)}} & BabyVision & \% accuracy & 388 & \href{https://huggingface.co/datasets/UnipatAI/BabyVision}{\faExternalLink*} \\
 & CharXiv Descriptive & \% accuracy & 4{,}000 & \href{https://charxiv.github.io/}{\faExternalLink*} \\
 & CharXiv Reasoning & \% accuracy & 1{,}000 & \href{https://charxiv.github.io/}{\faExternalLink*} \\
 & ERQA & \% & 400 & \href{https://github.com/embodiedreasoning/ERQA}{\faExternalLink*} \\
 & MMMU & \% correct & 900 & \href{https://mmmu-benchmark.github.io/}{\faExternalLink*} \\
 & MMMU-Pro & \% correct & 3{,}460 & \href{https://huggingface.co/datasets/MMMU/MMMU_Pro}{\faExternalLink*} \\
 & OmniDocBench (normalized edit distance, lower is better) & edit distance (lower=better) & 1{,}651 & \href{https://huggingface.co/datasets/opendatalab/OmniDocBench}{\faExternalLink*} \\
 & OmniDocBench 1.5 & edit distance (lower=better) & 1{,}355 & \href{https://github.com/opendatalab/OmniDocBench}{\faExternalLink*} \\
 & ScreenSpot-Pro & --- & 1{,}581 & \href{https://github.com/likaixin2000/ScreenSpot-Pro-GUI-Grounding}{\faExternalLink*} \\
 & Vibe-Eval & --- & 269 & \href{https://github.com/reka-ai/reka-vibe-eval}{\faExternalLink*} \\
 & Video-MME & --- & 2{,}700 & \href{https://video-mme.github.io/}{\faExternalLink*} \\
 & Video-MMMU & \% & 900 & \href{https://videommmu.github.io/}{\faExternalLink*} \\
\cmidrule(lr){1-5}
\multirow{9}{*}{\parbox[t]{0.13\textwidth}{Long context (9)}} & AA Long Context Reasoning & \% correct & 300 & \href{https://artificialanalysis.ai/methodology/intelligence-benchmarking}{\faExternalLink*} \\
 & BrowseComp Long Context 128k & \% accuracy & 1{,}266 & \href{https://openai.com/index/gpt-5-1-for-developers/}{\faExternalLink*} \\
 & GraphWalks BFS 0-128K & \% & 300 & \href{https://huggingface.co/datasets/openai/graphwalks}{\faExternalLink*} \\
 & GraphWalks parents 0-128K & \% & 350 & \href{https://huggingface.co/datasets/openai/graphwalks}{\faExternalLink*} \\
 & LongBench-V2 & \% & 503 & \href{https://huggingface.co/datasets/THUDM/LongBench-v2}{\faExternalLink*} \\
 & MRCR v1 & --- & 2{,}000 & \href{https://storage.googleapis.com/deepmind-media/gemini/gemini_v1_5_report.pdf}{\faExternalLink*} \\
 & MRCR v2 & \% correct & 2{,}400 & \href{https://huggingface.co/datasets/openai/mrcr}{\faExternalLink*} \\
 & OpenAI MRCR v2 (2 needle, 128k) & \% & 500 & \href{https://huggingface.co/datasets/openai/mrcr}{\faExternalLink*} \\
 & OpenAI MRCR v2 (8-needle) & \% & 800 & \href{https://huggingface.co/datasets/openai/mrcr}{\faExternalLink*} \\
\cmidrule(lr){1-5}
\multirow{9}{*}{\parbox[t]{0.13\textwidth}{Instruction following (9)}} & Arena-Hard Auto & \% win rate & 500 & \href{https://lmarena.ai/}{\faExternalLink*} \\
 & COLLIE & \% & 2{,}080 & \href{https://arxiv.org/abs/2307.08689}{\faExternalLink*} \\
 & IFBench & \% correct & 300 & \href{https://github.com/allenai/IFBench}{\faExternalLink*} \\
 & IFEval & \% correct (prompt strict) & 541 & \href{https://arxiv.org/abs/2311.07911}{\faExternalLink*} \\
 & InFoBench & --- & 2{,}250 & \href{https://github.com/qinyiwei/InfoBench}{\faExternalLink*} \\
 & Internal API IF Hard & \% & --- & \href{https://openai.com/index/introducing-gpt-5-for-developers/}{\faExternalLink*} \\
 & Multi-IF & \% & 13{,}503 & \href{https://huggingface.co/datasets/facebook/Multi-IF}{\faExternalLink*} \\
 & MultiChallenge & \% & 273 & \href{https://github.com/ekwinox117/multi-challenge}{\faExternalLink*} \\
 & MultiChallenge (o3-mini grader) & \% & 273 & \href{https://github.com/ekwinox117/multi-challenge}{\faExternalLink*} \\
\cmidrule(lr){1-5}
\multirow{9}{*}{\parbox[t]{0.13\textwidth}{Knowledge \& QA (9)}} & C-Eval (Chinese) & \% & 12{,}342 & \href{https://huggingface.co/datasets/ceval/ceval-exam}{\faExternalLink*} \\
 & Chinese-SimpleQA & \% & 3{,}000 & \href{https://huggingface.co/datasets/OpenStellarTeam/Chinese-SimpleQA}{\faExternalLink*} \\
 & GDPval (Artificial Analysis ELO) & score & 220 & \href{https://huggingface.co/datasets/openai/gdpval}{\faExternalLink*} \\
 & HealthBench & \% & 5{,}000 & \href{https://huggingface.co/datasets/openai/healthbench}{\faExternalLink*} \\
 & MMLU-Pro & \% correct & 12{,}032 & \href{https://arxiv.org/abs/2406.01574}{\faExternalLink*} \\
 & MMMLU & \% correct & 258{,}090 & \href{https://huggingface.co/datasets/openai/MMMLU}{\faExternalLink*} \\
 & PopQA & --- & 14{,}267 & \href{https://huggingface.co/datasets/akariasai/PopQA}{\faExternalLink*} \\
 & SimpleQA & \% correct & 4{,}326 & \href{https://openai.com/index/introducing-simpleqa/}{\faExternalLink*} \\
 & SimpleQA-Verified & \% correct (pass@1) & --- & \href{https://huggingface.co/deepseek-ai/DeepSeek-V4-Pro}{\faExternalLink*} \\
\cmidrule(lr){1-5}
\multirow{8}{*}{\parbox[t]{0.13\textwidth}{Reasoning (8)}} & ARC-AGI-1 & \% correct & 400 & \href{https://arcprize.org/arc-agi/1/}{\faExternalLink*} \\
 & ARC-AGI-2 & \% correct & 400 & \href{https://arcprize.org/arc-agi/2/}{\faExternalLink*} \\
 & BigBench Hard (BBH) & --- & --- & \href{https://arxiv.org/abs/2210.09261}{\faExternalLink*} \\
 & DROP & \% & 9{,}536 & \href{https://huggingface.co/datasets/EleutherAI/drop}{\faExternalLink*} \\
 & Global PIQA & --- & 6{,}283 & \href{https://huggingface.co/datasets/mrlbenchmarks/global-piqa-parallel}{\faExternalLink*} \\
 & HLE (Humanity's Last Exam) & \% correct & 2{,}500 & \href{https://lastexam.ai/}{\faExternalLink*} \\
 & HLE (w/ tools) & accuracy (\%) & 2{,}500 & \href{https://huggingface.co/moonshotai/Kimi-K2.5}{\faExternalLink*} \\
 & HLE Text & \% & 2{,}158 & \href{https://labs.scale.com/leaderboard/humanitys_last_exam_text_only}{\faExternalLink*} \\
\cmidrule(lr){1-5}
\multirow{5}{*}{\parbox[t]{0.13\textwidth}{Hallucination \& factuality (5)}} & FACTS Grounding & --- & 1{,}719 & \href{https://arxiv.org/abs/2501.03200}{\faExternalLink*} \\
 & FActScore (hallucination rate) & \% & 500 & \href{https://github.com/shmsw25/FActScore}{\faExternalLink*} \\
 & LongFact-Concepts (hallucination rate) & \% & 1{,}140 & \href{https://github.com/google-deepmind/long-form-factuality/tree/main/longfact}{\faExternalLink*} \\
 & LongFact-Objects (hallucination rate) & \% & 1{,}140 & \href{https://github.com/google-deepmind/long-form-factuality/tree/main/longfact}{\faExternalLink*} \\
 & TruthfulQA & --- & 817 & \href{https://github.com/sylinrl/TruthfulQA}{\faExternalLink*} \\
\cmidrule(lr){1-5}
\multirow{4}{*}{\parbox[t]{0.13\textwidth}{Science (4)}} & CritPt & \% correct & 70 & \href{https://huggingface.co/datasets/CritPt-Benchmark/CritPt}{\faExternalLink*} \\
 & GPQA Diamond & \% correct & 198 & \href{https://arxiv.org/abs/2311.12022}{\faExternalLink*} \\
 & GPQA Main (full set) & --- & 448 & \href{https://arxiv.org/abs/2311.12022}{\faExternalLink*} \\
 & SuperGPQA & \% & 26{,}529 & \href{https://huggingface.co/datasets/m-a-p/SuperGPQA}{\faExternalLink*} \\
\cmidrule(lr){1-5}
\multirow{7}{*}{\parbox[t]{0.13\textwidth}{Other (7)}} & AA Intelligence Index & index score & 12{,}826 & \href{https://artificialanalysis.ai/methodology/intelligence-benchmarking}{\faExternalLink*} \\
 & AlpacaEval 2.0 (LC-winrate) & \% & --- & \href{https://arxiv.org/abs/2501.12948}{\faExternalLink*} \\
 & Bullshit-Bench (Clear Pushback) & \% clear pushback & 55 & \href{https://github.com/petergpt/bullshit-benchmark}{\faExternalLink*} \\
 & Chatbot Arena Elo & Elo rating & 8{,}000 & \href{https://arxiv.org/abs/2403.04132}{\faExternalLink*} \\
 & CLUEWSC & \% & 2{,}574 & \href{https://huggingface.co/datasets/clue/clue}{\faExternalLink*} \\
 & LiveBench & overall score & 1{,}000 & \href{https://github.com/LiveBench/LiveBench}{\faExternalLink*} \\
 & Safety (OLMES suite) & --- & --- & \href{https://arxiv.org/abs/2501.00656}{\faExternalLink*} \\
\end{longtable}
}

\begin{table*}[!htbp]
\caption{\textbf{Model inventory.} All 84 models from 13 providers. \emph{R} = reasoning (chain-of-thought). \emph{O} = open-weight. Parameter counts in billions; ``--- = undisclosed. Active parameters shown only for MoE models.}
\label{tab:models}
\tiny
\setlength{\tabcolsep}{0.5pt}
\begin{minipage}[t]{0.495\textwidth}
\centering
\begin{tabular}{@{}p{0.16\linewidth}p{0.39\linewidth}p{0.07\linewidth}p{0.07\linewidth}ccp{0.11\linewidth}c@{}}
\toprule
\textbf{Provider} & \textbf{Model} & \textbf{B} & \textbf{Act.} & \textbf{R} & \textbf{O} & \textbf{Rel.} & \textbf{\faExternalLink*} \\
\midrule
\multirow{20}{*}{\parbox[t]{0.16\linewidth}{OpenAI}} & GPT-3.5 Turbo (0125) & --- & --- & \xmark & \xmark & 2024-01 & \href{https://platform.openai.com/docs/models}{\faExternalLink*} \\
 & GPT-4o (2024-05-13) & --- & --- & \xmark & \xmark & 2024-05 & \href{https://openai.com/index/hello-gpt-4o/}{\faExternalLink*} \\
 & GPT-4o mini & --- & --- & \xmark & \xmark & 2024-07 & \href{https://openai.com/index/gpt-4o-mini-advancing-cost-efficient-intelligence/}{\faExternalLink*} \\
 & GPT-4o (2024-11-20) & --- & --- & \xmark & \xmark & 2024-11 & \href{https://openai.com/index/hello-gpt-4o/}{\faExternalLink*} \\
 & OpenAI o1 (high) & --- & --- & \cmark & \xmark & 2024-12 &  \\
 & o3-mini (high) & --- & --- & \cmark & \xmark & 2025-01 & \href{https://github.com/openai/simple-evals}{\faExternalLink*} \\
 & GPT-4.5 & --- & --- & \xmark & \xmark & 2025-02 & \href{https://www.helicone.ai/blog/gpt-4.5-benchmarks}{\faExternalLink*} \\
 & GPT-4.1 & --- & --- & \xmark & \xmark & 2025-04 & \href{https://arxiv.org/abs/2507.20534}{\faExternalLink*} \\
 & GPT-4.1 mini & --- & --- & \xmark & \xmark & 2025-04 & \href{https://www.helicone.ai/blog/gpt-4.1-full-developer-guide}{\faExternalLink*} \\
 & GPT-4.1 nano & --- & --- & \xmark & \xmark & 2025-04 & \href{https://www.datacamp.com/blog/gpt-4-1}{\faExternalLink*} \\
 & o3 (high) & --- & --- & \cmark & \xmark & 2025-04 & \href{https://openai.com/index/introducing-o3-and-o4-mini/}{\faExternalLink*} \\
 & o4-mini (high) & --- & --- & \cmark & \xmark & 2025-04 & \href{https://www.datacamp.com/blog/o4-mini}{\faExternalLink*} \\
 & GPT-5 mini & --- & --- & \cmark & \xmark & 2025-07 & \href{https://openai.com/index/introducing-gpt-5-2/}{\faExternalLink*} \\
 & GPT-5 nano & --- & --- & \cmark & \xmark & 2025-07 & \href{https://openai.com/index/introducing-gpt-5-2/}{\faExternalLink*} \\
 & GPT-5 & --- & --- & \cmark & \xmark & 2025-08 & \href{https://openai.com/index/introducing-gpt-5/}{\faExternalLink*} \\
 & gpt-oss-120B & 116.8 & 5.1 & \cmark & \cmark & 2025-08 & \href{https://arxiv.org/abs/2508.10925}{\faExternalLink*} \\
 & GPT-5.1 & --- & --- & \cmark & \xmark & 2025-11 & \href{https://www.vellum.ai/blog/gpt-5-2-benchmarks}{\faExternalLink*} \\
 & GPT-5.2 & --- & --- & \cmark & \xmark & 2025-12 & \href{https://openai.com/index/introducing-gpt-5-2/}{\faExternalLink*} \\
 & GPT-5.4 & --- & --- & \cmark & \xmark & 2026-03 &  \\
 & GPT-5.5 & --- & --- & \cmark & \xmark & 2026-04 &  \\
\cmidrule(lr){1-8}
\multirow{12}{*}{\parbox[t]{0.16\linewidth}{Google}} & Gemini 1.5 Flash & --- & --- & \xmark & \xmark & 2024-05 & \href{https://deepmind.google/technologies/gemini/flash/}{\faExternalLink*} \\
 & Gemini 1.5 Pro & --- & --- & \xmark & \xmark & 2024-05 & \href{https://deepmind.google/technologies/gemini/pro/}{\faExternalLink*} \\
 & Gemma 2 27B & 27 & 27 & \xmark & \cmark & 2024-06 & \href{https://blog.google/technology/developers/google-gemma-2/}{\faExternalLink*} \\
 & Gemma 2 9B & 9 & 9 & \xmark & \cmark & 2024-06 & \href{https://blog.google/technology/developers/google-gemma-2/}{\faExternalLink*} \\
 & Gemma 3 1B & --- & --- & \cmark & \cmark & 2025 & \href{https://blog.google/technology/developers/gemma-3/}{\faExternalLink*} \\
 & Gemini 2.0 Flash & --- & --- & \xmark & \xmark & 2025-02 & \href{https://artificialanalysis.ai/models/gemini-2-0-flash}{\faExternalLink*} \\
 & Gemma 3 27B & 27 & 27 & \xmark & \cmark & 2025-03 & \href{https://llm-stats.com/benchmarks}{\faExternalLink*} \\
 & Gemini 2.5 Flash & --- & --- & \cmark & \xmark & 2025-05 & \href{https://llm-stats.com/models/gemini-2.5-flash}{\faExternalLink*} \\
 & Gemini 2.5 Pro (GA) & --- & --- & \cmark & \xmark & 2025-06 & \href{https://blog.google/technology/google-deepmind/gemini-model-thinking-updates-march-2025/}{\faExternalLink*} \\
 & Gemini 3 Flash & --- & --- & \cmark & \xmark & 2025-11 & \href{https://www.vellum.ai/blog/google-gemini-3-benchmarks}{\faExternalLink*} \\
 & Gemini 3 Pro & --- & --- & \cmark & \xmark & 2025-11 & \href{https://www.vellum.ai/blog/google-gemini-3-benchmarks}{\faExternalLink*} \\
 & Gemini 3.1 Pro & --- & --- & \cmark & \xmark & 2026-02 & \href{https://www.digitalapplied.com/blog/google-gemini-3-1-pro-benchmarks-pricing-guide}{\faExternalLink*} \\
\cmidrule(lr){1-8}
\multirow{11}{*}{\parbox[t]{0.16\linewidth}{Anthropic}} & Claude 3.5 Sonnet (1022) & --- & --- & \xmark & \xmark & 2024-10 & \href{https://www.anthropic.com/news/claude-3-5-sonnet}{\faExternalLink*} \\
 & Claude 3.7 Sonnet & --- & --- & \cmark & \xmark & 2025-02 & \href{https://www.anthropic.com/news/claude-3-7-sonnet}{\faExternalLink*} \\
 & Claude Opus 4 & --- & --- & \cmark & \xmark & 2025-05 & \href{https://arxiv.org/abs/2507.20534}{\faExternalLink*} \\
 & Claude Sonnet 4 & --- & --- & \xmark & \xmark & 2025-05 & \href{https://arxiv.org/abs/2507.20534}{\faExternalLink*} \\
 & Claude Opus 4.1 & --- & --- & \cmark & \xmark & 2025-08 & \href{https://www.anthropic.com/news/claude-opus-4-1}{\faExternalLink*} \\
 & Claude Sonnet 4.5 & --- & --- & \cmark & \xmark & 2025-09 & \href{https://www.anthropic.com/news/claude-sonnet-4-5}{\faExternalLink*} \\
 & Claude Haiku 4.5 & --- & --- & \cmark & \xmark & 2025-10 & \href{https://www.anthropic.com/news/claude-haiku-4-5}{\faExternalLink*} \\
 & Claude Opus 4.5 & --- & --- & \cmark & \xmark & 2025-11 & \href{https://www.anthropic.com/news/claude-opus-4-5}{\faExternalLink*} \\
 & Claude Opus 4.6 & --- & --- & \cmark & \xmark & 2026-02 & \href{https://www.vellum.ai/blog/claude-opus-4-6-benchmarks}{\faExternalLink*} \\
 & Claude Sonnet 4.6 & --- & --- & \cmark & \xmark & 2026-02 & \href{https://www.anthropic.com/news/claude-sonnet-4-6}{\faExternalLink*} \\
 & Claude Opus 4.7 & --- & --- & \cmark & \xmark & 2026-04 &  \\
\bottomrule
\end{tabular}
\end{minipage}%
\hfill
\begin{minipage}[t]{0.495\textwidth}
\centering
\begin{tabular}{@{}p{0.16\linewidth}p{0.39\linewidth}p{0.07\linewidth}p{0.07\linewidth}ccp{0.11\linewidth}c@{}}
\toprule
\textbf{Provider} & \textbf{Model} & \textbf{B} & \textbf{Act.} & \textbf{R} & \textbf{O} & \textbf{Rel.} & \textbf{\faExternalLink*} \\
\midrule
\multirow{11}{*}{\parbox[t]{0.16\linewidth}{Alibaba}} & Qwen2.5 72B Instruct & --- & --- & \xmark & \cmark & 2024-09 & \href{https://qwenlm.github.io/blog/qwen2.5/}{\faExternalLink*} \\
 & Qwen2.5-14B & 14 & 14 & \xmark & \cmark & 2024-09 &  \\
 & Qwen2.5-32B-Instruct & 32 & 32 & \xmark & \cmark & 2024-09 &  \\
 & Qwen2.5-7B-Instruct & 7 & 7 & \xmark & \cmark & 2024-09 &  \\
 & QwQ-32B & 32.8 & 32.8 & \cmark & \cmark & 2025-03 & \href{https://qwenlm.github.io/blog/qwq-32b/}{\faExternalLink*} \\
 & Qwen3-235B-A22B & 235 & 22 & \cmark & \cmark & 2025-05 & \href{https://qwenlm.github.io/blog/qwen3/}{\faExternalLink*} \\
 & Qwen3-30B-A3B & 30 & 3 & \cmark & \cmark & 2025-05 &  \\
 & Qwen3-32B & 32 & 32 & \cmark & \cmark & 2025-05 & \href{https://arxiv.org/abs/2505.09388}{\faExternalLink*} \\
 & Qwen3-8B & 8 & 8 & \cmark & \cmark & 2025-05 &  \\
 & Qwen3.5-397B-A17B & 397 & 17 & \cmark & \cmark & 2026-02 & \href{https://venturebeat.com/technology/alibabas-qwen-3-5-397b-a17/}{\faExternalLink*} \\
 & Qwen3.6-Plus & --- & --- & \cmark & \xmark & 2026-03 & \href{https://docs.apiyi.com/en/news/qwen-3-6-plus-launch}{\faExternalLink*} \\
\cmidrule(lr){1-8}
\multirow{9}{*}{\parbox[t]{0.16\linewidth}{DeepSeek}} & DeepSeek-V2-0506 & --- & --- & \xmark & \cmark & 2024-05 & \href{https://github.com/deepseek-ai/DeepSeek-V2}{\faExternalLink*} \\
 & DeepSeek-V2.5-0905 & --- & --- & \xmark & \cmark & 2024-09 & \href{https://github.com/deepseek-ai/DeepSeek-V2.5}{\faExternalLink*} \\
 & DeepSeek-V3 & 671 & 37 & \xmark & \cmark & 2025-01 & \href{https://github.com/deepseek-ai/DeepSeek-V3}{\faExternalLink*} \\
 & DeepSeek-R1 & 671 & 37 & \cmark & \cmark & 2025-01 & \href{https://github.com/deepseek-ai/DeepSeek-R1}{\faExternalLink*} \\
 & DeepSeek-R1-Distill-Llama-70B & 70 & 70 & \cmark & \cmark & 2025-01 &  \\
 & DeepSeek-R1-0528 & 671 & 37 & \cmark & \cmark & 2025-05 & \href{https://huggingface.co/deepseek-ai/DeepSeek-R1-0528}{\faExternalLink*} \\
 & DeepSeek-V3.2 & 671 & 37 & \cmark & \cmark & 2025-12 & \href{https://arxiv.org/abs/2512.02556}{\faExternalLink*} \\
 & DeepSeek-V4-Flash & 284 & 13 & \cmark & \cmark & 2026-04 &  \\
 & DeepSeek-V4-Pro & 1600 & 49 & \cmark & \cmark & 2026-04 &  \\
\cmidrule(lr){1-8}
\multirow{6}{*}{\parbox[t]{0.16\linewidth}{Meta}} & LLaMA-3.1 405B Instruct & --- & --- & \xmark & \cmark & 2024-07 &  \\
 & Llama 3.1 8B Instruct & 8 & 8 & \xmark & \cmark & 2024-07 &  \\
 & Llama 3.2 1B & --- & --- & \cmark & \cmark & 2024-09 &  \\
 & Llama-3.3-70B-Instruct & 70 & 70 & \xmark & \cmark & 2024-12 &  \\
 & Llama 4 Maverick & 402 & 17 & \xmark & \cmark & 2025-04 & \href{https://ai.meta.com/blog/llama-4-multimodal-intelligence/}{\faExternalLink*} \\
 & Muse Spark & --- & --- & \cmark & \xmark & 2026-04 & \href{https://about.fb.com/news/2026/04/introducing-muse-spark-meta-superintelligence-labs/}{\faExternalLink*} \\
\cmidrule(lr){1-8}
\multirow{4}{*}{\parbox[t]{0.16\linewidth}{Zhipu AI}} & GLM-4.6 & --- & --- & \cmark & \xmark & 2025-09 & \href{https://llm-stats.com/models/glm-4.6}{\faExternalLink*} \\
 & GLM-4.7 & --- & --- & \cmark & \xmark & 2025-12 & \href{https://medium.com/@leucopsis/a-technical-analysis-of-glm-4-7}{\faExternalLink*} \\
 & GLM-5 & --- & --- & \cmark & \cmark & 2026-03 &  \\
 & GLM-5.1 & --- & --- & \cmark & \cmark & 2026-04 & \href{https://docs.apiyi.com/en/news/glm-5-1-launch}{\faExternalLink*} \\
\cmidrule(lr){1-8}
\multirow{3}{*}{\parbox[t]{0.16\linewidth}{Moonshot AI}} & Kimi K2 & --- & --- & \xmark & \cmark & 2025-07 & \href{https://arxiv.org/abs/2507.20534}{\faExternalLink*} \\
 & Kimi K2.5 & --- & --- & \cmark & \cmark & 2026-01 & \href{https://huggingface.co/moonshotai/Kimi-K2.5}{\faExternalLink*} \\
 & Kimi K2.6 & --- & --- & \cmark & \cmark & 2026-04 &  \\
\cmidrule(lr){1-8}
\multirow{3}{*}{\parbox[t]{0.16\linewidth}{xAI}} & Grok 3 Beta & --- & --- & \xmark & \xmark & 2025-02 & \href{https://x.ai/news/grok-3}{\faExternalLink*} \\
 & Grok 4 & --- & --- & \cmark & \xmark & 2025-07 & \href{https://matharena.ai/}{\faExternalLink*} \\
 & Grok 4.1 & --- & --- & \cmark & \xmark & 2025-11 & \href{https://matharena.ai/}{\faExternalLink*} \\
\cmidrule(lr){1-8}
\multirow{2}{*}{\parbox[t]{0.16\linewidth}{MiniMax}} & MiniMax-M2 & --- & --- & \cmark & \xmark & 2025-10 & \href{https://artificialanalysis.ai/models/minimax-m2}{\faExternalLink*} \\
 & MiniMax M2.1 & --- & --- & \cmark & \cmark & 2025-12 &  \\
\cmidrule(lr){1-8}
Cohere & Command A & 111 & --- & \xmark & \cmark & 2025-03 & \href{https://cohere.com/research/papers/command-a-technical-report.pdf}{\faExternalLink*} \\
\cmidrule(lr){1-8}
ByteDance & Doubao Seed 2.0 Pro & --- & --- & \cmark & \xmark & 2026-02 & \href{https://www.digitalapplied.com/blog/bytedance-seed-2-doubao-ai-model-benchmarks-guide}{\faExternalLink*} \\
\cmidrule(lr){1-8}
Mistral & Ministral 8B Instruct 2410 & 8 & 8 & \xmark & \cmark & 2024-10 &  \\
\bottomrule
\end{tabular}
\end{minipage}
\end{table*}

\subsection{Rank-2 Geometry}
\label{app:rank_geometry}

\begin{figure}[!htbp]
\centering
\includegraphics[width=0.62\linewidth]{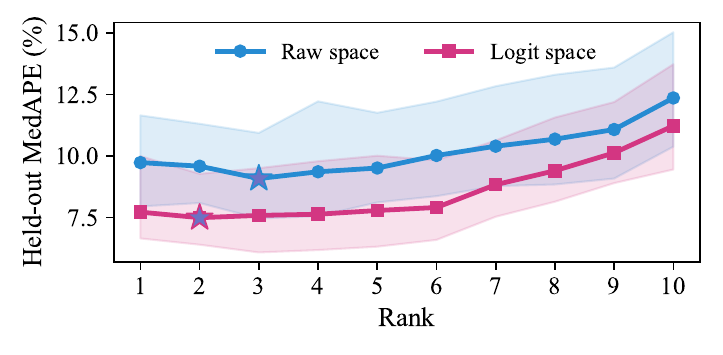}
\caption{\textbf{Rank sweep with the newest $20\%$ of models held out as a block.} Held-out MedAPE (\%) versus Soft-Impute rank, in raw and logit score space, with $95\%$ cluster-bootstrap intervals over the $17$ held-out models; stars mark each curve's minimum. In logit space (the space \benchpress{} uses) rank~2 remains the minimum even though no sibling of a target model is in the training matrix; in raw space ranks~2 and~3 fall within each other's interval.}
\label{fig:rank_ucurve_holdout}
\end{figure}

\paragraph{Rank selection under held-out models.}
The rank sweep in \Cref{sec:bp_svd} selects each rank on the canonical per-model folds, where every model contributes some training scores. A skeptical reading is that the rank-2 minimum could come from copying a near-duplicate sibling row of the target rather than from genuine low-rank structure, and \Cref{fig:rank_ucurve} reports no uncertainty. We therefore repeat the sweep under a stricter split: we hold out the newest $17$ models (the most recent $20\%$ by release date) as a block, removing their rows from the training matrix entirely, and predict each held-out model on its own from the older models plus two thirds of its own observed scores, so no other recent model is available to copy from. \Cref{fig:rank_ucurve_holdout} plots held-out MedAPE against rank with $95\%$ cluster-bootstrap intervals over the $17$ target models.

Low rank continues to predict best under this split: the logit-space minimum stays at rank~2, and the raw-space curve is flat across the first three ranks, whose intervals overlap. The rank-2 geometry is therefore not an artifact of near-duplicate sibling rows.

\section{Supplemental to \texorpdfstring{\Cref{sec:bp_methods}}{Section 4}: \benchpress{}: A Low-rank Benchmark Score Predictor}
\label{app:methods}

\subsection{Candidate Methods}
\label{app:method_details}

This appendix gives the formal definitions of the matrix-completion methods compared in \Cref{sec:method_comparison}.
All methods below operate on the same transformed, column-standardized matrix.

We use the following notation throughout this subsection:
\begin{itemize}[nosep,leftmargin=*]
  \item $X\in\mathbb{R}^{M\times B}$ is the transformed, column-standardized version of the adopted score matrix, with $M=84$ models and $B=133$ benchmarks.
  \item $x_{mb}$ is the entry for model $m$ on benchmark $b$ in this transformed space.
  \item $\Omega$ is the set of observed cells.
  \item $\Omega_m=\{b:(m,b)\in\Omega\}$ is the set of benchmarks observed for model $m$.
  \item $\Omega^b=\{m:(m,b)\in\Omega\}$ is the set of models observed for benchmark $b$.
  \item $\bar{x}_{m\cdot}=|\Omega_m|^{-1}\sum_{b\in\Omega_m}x_{mb}$ is the observed mean for model $m$.
  \item $\bar{x}_{\cdot b}=|\Omega^b|^{-1}\sum_{m\in\Omega^b}x_{mb}$ is the observed mean for benchmark $b$.
  \item $\bar{x}=|\Omega|^{-1}\sum_{(m,b)\in\Omega}x_{mb}$ is the observed global mean.
  \item $\hat{x}_{mb}$ is a method's prediction for cell $(m,b)$.
  \item $R$ is the rank used by low-rank methods in this subsection.
  \item $N_k(t;\rho)$ is a method-local top-$k$ neighbor set. For benchmark targets, $t=b\in\{1,\ldots,B\}$ and candidate neighbors are $u\in\{1,\ldots,B\}\setminus\{b\}$; for model targets, $t=m\in\{1,\ldots,M\}$ and candidate neighbors are $u\in\{1,\ldots,M\}\setminus\{m\}$. The scoring function $\rho(t,u)\in\mathbb{R}$ ranks each eligible candidate $u$ for target $t$; larger scores are preferred, so distances are used with a minus sign.
\end{itemize}
After a method produces predictions in this space, the pipeline maps them back to the original score scale by undoing the standardization and feature transform.

\paragraph{Benchmark mean.}
The benchmark-mean baseline predicts each missing cell from the observed mean of the target benchmark:
\[
  \hat{x}_{mb}=\bar{x}_{\cdot b}.
\]

\paragraph{Model mean.}
The model-mean baseline predicts each missing cell from the observed mean of the target model:
\[
  \hat{x}_{mb}=\bar{x}_{m\cdot}.
\]

\paragraph{Bench-KNN.}
Let $\Omega^{bb'}=\Omega^b\cap\Omega^{b'}$, and let $\operatorname{corr}(b,b')\in[-1,1]$ be the Pearson correlation between benchmark columns $b$ and $b'$ over $\Omega^{bb'}$ when this correlation is defined.
Here $N_k(b;\operatorname{corr})$ selects benchmark neighbors $b'\ne b$ using score $\rho(b,b')=\operatorname{corr}(b,b')$.
For a missing cell $(m,b)$, Bench-KNN predicts by the correlation-weighted average over $N_k(b;\operatorname{corr})\cap\Omega_m$:
\[
  \hat{x}_{mb}=
  \frac{\sum_{b'\in N_k(b;\operatorname{corr})\cap\Omega_m}\max(\operatorname{corr}(b,b'),0.01)x_{mb'}}
       {\sum_{b'\in N_k(b;\operatorname{corr})\cap\Omega_m}\max(\operatorname{corr}(b,b'),0.01)}.
\]
If $N_k(b;\operatorname{corr})\cap\Omega_m$ is empty, it falls back to $\bar{x}_{\cdot b}$.

\paragraph{Model-KNN.}
Let $\Omega_{mm'}=\Omega_m\cap\Omega_{m'}$.
For each model pair $(m,m')$ with $|\Omega_{mm'}|\ge3$, define the shared-benchmark distance function $\Delta(m,m')\in\mathbb{R}_{\ge0}$ by
\[
  \Delta(m,m')=
  \sqrt{\frac{1}{|\Omega_{mm'}|}
  \sum_{b\in\Omega_{mm'}}(x_{mb}-x_{m'b})^2}.
\]
Here $N_k(m;-\Delta)$ selects model neighbors $m'\ne m$ using score $\rho(m,m')=-\Delta(m,m')$.
For a missing cell $(m,b)$, Model-KNN predicts by the average over $N_k(m;-\Delta)\cap\Omega^b$:
\[
  \hat{x}_{mb}=
  \frac{1}{|N_k(m;-\Delta)\cap\Omega^b|}
  \sum_{m'\in N_k(m;-\Delta)\cap\Omega^b}x_{m'b}.
\]
If $N_k(m;-\Delta)\cap\Omega^b$ is empty, it falls back to $\bar{x}_{\cdot b}$.

\paragraph{BenchReg.}
For each target benchmark $b$ and predictor benchmark $b'$, BenchReg fits a one-dimensional linear predictor $f_{bb'}:\mathbb R\to\mathbb R$ on $\Omega^{bb'}=\Omega^b\cap\Omega^{b'}$.
Here $R^2(f_{bb'})\in\mathbb{R}$ is the coefficient of determination of this linear fit on the shared observations.
For BenchReg, $N_k(b;R^2)$ selects benchmark neighbors $b'\ne b$ with $|\Omega^{bb'}|\ge5$ and $R^2(f_{bb'})\ge R^2_{\min}$ using score $\rho(b,b')=R^2(f_{bb'})$.
For a missing cell $(m,b)$, BenchReg predicts by the $R^2$-weighted average over $N_k(b;R^2)\cap\Omega_m$:
\[
  \hat{x}_{mb}=
  \frac{\sum_{b'\in N_k(b;R^2)\cap\Omega_m}R^2(f_{bb'})\,f_{bb'}(x_{mb'})}
       {\sum_{b'\in N_k(b;R^2)\cap\Omega_m}R^2(f_{bb'})}.
\]
If $N_k(b;R^2)\cap\Omega_m$ is empty, BenchReg leaves the cell unpredicted.

\paragraph{ModelReg.}
ModelReg is the row-wise counterpart of BenchReg.
For each target model $m$ and predictor model $m'$, it fits a one-dimensional linear predictor $f_{mm'}:\mathbb R\to\mathbb R$ on $\Omega_{mm'}=\Omega_m\cap\Omega_{m'}$.
Here $R^2(f_{mm'})\in\mathbb{R}$ is the coefficient of determination of this linear fit on the shared benchmarks.
For ModelReg, $N_k(m;R^2)$ selects model neighbors $m'\ne m$ with $|\Omega_{mm'}|\ge5$ and $R^2(f_{mm'})\ge R^2_{\min}$ using score $\rho(m,m')=R^2(f_{mm'})$.
For a missing cell $(m,b)$, ModelReg predicts by the $R^2$-weighted average over $N_k(m;R^2)\cap\Omega^b$:
\[
  \hat{x}_{mb}=
  \frac{\sum_{m'\in N_k(m;R^2)\cap\Omega^b}R^2(f_{mm'})\,f_{mm'}(x_{m'b})}
       {\sum_{m'\in N_k(m;R^2)\cap\Omega^b}R^2(f_{mm'})}.
\]
If $N_k(m;R^2)\cap\Omega^b$ is empty, ModelReg leaves the cell unpredicted.

\paragraph{Soft-Impute.}
Soft-Impute~\citep{mazumder2010} initializes missing cells and then alternates between a rank-$R$ SVD projection and clamping the observed entries:
\[
  X^{(\ell+1)}_{mb}=x_{mb}\quad\text{for }(m,b)\in\Omega,\qquad
  X^{(\ell+1)}_{\Omega^c}=\bigl[\operatorname{SVD}_R(X^{(\ell)})\bigr]_{\Omega^c}.
\]
The fixed point gives predictions $\hat{x}_{mb}=X^{(\infty)}_{mb}$.

\paragraph{Bias-decomposed ALS.}
Bias-decomposed ALS adds a residual correction $UV^\top$, with $U\in\mathbb{R}^{M\times R}$ and $V\in\mathbb{R}^{B\times R}$, fitted by
\[
  \begin{aligned}
  (U,V)=\arg\min_{U,V}\;&
  \sum_{(m,b)\in\Omega}
  \Bigl[
    x_{mb}
    -\bigl(\bar{x}+(\bar{x}_{m\cdot}-\bar{x})+(\bar{x}_{\cdot b}-\bar{x})\bigr)
    -(UV^\top)_{mb}
  \Bigr]^2 \\
  &+\lambda\left(\|U\|_F^2+\|V\|_F^2\right).
  \end{aligned}
\]
The prediction is therefore
\[
  \hat{x}_{mb}
  =
  \underbrace{\bar{x}}_{\text{global level}}
  +\underbrace{(\bar{x}_{m\cdot}-\bar{x})}_{\text{model }m\text{ offset}}
  +\underbrace{(\bar{x}_{\cdot b}-\bar{x})}_{\text{benchmark }b\text{ offset}}
  +\underbrace{(UV^\top)_{mb}}_{\text{rank-}R\text{ residual correction}}.
\]
The correction satisfies $UV^\top\in\mathbb{R}^{M\times B}$ and has rank at most $R$ because $U$ and $V$ each have $R$ columns.
The default predictor uses rank $R=2$, $\lambda=0.1$, and averages the completed matrices from 10 random initializations (seeds 42--51).

\paragraph{NMF.}
Non-negative matrix factorization (NMF)~\citep{lee1999nmf} first shifts each benchmark column, if needed, so that the observed entries are non-negative.
Writing the shifted observed entries as $x'_{mb}\in\mathbb{R}_{\ge0}$, it solves
\[
  (U,V)
  =
  \arg\min_{\substack{U\in\mathbb{R}_+^{M\times R}\\V\in\mathbb{R}_+^{B\times R}}}
  \sum_{(m,b)\in\Omega}\bigl[x'_{mb}-(UV^\top)_{mb}\bigr]^2
  + \lambda(\|U\|_F^2+\|V\|_F^2),
\]
then subtracts the column shifts from $UV^\top$ to obtain predictions in the original transformed space.

\paragraph{PMF.}
Probabilistic matrix factorization (PMF)~\citep{salakhutdinov2008pmf} uses the same factor-matrix dimensions without the non-negativity constraint:
\[
  (U,V)
  =
  \arg\min_{\substack{U\in\mathbb{R}^{M\times R}\\V\in\mathbb{R}^{B\times R}}}
  \sum_{(m,b)\in\Omega}\bigl[x_{mb}-(UV^\top)_{mb}\bigr]^2
  + \lambda(\|U\|_F^2+\|V\|_F^2),
\]
with prediction $\hat{x}_{mb}=(UV^\top)_{mb}$.

\paragraph{Nuclear norm minimization.}
The nuclear-norm baseline~\citep{candes2009} solves the convex low-rank surrogate
\[
  Z^\star
  =
  \arg\min_{Z\in\mathbb{R}^{M\times B}}
  \frac12\sum_{(m,b)\in\Omega}(Z_{mb}-x_{mb})^2+\lambda\|Z\|_*,
\]
and predicts $\hat{x}_{mb}=Z^\star_{mb}$.

\paragraph{Neural baseline.}
Let $\tilde{x}_m\in\mathbb{R}^B$ be row $m$ with missing entries filled by zero, and let $o_m\in\{0,1\}^B$ be its binary observation mask.
The MLP baseline trains a two-layer network $f_\theta$ with masked reconstruction loss, where $\odot$ denotes elementwise multiplication:
\[
  \min_\theta
  \sum_m \bigl\|o_m\odot(f_\theta(\tilde{x}_m)-\tilde{x}_m)\bigr\|_2^2,
\]
and predicts $\hat{x}_{mb}=[f_\theta(\tilde{x}_m)]_b$, averaged over three random seeds.

\subsection{From Candidate Methods to \benchpress{}}
\label{app:method_comparison}

This appendix expands the head-to-head comparison of \Cref{sec:method_comparison} from a small selected set into the full transform $\times$ method grid, reported both as a heatmap and as a sortable table.

\begin{figure*}[!htbp]
\centering
\includegraphics[width=0.95\textwidth]{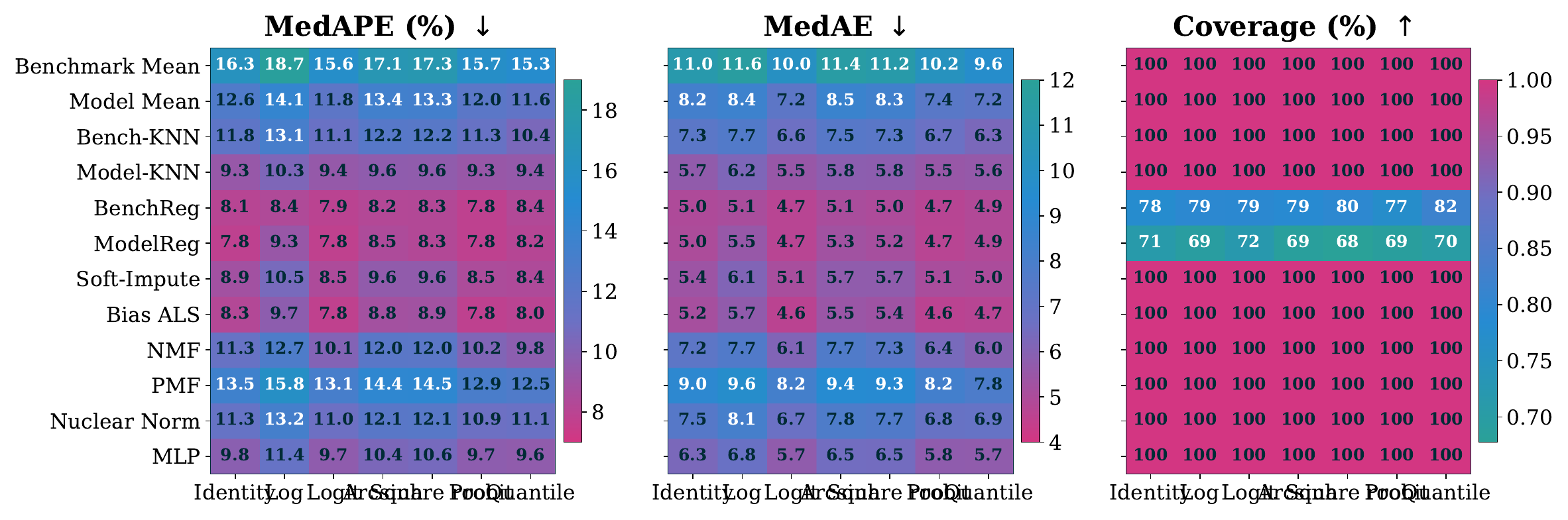}
\caption{Full transform $\times$ method grid (7 transforms, 12 methods) across the Section 4 metrics: MedAPE, MedAE, and coverage. Each cell reports the hyperparameter selected on nested validation cells, evaluated on the outer test cells as the median over 10 seeds $\times$ 3 folds. All methods operate in standardized space. Green = better.}
\label{fig:transform_method_grid}
\end{figure*}

The full transform $\times$ method grid (\Cref{fig:transform_method_grid}) evaluates all 84 combinations on a common experiment setting; \Cref{tab:full_grid} reports the same numbers in tabular form, sorted by $\mathsf{MedAPE}$.

\paragraph{Model selection.}
\Cref{sec:method_comparison} selects each pair's hyperparameter on validation cells rather than on the held-out folds it reports, using nested cross-validation. The outer folds are unchanged. Within each outer fold we draw a second per-model partition of the training cells and use two of its three parts as validation cells, so the inner training set retains roughly the same fraction of the outer training cells that the outer training set retains of the matrix. Every candidate configuration is fit on the inner training cells and scored on the validation cells, and the configuration with the lowest validation $\mathsf{MedAPE}$ is then evaluated on the untouched outer test cells. \Cref{tab:model_selection} ranks the 203 configurations that achieve 100\% coverage on the outer held-out cells. Within this full-coverage set, the pipeline, seeds, metrics, and aggregation match \Cref{tab:top15}; only the scored cells change from outer test to inner validation. \Cref{tab:full_grid} and \Cref{fig:transform_method_grid} report each pair under this nested selection, and where the winning hyperparameter varies across outer folds they report the one selected most often.

\begin{table*}[!htbp]
\centering
\caption{Top-15 transform $\times$ method configurations ranked independently by validation error among the 203 configurations with 100\% coverage on the outer held-out cells. The full-coverage filter is applied before ranking. Within this set, the prediction pipeline, seeds, metrics, and aggregation match \Cref{tab:top15}; each configuration is fit on inner training cells and scored on inner validation cells. \# is per-metric rank; values are shown as metric value followed by validation coverage, rounded to the nearest whole percent, in parentheses. All results use standardization and report the median over 10 seeds $\times$ 3 outer folds $\times$ 2 inner validation folds. Pink highlights mark the Logit Bias ALS configuration adopted as \benchpress{}'s main score predictor.}
\label{tab:model_selection}
\resizebox{\textwidth}{!}{%
\begin{tabular}{@{}rlllr@{\hspace{12pt}}rlllr@{}}
\toprule
\multicolumn{5}{c}{\textbf{Validation MedAPE (\%) $\downarrow$}} &
\multicolumn{5}{c}{\textbf{Validation MedAE $\downarrow$}} \\
\cmidrule(r){1-5} \cmidrule(l){6-10}
\# & Transform & Method & Hyperparameter & Value &
\# & Transform & Method & Hyperparameter & Value \\
\midrule
1 & \cBP\textbf{Logit} & \cBP\textbf{Bias ALS} & \cBP\textbf{$\lambda{=}0.1$, $r{=}2$} & \cBP\textbf{8.9 (100\%)} & 1 & \cBP\textbf{Logit} & \cBP\textbf{Bias ALS} & \cBP\textbf{$\lambda{=}0.1$, $r{=}2$} & \cBP\textbf{5.20 (100\%)} \\
2 & Probit & Bias ALS & $\lambda{=}0.1$, $r{=}2$ & 9.0 (100\%) & 2 & Probit & Bias ALS & $\lambda{=}0.1$, $r{=}2$ & 5.29 (100\%) \\
3 & Quantile & Bias ALS & $\lambda{=}0.1$, $r{=}2$ & 9.0 (100\%) & 3 & Quantile & Bias ALS & $\lambda{=}0.1$, $r{=}2$ & 5.39 (100\%) \\
4 & Logit & Bias ALS & $\lambda{=}0.01$, $r{=}2$ & 9.3 (100\%) & 4 & Logit & Bias ALS & $\lambda{=}0.01$, $r{=}2$ & 5.39 (100\%) \\
5 & Quantile & Bias ALS & $\lambda{=}0.01$, $r{=}2$ & 9.3 (100\%) & 5 & Quantile & Bias ALS & $\lambda{=}0.01$, $r{=}2$ & 5.45 (100\%) \\
6 & Logit & Bias ALS & $\lambda{=}1.0$, $r{=}2$ & 9.4 (100\%) & 6 & Probit & Bias ALS & $\lambda{=}0.01$, $r{=}2$ & 5.53 (100\%) \\
7 & Probit & Bias ALS & $\lambda{=}0.01$, $r{=}2$ & 9.4 (100\%) & 7 & Logit & Bias ALS & $\lambda{=}1.0$, $r{=}2$ & 5.62 (100\%) \\
8 & Probit & Bias ALS & $\lambda{=}1.0$, $r{=}2$ & 9.5 (100\%) & 8 & Probit & Bias ALS & $\lambda{=}1.0$, $r{=}2$ & 5.78 (100\%) \\
9 & Identity & Bias ALS & $\lambda{=}0.1$, $r{=}2$ & 9.5 (100\%) & 9 & Logit & Soft-Impute & $r{=}2$ & 5.86 (100\%) \\
10 & Identity & Bias ALS & $\lambda{=}1.0$, $r{=}2$ & 9.8 (100\%) & 10 & Probit & Soft-Impute & $r{=}2$ & 5.88 (100\%) \\
11 & Quantile & Bias ALS & $\lambda{=}1.0$, $r{=}2$ & 9.8 (100\%) & 11 & Quantile & Bias ALS & $\lambda{=}1.0$, $r{=}2$ & 5.89 (100\%) \\
12 & Quantile & Soft-Impute & $r{=}2$ & 9.9 (100\%) & 12 & Quantile & Soft-Impute & $r{=}2$ & 5.93 (100\%) \\
13 & Probit & Soft-Impute & $r{=}2$ & 10.0 (100\%) & 13 & Identity & Bias ALS & $\lambda{=}0.1$, $r{=}2$ & 5.95 (100\%) \\
14 & Logit & Soft-Impute & $r{=}2$ & 10.0 (100\%) & 14 & Square root & Bias ALS & $\lambda{=}0.1$, $r{=}2$ & 6.09 (100\%) \\
15 & Identity & Bias ALS & $\lambda{=}0.01$, $r{=}2$ & 10.0 (100\%) & 15 & Identity & Bias ALS & $\lambda{=}0.01$, $r{=}2$ & 6.14 (100\%) \\
\bottomrule
\end{tabular}
}
\end{table*}

{\footnotesize\setlength{\tabcolsep}{2pt}
\begin{longtable}{@{}rlllrrr@{}}
\caption{Full transform $\times$ method grid: all 84 pairs of feature transform and prediction method from \Cref{sec:method_comparison}, sorted by $\mathsf{MedAPE}$. Each row reports the hyperparameter selected on nested validation cells (\Cref{tab:model_selection}); where that choice varies across outer folds, the one selected most often is shown. Errors are measured on the outer test cells as the median over 10 seeds $\times$ 3 folds in standardized space.}\label{tab:full_grid} \\
\toprule
\# & Transform & Method & Hyperparameter & MedAPE (\%) $\downarrow$ & MedAE $\downarrow$ & Cov. \\
\midrule
\endfirsthead
\multicolumn{7}{c}{\tablename\ \thetable\ -- continued from previous page} \\
\toprule
\# & Transform & Method & Hyperparameter & MedAPE (\%) $\downarrow$ & MedAE $\downarrow$ & Cov. \\
\midrule
\endhead
\midrule \multicolumn{7}{r}{\textit{continued on next page}} \\
\endfoot
\bottomrule
\endlastfoot
1 & Logit & ModelReg & $R^2_{\min}{=}0.3$, $k{=}3$ & 7.75 & 4.69 & 72\% \\
2 & Logit & Bias ALS & $\lambda{=}0.1$, $r{=}2$ & 7.77 & 4.63 & 100\% \\
3 & Probit & Bias ALS & $\lambda{=}0.1$, $r{=}2$ & 7.80 & 4.64 & 100\% \\
4 & Probit & ModelReg & $R^2_{\min}{=}0.3$, $k{=}3$ & 7.81 & 4.71 & 69\% \\
5 & Identity & ModelReg & $R^2_{\min}{=}0.2$, $k{=}3$ & 7.83 & 5.03 & 71\% \\
6 & Probit & BenchReg & $R^2_{\min}{=}0.3$, $k{=}7$ & 7.84 & 4.73 & 77\% \\
7 & Logit & BenchReg & $R^2_{\min}{=}0.3$, $k{=}7$ & 7.95 & 4.69 & 79\% \\
8 & Quantile & Bias ALS & $\lambda{=}0.1$, $r{=}2$ & 8.00 & 4.73 & 100\% \\
9 & Identity & BenchReg & $R^2_{\min}{=}0.3$, $k{=}7$ & 8.10 & 4.96 & 78\% \\
10 & Quantile & ModelReg & $R^2_{\min}{=}0.3$, $k{=}3$ & 8.17 & 4.90 & 70\% \\
11 & Arcsinh & BenchReg & $R^2_{\min}{=}0.3$, $k{=}7$ & 8.23 & 5.06 & 79\% \\
12 & Identity & Bias ALS & $\lambda{=}0.1$, $r{=}2$ & 8.25 & 5.17 & 100\% \\
13 & Square root & ModelReg & $R^2_{\min}{=}0.2$, $k{=}3$ & 8.27 & 5.17 & 68\% \\
14 & Square root & BenchReg & $R^2_{\min}{=}0.3$, $k{=}5$ & 8.29 & 4.99 & 80\% \\
15 & Quantile & BenchReg & $R^2_{\min}{=}0.3$, $k{=}7$ & 8.35 & 4.88 & 82\% \\
16 & Log & BenchReg & $R^2_{\min}{=}0.3$, $k{=}7$ & 8.37 & 5.08 & 79\% \\
17 & Quantile & Soft-Impute & $r{=}2$ & 8.42 & 5.04 & 100\% \\
18 & Arcsinh & ModelReg & $R^2_{\min}{=}0.3$, $k{=}3$ & 8.46 & 5.29 & 69\% \\
19 & Logit & Soft-Impute & $r{=}2$ & 8.49 & 5.10 & 100\% \\
20 & Probit & Soft-Impute & $r{=}2$ & 8.53 & 5.08 & 100\% \\
21 & Arcsinh & Bias ALS & $\lambda{=}0.1$, $r{=}2$ & 8.84 & 5.47 & 100\% \\
22 & Identity & Soft-Impute & $r{=}2$ & 8.89 & 5.40 & 100\% \\
23 & Square root & Bias ALS & $\lambda{=}0.1$, $r{=}2$ & 8.95 & 5.36 & 100\% \\
24 & Probit & Model-KNN & $k{=}10$ & 9.26 & 5.50 & 100\% \\
25 & Log & ModelReg & $R^2_{\min}{=}0.3$, $k{=}3$ & 9.28 & 5.54 & 69\% \\
26 & Identity & Model-KNN & $k{=}10$ & 9.33 & 5.69 & 100\% \\
27 & Logit & Model-KNN & $k{=}10$ & 9.40 & 5.51 & 100\% \\
28 & Quantile & Model-KNN & $k{=}10$ & 9.41 & 5.61 & 100\% \\
29 & Arcsinh & Soft-Impute & $r{=}2$ & 9.55 & 5.74 & 100\% \\
30 & Square root & Soft-Impute & $r{=}2$ & 9.57 & 5.74 & 100\% \\
31 & Quantile & MLP & lr${=}0.001$ & 9.59 & 5.74 & 100\% \\
32 & Square root & Model-KNN & $k{=}10$ & 9.62 & 5.79 & 100\% \\
33 & Arcsinh & Model-KNN & $k{=}10$ & 9.63 & 5.83 & 100\% \\
34 & Logit & MLP & lr${=}0.001$ & 9.65 & 5.73 & 100\% \\
35 & Log & Bias ALS & $\lambda{=}0.1$, $r{=}2$ & 9.68 & 5.72 & 100\% \\
36 & Probit & MLP & lr${=}0.001$ & 9.70 & 5.78 & 100\% \\
37 & Quantile & NMF & $r{=}1$ & 9.76 & 5.95 & 100\% \\
38 & Identity & MLP & lr${=}0.001$ & 9.84 & 6.30 & 100\% \\
39 & Logit & NMF & $r{=}1$ & 10.12 & 6.10 & 100\% \\
40 & Probit & NMF & $r{=}1$ & 10.21 & 6.36 & 100\% \\
41 & Log & Model-KNN & $k{=}10$ & 10.30 & 6.21 & 100\% \\
42 & Arcsinh & MLP & lr${=}0.01$ & 10.42 & 6.52 & 100\% \\
43 & Quantile & Bench-KNN & $k{=}10$ & 10.44 & 6.29 & 100\% \\
44 & Log & Soft-Impute & $r{=}2$ & 10.53 & 6.10 & 100\% \\
45 & Square root & MLP & lr${=}0.001$ & 10.57 & 6.47 & 100\% \\
46 & Probit & Nuclear & $\lambda{=}5.0$ & 10.95 & 6.82 & 100\% \\
47 & Logit & Nuclear & $\lambda{=}5.0$ & 10.97 & 6.71 & 100\% \\
48 & Quantile & Nuclear & $\lambda{=}5.0$ & 11.10 & 6.88 & 100\% \\
49 & Logit & Bench-KNN & $k{=}10$ & 11.15 & 6.65 & 100\% \\
50 & Identity & NMF & $r{=}1$ & 11.26 & 7.22 & 100\% \\
51 & Probit & Bench-KNN & $k{=}10$ & 11.27 & 6.69 & 100\% \\
52 & Identity & Nuclear & $\lambda{=}5.0$ & 11.34 & 7.48 & 100\% \\
53 & Log & MLP & lr${=}0.01$ & 11.35 & 6.76 & 100\% \\
54 & Quantile & Model-Mean & --- & 11.62 & 7.21 & 100\% \\
55 & Identity & Bench-KNN & $k{=}10$ & 11.76 & 7.29 & 100\% \\
56 & Logit & Model-Mean & --- & 11.83 & 7.22 & 100\% \\
57 & Probit & Model-Mean & --- & 11.95 & 7.38 & 100\% \\
58 & Square root & NMF & $r{=}1$ & 11.96 & 7.32 & 100\% \\
59 & Arcsinh & NMF & $r{=}1$ & 11.97 & 7.70 & 100\% \\
60 & Arcsinh & Nuclear & $\lambda{=}5.0$ & 12.05 & 7.79 & 100\% \\
61 & Square root & Nuclear & $\lambda{=}5.0$ & 12.14 & 7.67 & 100\% \\
62 & Square root & Bench-KNN & $k{=}10$ & 12.16 & 7.33 & 100\% \\
63 & Arcsinh & Bench-KNN & $k{=}10$ & 12.17 & 7.45 & 100\% \\
64 & Quantile & PMF & $r{=}5$ & 12.54 & 7.82 & 100\% \\
65 & Identity & Model-Mean & --- & 12.61 & 8.19 & 100\% \\
66 & Log & NMF & $r{=}1$ & 12.70 & 7.66 & 100\% \\
67 & Probit & PMF & $r{=}5$ & 12.91 & 8.19 & 100\% \\
68 & Logit & PMF & $r{=}3$ & 13.07 & 8.16 & 100\% \\
69 & Log & Bench-KNN & $k{=}10$ & 13.09 & 7.65 & 100\% \\
70 & Log & Nuclear & $\lambda{=}5.0$ & 13.20 & 8.10 & 100\% \\
71 & Square root & Model-Mean & --- & 13.33 & 8.29 & 100\% \\
72 & Arcsinh & Model-Mean & --- & 13.36 & 8.53 & 100\% \\
73 & Identity & PMF & $r{=}5$ & 13.54 & 9.04 & 100\% \\
74 & Log & Model-Mean & --- & 14.14 & 8.44 & 100\% \\
75 & Arcsinh & PMF & $r{=}3$ & 14.42 & 9.36 & 100\% \\
76 & Square root & PMF & $r{=}3$ & 14.53 & 9.26 & 100\% \\
77 & Quantile & Bench-Mean & --- & 15.26 & 9.65 & 100\% \\
78 & Logit & Bench-Mean & --- & 15.56 & 10.04 & 100\% \\
79 & Probit & Bench-Mean & --- & 15.70 & 10.21 & 100\% \\
80 & Log & PMF & $r{=}3$ & 15.85 & 9.61 & 100\% \\
81 & Identity & Bench-Mean & --- & 16.29 & 11.04 & 100\% \\
82 & Arcsinh & Bench-Mean & --- & 17.06 & 11.40 & 100\% \\
83 & Square root & Bench-Mean & --- & 17.28 & 11.18 & 100\% \\
84 & Log & Bench-Mean & --- & 18.72 & 11.56 & 100\% \\
\end{longtable}

}

\subsection{\benchpress{} vs. LLMs as Benchmark Score Predictors}
\label{app:llm_five_shot_prompt}

The LLM diagnostic in \Cref{sec:additional_llm_baseline} uses no separate system prompt: the API call passes \texttt{system\_message=None}, and all task instructions are contained in the user message.
The user prompt is generated once per batch of target cells.
In the named condition, model and benchmark fields use the real names and the benchmark line also includes the benchmark scale.
In the blind condition, those fields are replaced by local labels such as \texttt{Target model q0}, \texttt{Benchmark A}, and \texttt{Peer model q0-1}; scores and the five-shot peer-example structure are unchanged.
Here \texttt{query\_id} is the local identifier for a target cell within the batch, such as \texttt{q0}; it is used only to match the returned JSON value to the corresponding query.

\begin{titlebox}[title={Five-shot LLM user prompt template}, fontupper=\footnotesize\ttfamily]
You are estimating benchmark results before running expensive evaluations.\\
Each query gives compact known scores for a target model and five nearest peer-model examples.\\
Make a quick numerical estimate from the nearest peers; do not explain or show calculations.\\
Return ONLY valid JSON mapping each query id to a numeric score, e.g. \{"q0": 72.5\}.\\
\\
Query \{query id, e.g. q0\}\\
Target model: \{target model name or blind target label\}\\
Target known scores: [\{benchmark label: score\}, ...]\\
Estimate the target model's score on: \{target benchmark name and scale, or blind benchmark label\}\\
Nearest peer examples:\\
Example 1: model=\{peer model name or blind peer label\}; shared\_scores=[\{benchmark label: score\}, ...]; \{target benchmark label\} score: \{peer target score\}\\
...\\
Example 5: model=\{peer model name or blind peer label\}; shared\_scores=[\{benchmark label: score\}, ...]; \{target benchmark label\} score: \{peer target score\}
\end{titlebox}

\subsection{Comparison to Observational Scaling Laws}
\label{app:osl_comparison}

\emph{Observational Scaling Laws} (OSL)~\citep{ruan2024} construct low-dimensional capability measures from a fixed block of standard benchmarks and fit a separate predictor for each selected downstream capability, including under weak-to-strong extrapolation. \benchpress{} instead considers the general setting in which any available benchmark scores may be observed and any remaining score may be the prediction target.

This difference shows up in how much data each method needs. OSL's released matrices are nearly complete: benchmarks miss under 5\% of their entries and every training benchmark carries at least ten model scores, so its per-benchmark regressions always see dense columns. Our matrix is 76.7\% missing, and some benchmarks are observed for as few as two models, so a per-benchmark regression frequently has fewer training models than free parameters. OSL's largest completion task predicts two agentic targets for twenty-two models (their Section~4.2), whereas ours predicts 133 benchmarks for 84 models.

To run OSL on our matrix, we supply the dense capability block it assumes in two ways: (i) restrict to the eight most-observed benchmark columns and impute their remaining gaps before fitting OSL's capability regression; and (ii) impute the full 76.7\%-missing matrix before the same fit. Both variants and \benchpress{} are evaluated with the same 30-fold protocol and the same evaluation harness, and \Cref{tab:osl_comparison} reports the median error over each method's finite predictions on the held-out cells. Even the denser variant~(i) trails \benchpress{}, and imputing the full sparse matrix~(ii) degrades OSL further.

\begin{table}[!htbp]
\centering
\caption{Applying OSL~\citep{ruan2024} to our score matrix, against \benchpress{}. OSL needs a densely scored capability block, supplied either by (i) the eight most-observed benchmark columns or (ii) imputing the full matrix. Median error is computed by the shared evaluation harness on each method's finite predictions under the same 30-fold protocol; lower is better.}
\label{tab:osl_comparison}
\small
\setlength{\tabcolsep}{6pt}
\renewcommand{\arraystretch}{1.2}
\begin{tabular}{@{}lcc@{}}
\toprule
Method & MedAPE (\%) $\downarrow$ & MedAE $\downarrow$ \\
\midrule
OSL, densest-8 block (i) & 9.85 & 5.94 \\
OSL, full-matrix impute (ii) & 21.87 & 12.17 \\
\cBP\textbf{\benchpress{}} & \cBP\textbf{7.77} & \cBP\textbf{4.63} \\
\bottomrule
\end{tabular}
\end{table}

One way to read the gap is that \benchpress{} draws on more of the scores a model already has. OSL summarizes each model with a few capability values fitted on one fixed benchmark block, so observed scores outside that block do not directly enter its prediction, whereas \benchpress{} conditions each estimate on all of that model's observed cells. On a sparse public matrix, whose coverage is uneven across models and benchmarks, this may be part of why \benchpress{} does better here.

The gap does not make OSL a weaker method on its own terms. Where a genuinely stronger model must be predicted from weaker models alone, with no comparably strong peer already in the matrix, OSL's scaling formulation is the better-matched tool; that regime lies outside the general completion problem \benchpress{} targets.

\subsection{\benchpress{} on Other Score Matrices}
\label{app:other_matrices}

\paragraph{Third-party score matrices.}
To check that \benchpress{} is not specific to our curated matrix, we rebuild two independent score matrices, one from the public HELM leaderboard~\citep{liang2023helm} and one from Every Eval Ever (EEE)~\citep{batzner2026everyeval}, using the same construction rules and the same predictor. Each source is curated into a \benchpress{}-compatible matrix, restricted to its percentage-scale benchmarks and filtered to the paper's observation thresholds, giving an $83 \times 207$ HELM matrix ($9{,}776$ cells) and a $653 \times 498$ EEE matrix ($24{,}879$ cells). \Cref{tab:other_matrices} scores \benchpress{} on both against per-benchmark-median and global-median baselines under the paper's per-model holdout: \benchpress{} is the most accurate under both metrics on both matrices, matching the pattern on our own matrix.

\begin{table}[!htbp]
\centering
\caption{\benchpress{} on two independently built score matrices, curated from the HELM leaderboard ($83 \times 207$, $9{,}776$ cells) and Every Eval Ever ($653 \times 498$, $24{,}879$ cells), against per-benchmark-median and global-median baselines under the paper's per-model holdout; lower is better. \benchpress{} is the most accurate on both.}
\label{tab:other_matrices}
\small
\setlength{\tabcolsep}{6pt}
\renewcommand{\arraystretch}{1.2}
\begin{tabular}{@{}lcccc@{}}
\toprule
& \multicolumn{2}{c}{HELM} & \multicolumn{2}{c}{EEE} \\
\cmidrule(lr){2-3}\cmidrule(lr){4-5}
Predictor & MedAE $\downarrow$ & MedAPE $\downarrow$ & MedAE $\downarrow$ & MedAPE $\downarrow$ \\
\midrule
\cBP\textbf{\benchpress{}} & \cBP\textbf{2.87} & \cBP\textbf{10.67\%} & \cBP\textbf{2.89} & \cBP\textbf{5.02\%} \\
Per-benchmark median & 4.12 & 15.31\% & 7.00 & 11.24\% \\
Global median & 19.31 & 46.60\% & 18.25 & 23.18\% \\
\bottomrule
\end{tabular}
\end{table}

\paragraph{Version 2: August 26, 2026 score matrix.}
We also rerun the full 329-configuration method comparison on our updated score matrix using the same filtering thresholds and 10-seed, three-fold per-model holdout. After filtering, this version contains 129 models, 253 benchmarks, and 4{,}905 observed cells. \Cref{tab:updated_score_matrix_top15} shows that the adopted Logit Bias ALS predictor remains fully covered and ranks sixth by MedAPE and first by MedAE. The lowest-MedAPE configuration is only slightly more accurate but covers 79\% of held-out cells, so the updated matrix strengthens the original choice of Logit Bias ALS as the full-coverage predictor.

\begin{table*}[!htbp]
\centering
\caption{Top-15 transform and method configurations on the August 26, 2026 score matrix after applying the paper's observation thresholds ($129 \times 253$, 4{,}905 observed cells). The two halves rank configurations independently by median absolute percentage error (MedAPE) and median absolute error (MedAE); each value is followed by prediction coverage in parentheses. All results report the median over 10 seeds and three folds. Pink highlights mark the Logit Bias ALS configuration adopted as \benchpress{}'s main score predictor.}
\label{tab:updated_score_matrix_top15}
\small
\setlength{\tabcolsep}{2.5pt}
\resizebox{\textwidth}{!}{%
\begin{tabular}{@{}rlllr@{\hspace{12pt}}rlllr@{}}
\toprule
\multicolumn{5}{c}{\textbf{MedAPE (\%) $\downarrow$}} &
\multicolumn{5}{c}{\textbf{MedAE $\downarrow$}} \\
\cmidrule(r){1-5} \cmidrule(l){6-10}
\# & Transform & Method & Hyperparameter & Value &
\# & Transform & Method & Hyperparameter & Value \\
\midrule
1 & Probit & BenchReg & $R^2_{\min}{=}0.3$, $k{=}7$ & 7.6 (79\%) & 1 & \cBP\textbf{Logit} & \cBP\textbf{Bias ALS} & \cBP\textbf{$\lambda{=}0.1$, $r{=}2$} & \cBP\textbf{4.43 (100\%)} \\
2 & Logit & BenchReg & $R^2_{\min}{=}0.3$, $k{=}7$ & 7.6 (79\%) & 2 & Probit & Bias ALS & $\lambda{=}0.1$, $r{=}2$ & 4.47 (100\%) \\
3 & Probit & BenchReg & $R^2_{\min}{=}0.2$, $k{=}7$ & 7.7 (80\%) & 3 & Logit & BenchReg & $R^2_{\min}{=}0.3$, $k{=}7$ & 4.49 (79\%) \\
4 & Logit & BenchReg & $R^2_{\min}{=}0.2$, $k{=}7$ & 7.7 (80\%) & 4 & Logit & Bias ALS & $\lambda{=}0.01$, $r{=}2$ & 4.51 (100\%) \\
5 & Probit & BenchReg & $R^2_{\min}{=}0.1$, $k{=}7$ & 7.7 (81\%) & 5 & Logit & BenchReg & $R^2_{\min}{=}0.2$, $k{=}7$ & 4.54 (80\%) \\
6 & \cBP\textbf{Logit} & \cBP\textbf{Bias ALS} & \cBP\textbf{$\lambda{=}0.1$, $r{=}2$} & \cBP\textbf{7.7 (100\%)} & 6 & Probit & BenchReg & $R^2_{\min}{=}0.3$, $k{=}7$ & 4.55 (79\%) \\
7 & Probit & Bias ALS & $\lambda{=}0.1$, $r{=}2$ & 7.7 (100\%) & 7 & Logit & BenchReg & $R^2_{\min}{=}0.1$, $k{=}7$ & 4.56 (80\%) \\
8 & Logit & BenchReg & $R^2_{\min}{=}0.3$, $k{=}5$ & 7.8 (72\%) & 8 & Probit & BenchReg & $R^2_{\min}{=}0.2$, $k{=}7$ & 4.56 (80\%) \\
9 & Quantile & Bias ALS & $\lambda{=}0.1$, $r{=}2$ & 7.8 (100\%) & 9 & Probit & BenchReg & $R^2_{\min}{=}0.1$, $k{=}7$ & 4.56 (81\%) \\
10 & Logit & BenchReg & $R^2_{\min}{=}0.1$, $k{=}7$ & 7.8 (80\%) & 10 & Probit & Bias ALS & $\lambda{=}0.01$, $r{=}2$ & 4.57 (100\%) \\
11 & Logit & BenchReg & $R^2_{\min}{=}0.1$, $k{=}5$ & 7.8 (74\%) & 11 & Quantile & Bias ALS & $\lambda{=}0.1$, $r{=}2$ & 4.61 (100\%) \\
12 & Logit & Bias ALS & $\lambda{=}0.01$, $r{=}2$ & 7.8 (100\%) & 12 & Probit & Bias ALS & $\lambda{=}1.0$, $r{=}2$ & 4.62 (100\%) \\
13 & Probit & Bias ALS & $\lambda{=}0.01$, $r{=}2$ & 7.8 (100\%) & 13 & Quantile & Bias ALS & $\lambda{=}0.01$, $r{=}2$ & 4.64 (100\%) \\
14 & Logit & BenchReg & $R^2_{\min}{=}0.2$, $k{=}5$ & 7.8 (73\%) & 14 & Logit & Bias ALS & $\lambda{=}1.0$, $r{=}2$ & 4.65 (100\%) \\
15 & Identity & BenchReg & $R^2_{\min}{=}0.2$, $k{=}7$ & 7.8 (79\%) & 15 & Logit & BenchReg & $R^2_{\min}{=}0.3$, $k{=}5$ & 4.66 (72\%) \\
\bottomrule
\end{tabular}%
}
\end{table*}

\section{Supplemental to \texorpdfstring{\Cref{sec:findings}}{Section 5}: What \benchpress{} Enables for Model Evaluation}
\label{app:evaluation_design}

This section provides additional details for the model-evaluation analyses in \Cref{sec:findings}.
\Cref{app:probe_selection} supplements \Cref{sec:probe_selection} with a pairwise-redundancy diagnostic and more exhaustive probe-selection checks.
\Cref{app:ranking_preservation} reports an auxiliary shortlist-recovery metric for \Cref{sec:ranking_preservation}.
\Cref{app:temporal_deployment} gives the full per-target table for \Cref{sec:temporal_deployment}.

\subsection{Budgeted Scorecard Recovery}
\label{app:probe_selection}

This appendix supplements \Cref{sec:probe_selection} in two ways.
First, a pairwise-redundancy diagnostic explains why a small probe set can recover most of the matrix: the typical benchmark already has another benchmark column that predicts it nearly perfectly.
Second, a more exhaustive probe-selection analysis asks whether greedy's computational simplicity comes at a material accuracy cost, comparing it with exact enumeration over the low-cost allowlist and exact search after pruning in the unrestricted setting.

\paragraph{Widespread redundancy across benchmarks.}
\label{app:pairwise_redundancy}
Before choosing a probe set, we first ask whether many benchmark columns are redundant enough that a small probe set could plausibly recover the rest.
For every ordered pair of benchmarks $(b, b')$, we collect the scores $s_{mb}$ and $s_{mb'}$ of all models $m$ evaluated on both ($n \geq 5$ shared models required), apply a logit transform followed by per-column $z$-scoring, fit a univariate linear regression in this transformed space, and invert the transform to obtain predicted raw scores $\hat{s}_{mb}$.
For each target benchmark $b$, we identify its \emph{best predictor benchmark}, the predictor $b'$ that maximizes the absolute Pearson correlation, and report the signed correlation, $\mathsf{MedAE}$, and $\mathsf{MedAPE}$.
Of the 133 benchmarks, 132 have at least one neighbor pair with $\geq 5$ shared models; 1 is excluded for insufficient overlap.
Among these 132, 127 have a best-neighbor absolute correlation $\geq 0.85$ (129 reach $\geq 0.80$), and the median best-neighbor absolute correlation is $0.97$.
\Cref{tab:pairwise_ols} lists the five most and five least predictable benchmarks.
The most predictable pair, the LongFact Concepts and LongFact Objects hallucination-rate benchmarks, achieves a correlation of $0.997$.
At the other extreme, Safety (OLMES suite) has the weakest best-neighbor correlation ($0.62$), followed by MRCR v1 ($0.68$).

\begin{table}[!htbp]
\centering
\caption{Five most and least predictable benchmarks identified by pairwise linear regression in logit\,+\,$z$-score space. Rows are selected by absolute Pearson correlation between the target and its best predictor benchmark; the Corr. column reports the signed value.}
\label{tab:pairwise_ols}
\small
\setlength{\tabcolsep}{3pt}
\resizebox{.8\textwidth}{!}{%
\begin{tabular}{@{}llrrr@{}}
\toprule
\textbf{Bench.} & \textbf{Best neighbor} & \textbf{Corr.} & \textbf{MedAPE\,$\downarrow$} & \textbf{MedAE\,$\downarrow$} \\
\midrule
\multicolumn{5}{@{}l}{\textit{Highly predictable (highest correlation)}} \\
LongFact-Concepts & LongFact-Objects & .997 & 5.4\% & 0.37 \\
LongFact-Objects & LongFact-Concepts & .997 & 7.9\% & 0.33 \\
Beyond AIME & BRUMO 2025 & .994 & 1.9\% & 1.56 \\
BRUMO 2025 & Beyond AIME & .994 & 0.3\% & 0.29 \\
MathVista & MMMU & .992 & 3.9\% & 2.64 \\
\midrule
\multicolumn{5}{@{}l}{\textit{Hard to predict (lowest correlation)}} \\
Safety (OLMES suite) & HumanEval & .622 & 8.1\% & 5.81 \\
MRCR v1 & SimpleQA & .676 & 6.7\% & 4.86 \\
HealthBench & GraphWalks parents 0--128K & -.789 & 4.9\% & 2.75 \\
OmniDocBench 1.5 & GDPval (Artificial Analysis ELO) & .815 & 131.9\% & 2.42 \\
ComplexFuncBench & tau-bench Airline & .846 & 27.1\% & 11.41 \\
\bottomrule
\end{tabular}%
}
\end{table}

\textit{Caveat: high cross-category correlation does not imply semantic similarity.}
Some cross-category pairs appear surprisingly predictable; for example, GDPval (Artificial Analysis ELO) and WideSearch have correlation $0.99$.
This does not mean GDP-style task performance predicts search-agent performance.
The regression is fit on only 5 models that have scores on both benchmarks, all of which are frontier models whose scores are dominated by a single general-capability axis: whichever model is strongest overall tends to score highest on both.
With so few data points and so little capability diversity, nearly any two benchmarks will correlate.
These inflated cross-category correlation values reflect sample composition, not a meaningful relationship between the benchmarks.

\begin{finding}{Most benchmark scores are inferable from one well-chosen peer, reflecting shared variation across the matrix.}{redundancy}
\end{finding}

\paragraph{Pruned exhaustive probe selection.}
The budgeted scorecard recovery experiment in \Cref{sec:probe_selection} selects probe sets greedily.
Greedy selection is simple and fast, but it does not certify that the selected five probes are the best possible subset.
We therefore run two exhaustive checks, one where exact enumeration is feasible and one where the unrestricted search space first has to be pruned.
\begin{itemize}[leftmargin=*]
    \item \textit{Cost-aware exact exhaustive search.}
    The low-cost allowlist has only 25 candidate probes, so exact enumeration is feasible without pruning.
    We search all $\binom{25}{5}=53{,}130$ five-probe subsets to test whether the cost-aware greedy prefix misses a better cheap combination.
    \item \textit{Cost-unaware pruned exhaustive search.}
    The unrestricted setting has 133 candidate probes, making exact search over all $\binom{133}{5}$ five-probe subsets too large.
    We therefore build a top-30 candidate pool from the full ten-step MedAE greedy trajectory: at each step, every remaining candidate is ranked by the pooled MedAE it would achieve if added next, and ranks are averaged across steps.
    This pruning keeps the full MedAE greedy top-10 prefix, reduces the exact search to $\binom{30}{5}=142{,}506$ subsets, and lets us test whether the greedy five-probe solution is preserved when the final subset is selected exactly.
\end{itemize}
The results are reported in \Cref{tab:probe_exhaustive_check}.
In the unrestricted setting, pruned exhaustive search returns the same five probes as greedy.
In the low-cost setting, exact exhaustive search improves slightly over greedy, but the gain is small; the greedy probe sets are already close to optimal under both policies.

\begin{table}[!htbp]
\centering
\caption{\textbf{Five-probe MedAE selections.} Cost-unaware greedy already matches the best five-probe set found by exact search over the pruned top-30 universe. In the low-cost allowlist, exhaustive search slightly improves over cost-aware greedy.}
\label{tab:probe_exhaustive_check}
\scriptsize
\setlength{\tabcolsep}{2.5pt}
\resizebox{\textwidth}{!}{%
\begin{tabular}{@{}llp{0.52\textwidth}rr@{}}
\toprule
Policy & Selection method & Five-probe set & MedAE$\downarrow$ & MedAPE$\downarrow$ \\
\midrule
\multirow{2}{*}{Cost-unaware} & Greedy & GPQA Diamond; HLE; Codeforces Rating; MMLU-Pro; ARC-AGI-1 & 3.926 & 6.588 \\
 & Pruned exhaustive & GPQA Diamond; HLE; Codeforces Rating; MMLU-Pro; ARC-AGI-1 & 3.926 & 6.588 \\
 \midrule
\multirow{2}{*}{Cost-aware} & Greedy & GPQA Diamond; MMLU-Pro; Aider Polyglot (diff mode); MATH-500; AIME 2026 & 4.549 & 7.596 \\
 & Exact exhaustive & $\tau^2$-bench Telecom; MathArena Apex 2025; GPQA Diamond; Aider Polyglot (diff mode); MMLU-Pro & 4.465 & 7.412 \\
\bottomrule
\end{tabular}}
\end{table}

\subsection{Preserving Model Rankings}
\label{app:ranking_preservation}

\paragraph{Auxiliary metric: shortlist recovery.}
\Cref{sec:ranking_preservation} reports same-benchmark pairwise ranking accuracy as the main ranking metric.
As an auxiliary view, we also measure shortlist recovery.
For each benchmark and held-out fold, we complete the full observed leaderboard by keeping seen scores fixed and replacing held-out cells with \benchpress{} predictions.
We then compare the completed top fraction with the true top fraction on that same observed leaderboard.
Because the predicted and true shortlists have the same size, the overlap rate is the fraction of true top models recovered by the predicted shortlist.
\Cref{tab:ranking_shortlist} reports two summaries of this overlap: \emph{recovery} computes top-fraction recovery separately for each benchmark and then reports the median across benchmarks, while \emph{shortlist slots} counts selection positions rather than unique models, so a benchmark-fold group contributing four top-$20\%$ positions contributes four slots.

\begin{table}
\vspace{-0.8em}
\centering
\scriptsize
\setlength{\tabcolsep}{2.0pt}
\caption{\textbf{Auxiliary shortlist recovery.}
Recovery is median benchmark-level overlap; slots count top-fraction positions.}
\label{tab:ranking_shortlist}
\vspace{-0.3em}
\resizebox{.3\linewidth}{!}{%
\begin{tabular}{@{}lcc@{}}
\toprule
Setting & Recovery & Slots \\
\midrule
Top 10\% & $72.4\%$ & $9{,}515$ \\
Top 20\% & $79.3\%$ & $17{,}154$ \\
Top 30\% & $83.9\%$ & $25{,}091$ \\
\bottomrule
\end{tabular}%
}
\vspace{-1.0em}
\end{table}

\paragraph{Probe selection for ranking preservation.}
We also run a probe-selection diagnostic that optimizes the ranking metric directly.
This greedy search evaluates the same set of observed model--benchmark cells as \Cref{sec:probe_selection}, but scores each candidate prefix by same-benchmark pairwise ranking accuracy with a five-point score margin.
Probe cells are revealed exactly and remain in the denominator, so the question is which known benchmark scores most improve the completed leaderboard.
\Cref{tab:ranking_greedy_probe_set} reports two top-10 prefixes selected by this ranking-aware objective: a cost-unaware search that may choose any benchmark, and a cost-aware search restricted by the low-cost allowlist.
The cost-aware constraint costs only a small amount of ranking accuracy at ten probes ($86.2\%$ versus $88.9\%$) while selecting a more practical benchmark set.
\begin{table}[!htbp]
\centering
\small
\setlength{\tabcolsep}{2.0pt}
\resizebox{\textwidth}{!}{%
\begin{tabular}{clclcc}
\toprule
Step & Cost-unaware probe & Accuracy & Cost-aware probe & Accuracy & Comparable pairs \\
\midrule
1 & GPQA Diamond & $78.9\%$ & GPQA Diamond & $78.9\%$ & $27{,}245$ \\
2 & BrowseComp & $82.5\%$ & MMLU-Pro & $81.7\%$ & $27{,}245$ \\
3 & Codeforces Rating & $84.5\%$ & AIME 2025 & $83.0\%$ & $27{,}245$ \\
4 & DROP & $85.6\%$ & Bullshit-Bench (Clear Pushback) & $83.8\%$ & $27{,}245$ \\
5 & ScreenSpot-Pro & $86.3\%$ & HMMT Feb 2026 & $84.5\%$ & $27{,}245$ \\
6 & GraphWalks parents 0--128K & $87.1\%$ & MATH-500 & $84.9\%$ & $27{,}245$ \\
7 & AIME 2024 & $87.7\%$ & AlpacaEval 2.0 (LC-winrate) & $85.7\%$ & $27{,}245$ \\
8 & MGSM & $88.2\%$ & HMMT Feb 2025 & $86.1\%$ & $27{,}245$ \\
9 & NL2Repo-Bench & $88.4\%$ & Aider Polyglot (whole mode) & $86.1\%$ & $27{,}245$ \\
10 & AlpacaEval 2.0 (LC-winrate) & $88.9\%$ & Arena-Hard Auto & $86.2\%$ & $27{,}245$ \\
\bottomrule
\end{tabular}%
}
\caption{\textbf{Top-10 probe sets selected for ranking preservation.}
Both greedy searches optimize same-benchmark pairwise ranking accuracy at a five-point score margin.
The cost-unaware search may choose any benchmark; the cost-aware search is restricted by the low-cost allowlist.
At each step, the selected probe prefix is evaluated on the same fixed universe of observed cells; revealed probe cells are exact and unrevealed observed cells are predicted by \benchpress{}.}
\label{tab:ranking_greedy_probe_set}
\end{table}

\subsection{Predicting Newly Released Models}
\label{app:temporal_deployment}

The main text summarizes the temporal deployment stress test as a distribution across target models.
\Cref{tab:temporal_deployment_full} lists the per-target results.
Each target model is selected by the same pre-specified temporal-window rule used in \Cref{sec:temporal_deployment}: we use models from the post-DeepSeek-R1 reasoning era through GPT-5.1 and keep models with more than 20 observed scores in the final matrix.

\begin{table*}[!htbp]
\centering
\scriptsize
\setlength{\tabcolsep}{2.0pt}
\caption{\textbf{Full temporal deployment results.}
For each target model, Obs. is the number of observed benchmark scores in the final matrix and Train is the number of older models available before the target's release date. Each $k$ column reveals that many seed scores from the target model and predicts the rest; numbers are medians across 10 random seeds.}
\label{tab:temporal_deployment_full}
\resizebox{\textwidth}{!}{%
\begin{tabular}{@{}l r r rr rr rr@{}}
\toprule
& & & \multicolumn{2}{c}{$k = 1$} & \multicolumn{2}{c}{$k = 5$} & \multicolumn{2}{c}{$k = 10$} \\
\cmidrule(lr){4-5} \cmidrule(lr){6-7} \cmidrule(lr){8-9}
Target model & Obs. & Train & MedAPE & MedAE & MedAPE & MedAE & MedAPE & MedAE \\
\midrule
DeepSeek-R1                    & 40 & 23 & 22.9 & 12.0 & 11.8 & 7.2 & 5.1 & 3.7 \\
o3-mini (high)                 & 34 & 25 & 20.3 & 9.2 & 9.3 & 5.9 & 4.5 & 3.6 \\
Grok 3 Beta                    & 21 & 27 & 11.6 & 9.2 & 7.2 & 5.7 & 0.9 & 0.5 \\
Claude 3.7 Sonnet              & 25 & 28 & 25.0 & 12.4 & 14.5 & 7.5 & 3.4 & 1.3 \\
GPT-4.5                        & 28 & 29 & 13.8 & 6.0 & 8.4 & 4.4 & 3.1 & 1.3 \\
QwQ-32B                        & 23 & 30 & 12.9 & 9.9 & 6.6 & 5.0 & 2.0 & 1.8 \\
Llama 4 Maverick               & 22 & 33 & 10.0 & 6.4 & 6.4 & 4.1 & 1.6 & 1.2 \\
GPT-4.1                        & 48 & 34 & 14.8 & 7.3 & 6.2 & 3.3 & 5.1 & 2.4 \\
GPT-4.1 mini                   & 43 & 34 & 15.6 & 6.2 & 9.4 & 4.6 & 7.9 & 3.3 \\
GPT-4.1 nano                   & 35 & 34 & 36.1 & 8.6 & 16.1 & 4.8 & 16.4 & 5.4 \\
o3 (high)                      & 44 & 37 & 22.5 & 10.2 & 13.0 & 5.9 & 8.3 & 4.8 \\
o4-mini (high)                 & 45 & 37 & 22.2 & 10.4 & 12.9 & 6.1 & 9.5 & 4.3 \\
Qwen3-235B-A22B                & 34 & 39 & 9.8 & 6.9 & 8.4 & 5.6 & 5.7 & 3.7 \\
Gemini 2.5 Flash               & 43 & 43 & 26.8 & 11.8 & 8.0 & 4.7 & 6.3 & 2.3 \\
Claude Opus 4                  & 23 & 44 & 24.9 & 10.6 & 8.4 & 3.3 & 1.7 & 0.7 \\
Claude Sonnet 4                & 39 & 44 & 26.2 & 14.1 & 12.8 & 5.6 & 6.1 & 3.8 \\
DeepSeek-R1-0528               & 24 & 46 & 21.9 & 13.7 & 5.8 & 4.1 & 2.6 & 1.5 \\
Gemini 2.5 Pro (GA)            & 65 & 47 & 13.7 & 7.1 & 11.5 & 5.3 & 7.5 & 4.2 \\
Grok 4                         & 24 & 48 & 16.2 & 9.3 & 5.0 & 4.8 & 1.8 & 1.5 \\
GPT-5 mini                     & 30 & 49 & 11.3 & 6.1 & 7.2 & 4.3 & 3.7 & 2.6 \\
GPT-5 nano                     & 30 & 49 & 13.4 & 4.3 & 9.2 & 2.9 & 5.0 & 1.9 \\
Kimi K2                        & 45 & 51 & 15.5 & 7.5 & 13.0 & 6.1 & 7.6 & 3.6 \\
GPT-5                          & 67 & 52 & 23.4 & 14.0 & 9.0 & 4.7 & 7.3 & 4.2 \\
GLM-4.6                        & 27 & 55 & 12.3 & 6.2 & 8.3 & 4.6 & 4.0 & 2.1 \\
Claude Sonnet 4.5              & 79 & 56 & 21.0 & 9.1 & 13.3 & 7.0 & 10.1 & 5.7 \\
MiniMax-M2                     & 28 & 57 & 10.0 & 5.2 & 7.5 & 3.8 & 3.8 & 1.9 \\
GPT-5.1                        & 37 & 59 & 13.7 & 9.9 & 7.6 & 5.5 & 4.4 & 3.1 \\
\midrule
\textit{Median}          &    &    & 15.6 & 9.2 & 8.4 & 4.8 & 5.0 & 2.6 \\
\bottomrule
\end{tabular}%
}
\end{table*}

\section{Supplemental to \texorpdfstring{\Cref{sec:trust}}{Section 6}: When to Trust \benchpress{}'s Predictions}
\label{app:trust}

This section provides additional details for the trust analyses in \Cref{sec:trust}.
\Cref{app:reliability_analysis} supplements \Cref{sec:reliability_analysis} with expanded prediction-error diagnostics.
\Cref{app:confidence_calibration_details} spells out the reliability estimators used in \Cref{sec:confidence_calibration}.

\subsection{What Affects Prediction Reliability}
\label{app:reliability_analysis}

This subsection provides extended experimental analysis that complements the prediction-error analysis in \Cref{sec:reliability_analysis}.
\Cref{app:bench_analysis} extends the benchmark analysis of \Cref{sec:bench_predict} with the full benchmark-side hypothesis grid.
\Cref{app:model_analysis} extends the model analysis of \Cref{sec:model_predict} with the full model-side hypothesis grid.

\paragraph{Spearman rank correlation tests.}
\label{app:reliability_stat_tests}
\Cref{sec:reliability_analysis} uses two hypothesis-test families.
For observational hypotheses, each target contributes a single pair $(x_i, y_i)$: a feature $x_i$ that we measure but cannot intervene on (e.g.\ a benchmark's inherent rank-2 reconstruction quality, or a model's median observed score), and its \benchpress{} prediction error $y_i$.
We use the Spearman rank correlation test to ask whether targets with a higher feature value tend to have higher or lower prediction error.
We measure this association via the \emph{rank} of each value, its position in the sorted ordering of its column, where the smallest value has rank $1$ and the largest has rank $n$ (this sense of ``rank'' is unrelated to the matrix-rank quantity used in \Cref{sec:bp_svd}).
Using ranks rather than raw values makes the test sensitive to monotonic relationships, not only linear ones, and more robust to heavy-tailed errors and outliers.
Concretely, we replace each column by its within-column ranks and compute the Pearson correlation between the two ranked columns; the result $\rho \in [-1, 1]$ is the Spearman correlation.
Intuitively, $\rho > 0$ means targets with a higher feature value tend to also have higher prediction error, $\rho < 0$ means the opposite, and $|\rho|$ measures how consistently the ranking holds.
The $p$-value asks: if there were truly no monotonic association ($\rho_{\text{true}} = 0$), how often would we see a sample correlation at least as extreme as $\rho$?
It is computed from the $t$-approximation $t = \rho \sqrt{(n-2)/(1-\rho^2)}$, which under $H_0$ approximately follows a Student-$t$ distribution with $n-2$ degrees of freedom \citep{hollander2014nonparametric}.
The approximation is reliable when $n$ is at least a few dozen (all our targets satisfy $n \geq 25$).
Since we test for any deviation from $\rho = 0$, the $p$-value doubles the upper-tail probability, $p = 2\bigl(1 - F_{t_{n-2}}(|t|)\bigr)$, where $F_{t_{n-2}}$ is the CDF of the Student-$t_{n-2}$ distribution.
The intuition is: $|t|$ increases with both $|\rho|$ and the sample size $n$, so the $p$-value gets smaller only when the correlation is meaningful and backed by enough targets.

\paragraph{Paired Wilcoxon signed-rank tests.}
For intervention-style hypotheses, each target contributes a pair of errors $(y^{\text{baseline}}_i, y^{\text{intervention}}_i)$, measured on the same target under two different settings (e.g.\ \benchpress{} trained on the original matrix vs.\ on a matrix where every benchmark highly correlated with the target has been masked out).
We use the paired Wilcoxon signed-rank test to ask whether the intervention shifts each target's error in a consistent direction.
Because the comparison is within-target, each target serves as its own control, removing the effect of inherent target difficulty.
We form per-target differences $\Delta_i = y^{\text{intervention}}_i - y^{\text{baseline}}_i$ and ask whether their median is zero (i.e.\ the intervention has no typical effect on prediction error).
The Wilcoxon signed-rank test ranks $|\Delta_i|$ from smallest to largest, denotes the rank of pair $i$ by $R_i$, and uses as its statistic the sum of ranks for positive differences, $W^+ = \sum_{i:\,\Delta_i > 0} R_i$ (in the rare case of any $\Delta_i = 0$, that pair is dropped before ranking).
Under $H_0$ (the distribution of $\Delta$ is symmetric about $0$), $W^+$ has mean $n(n+1)/4$ and variance $n(n+1)(2n+1)/24$, and the standardized statistic $z = (W^+ - \mathbb{E}[W^+]) / \sqrt{\mathrm{Var}(W^+)}$ is approximately standard normal for $n \gtrsim 25$ \citep{hollander2014nonparametric}.
Since we test for any deviation from $\operatorname{median}\Delta = 0$, the $p$-value doubles the upper-tail probability, $p = 2\bigl(1 - \Phi(|z|)\bigr)$, where $\Phi$ is the standard normal CDF.
The same intuition applies: $|z|$ grows with both the magnitude of the per-target shift and the sample size $n$, and only the combination of a substantial intervention effect and enough paired targets drives $|z|$ large enough, and the $p$-value small enough, to rule out chance.
We use Wilcoxon rather than a paired $t$-test because $\Delta$ is heavy-tailed and not approximately Gaussian across targets.

\paragraph{Multiple-testing correction.}
Across the two hypothesis tables we test $16$ hypotheses in total: seven benchmark-side (H1--H7 in \Cref{tab:predictability_factors}) and nine model-side (H1--H9 in \Cref{tab:model_hypotheses}).
Each hypothesis is tested under both of our error metrics, MedAPE and MedAE, and we report a $p$-value for each, so the two tables together contain $16 \times 2 = 32$ tests.
Because this is a multiple-comparison setting, we assess significance against a Benjamini-Hochberg false-discovery-rate correction \citep{benjamini1995} over all $32$ $p$-values, which raises the bar in proportion to the number of tests.
Ordering the $p$-values and comparing the $k$-th smallest against $(k/32)\,\alpha$ at $\alpha=0.01$, the procedure rejects the $17$ tests with $p\le 0.004$.
These include both the MedAPE and MedAE tests for each of the eight factors we report as jointly supported (benchmark-side H3, H4, H5 and model-side H2, H3, H5, H8, H9), so all eight remain significant under both metrics after correction; no hypothesis that was not already rejected at $p<0.01$ becomes significant.

\subsubsection{Benchmark Analysis}
\label{app:bench_analysis}

This appendix reports two extensions of the benchmark-side analysis in \Cref{sec:bench_predict}: the full $7 \times 2$ hypothesis $\times$ metric grid, and a per-benchmark predictability ranking that names which benchmarks are easiest and hardest for \benchpress{} to predict.

\begin{figure}[p]
\centering
\includegraphics[width=\textwidth]{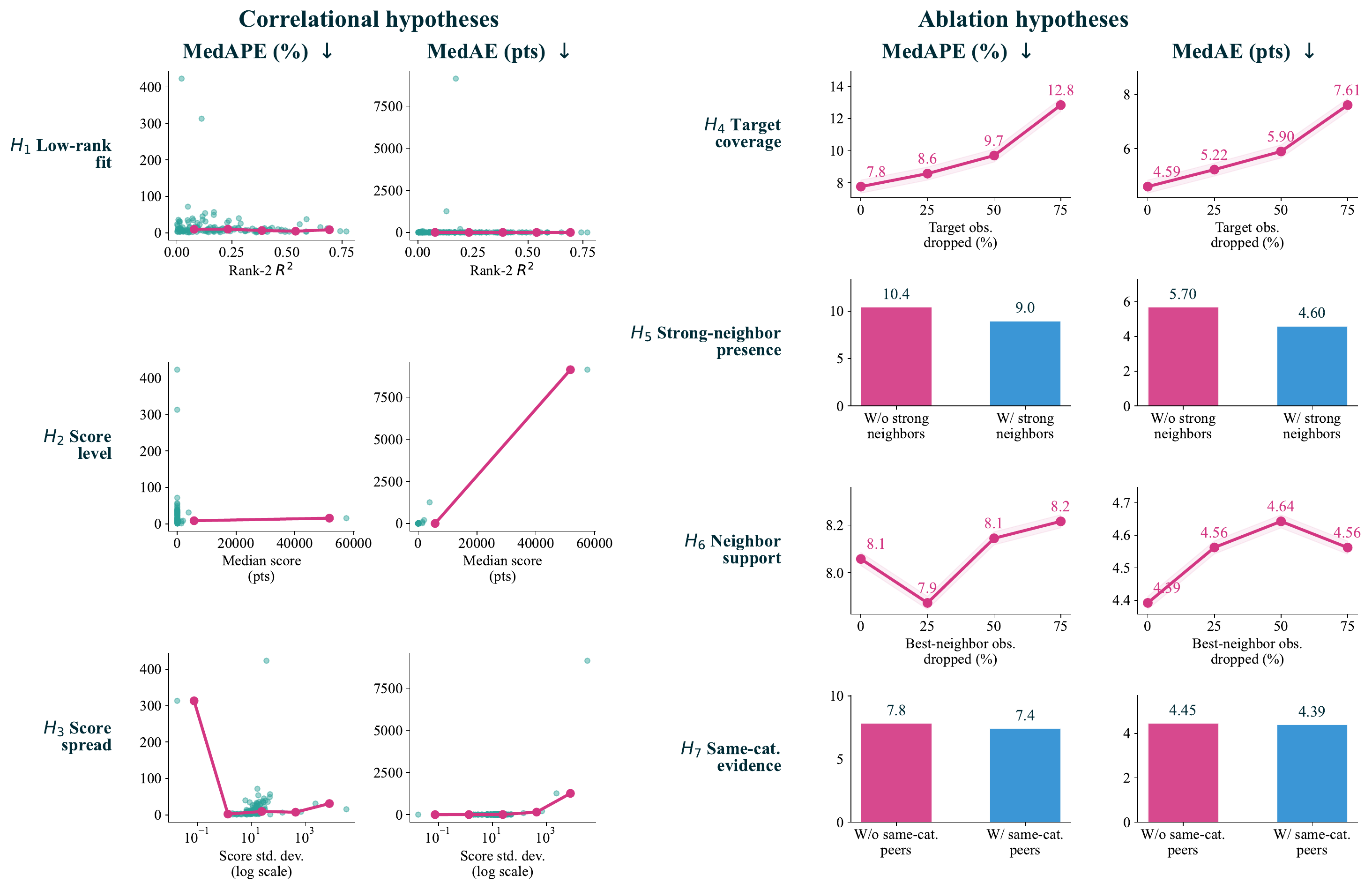}
\caption{\textbf{All seven active benchmark-level hypotheses against both score-error metrics.} The left block shows H1--H3 (correlational hypotheses) and the right block shows H4--H7 (ablations). Columns within each block are $\mathsf{MedAPE}$ ($\downarrow$) and $\mathsf{MedAE}$ ($\downarrow$). Correlational rows show scatter + binned trend; ablation rows show line plots across drop fractions (H4, H6) or paired bars at the headline intervention (H5 with $|r|\!\geq\!0.85$ peers; H7 with same-category peers).}
\label{fig:predictability_factors_full}
\end{figure}

\paragraph{Full hypothesis $\times$ metric grid.}
\label{app:predictability_factors_full}
\Cref{fig:predictability_factors_51} in the main text visualizes the benchmark-side patterns that pass the joint-support criterion (H3, H4, and H5). For completeness, \Cref{fig:predictability_factors_full} expands this to the full $7\times 2$ grid: every active benchmark-side hypothesis (H1--H7) against both score-error metrics, using the same correlational and ablation setups as \Cref{tab:predictability_factors}.

\paragraph{Per-benchmark predictability.}
\label{app:per_benchmark_predictability}
We apply a direct cell-holdout test for each benchmark column.
For each model we randomly hide half of its observed scores and predict them via the \benchpress{} predictor from \Cref{sec:method_comparison}; we then aggregate the test-cell errors by benchmark column. This is repeated over 10 random seeds for stability.

\begin{figure}[p]
\centering
\includegraphics[height=0.72\textheight,width=\textwidth,keepaspectratio]{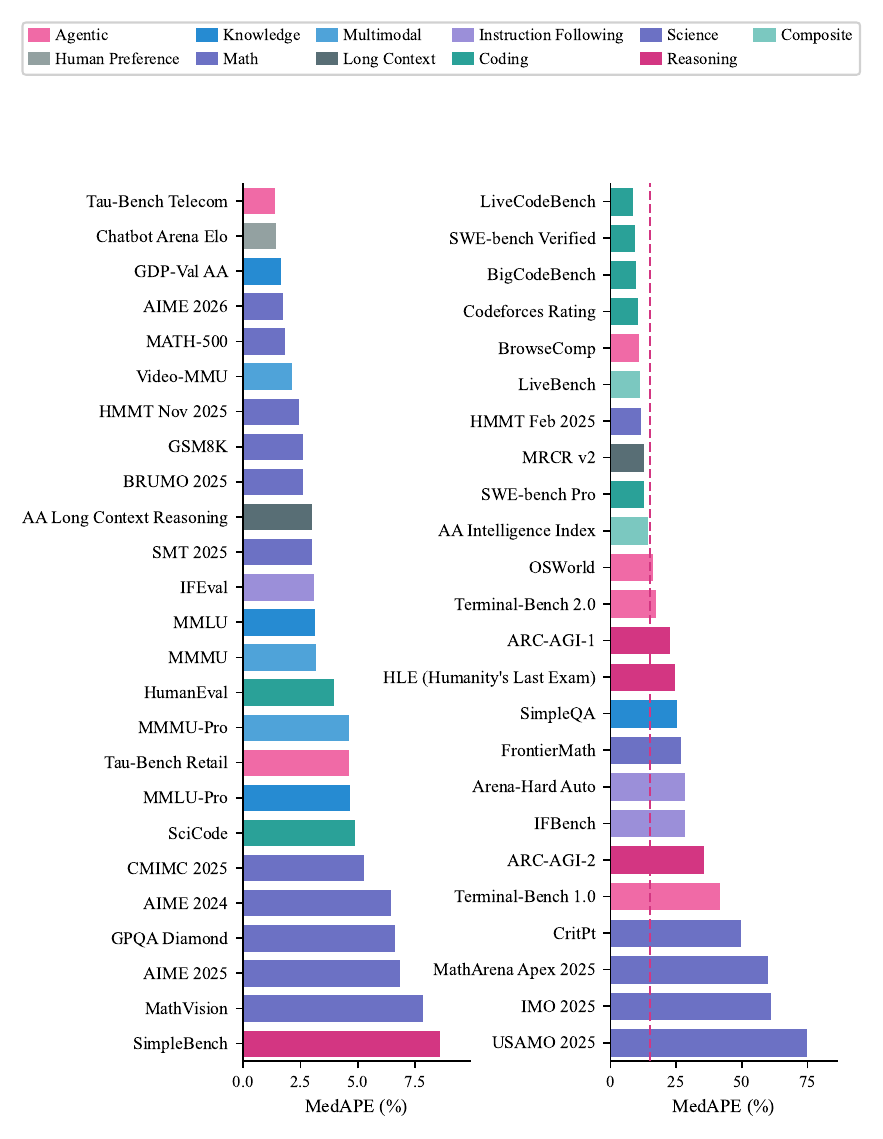}
\caption{\textbf{Per-benchmark predictability.} For each model, half of observed scores are held out and predicted via \benchpress{}; errors are aggregated by benchmark column (10 seeds). Benchmarks below the 15\% $\mathsf{MedAPE}$ threshold (dashed red line) are well-predicted by others. Color = benchmark category.}
\label{fig:benchmark_predictability}
\end{figure}

\Cref{fig:benchmark_predictability} shows the per-benchmark $\mathsf{MedAPE}$ for the evaluated benchmark columns.
Roughly 71\% (35/49) of benchmarks fall below the 15\% $\mathsf{MedAPE}$ threshold, indicating they are inferable with limited additional information loss.

\subsubsection{Model Analysis}
\label{app:model_analysis}

This appendix mirrors the benchmark-side extensions for the model-side analysis in \Cref{sec:model_predict}: the full $9 \times 2$ hypothesis $\times$ metric grid, and a per-model predictability ranking that names which models are easiest and hardest for \benchpress{} to predict.

\begin{figure}[p]
\centering
\includegraphics[height=0.85\textheight,width=\textwidth,keepaspectratio]{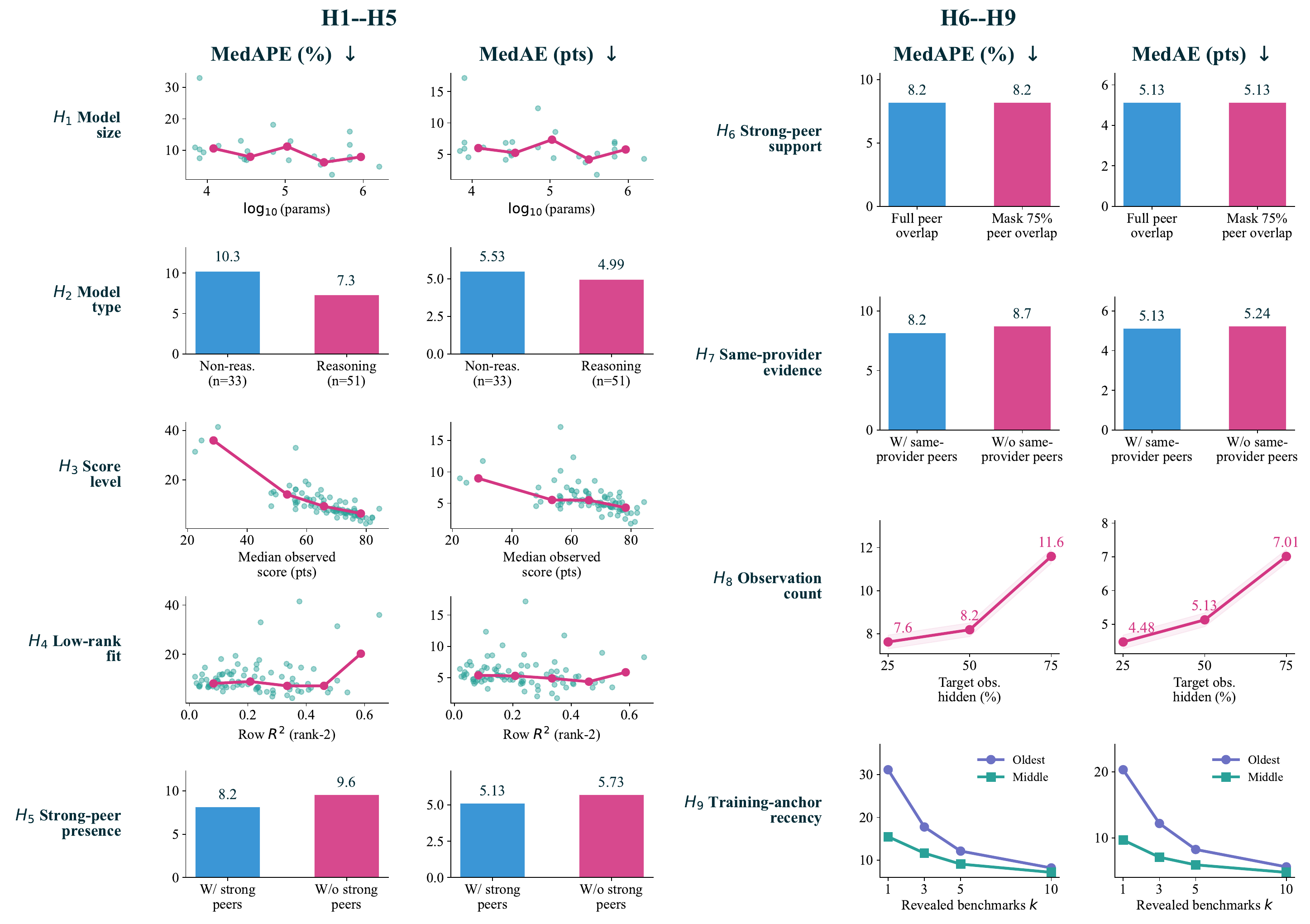}
\caption{\textbf{Selected settings for all nine model-level hypotheses against both score-error metrics.} The left block shows H1--H5 and the right block shows H6--H9. Columns within each block are $\mathsf{MedAPE}$ ($\downarrow$) and $\mathsf{MedAE}$ ($\downarrow$). H1--H4 are correlational rows (H2 grouped bars), H5--H8 are ablations, and H9 is temporal. Ablation rows show paired bars at the headline intervention (H5: $|r|\!\geq\!0.95$ peers; H6: 75\% strongest-peer overlap mask; H7: same-provider evidence) or a line across hide fractions (H8). H9 compares oldest vs.\ middle training anchors for the displayed revealed-benchmark counts $k\in\{1,3,5,10\}$; secondary H9 settings with $k=8$ and $k=15$ are not plotted in this figure.}
\label{fig:model_predictability_factors_full}
\end{figure}

\paragraph{Full hypothesis $\times$ metric grid.}
\label{app:model_predictability_factors_full}
\Cref{fig:error_hypotheses_52} in the main text visualizes five representative model-level hypotheses (H2, H3, H5, H8, H9) under a single metric per panel. For completeness, \Cref{fig:model_predictability_factors_full} expands this to the paper-facing setting for each of the nine hypotheses (H1--H9) against both score-error metrics, using the same correlational, ablation, and temporal setups as \Cref{tab:model_hypotheses}.

\paragraph{Per-model predictability.}
\label{app:per_model_predictability}
Mirroring the per-benchmark probe in \Cref{app:per_benchmark_predictability}, we apply the same half-per-model holdout but aggregate the test-cell errors by \emph{model row} instead of benchmark column. For each model we randomly hide half of its observed scores and predict them via the \benchpress{} predictor from \Cref{sec:method_comparison}, repeating over 10 random seeds for stability.

\begin{figure}[p]
\centering
\includegraphics[height=0.85\textheight,width=\textwidth,keepaspectratio]{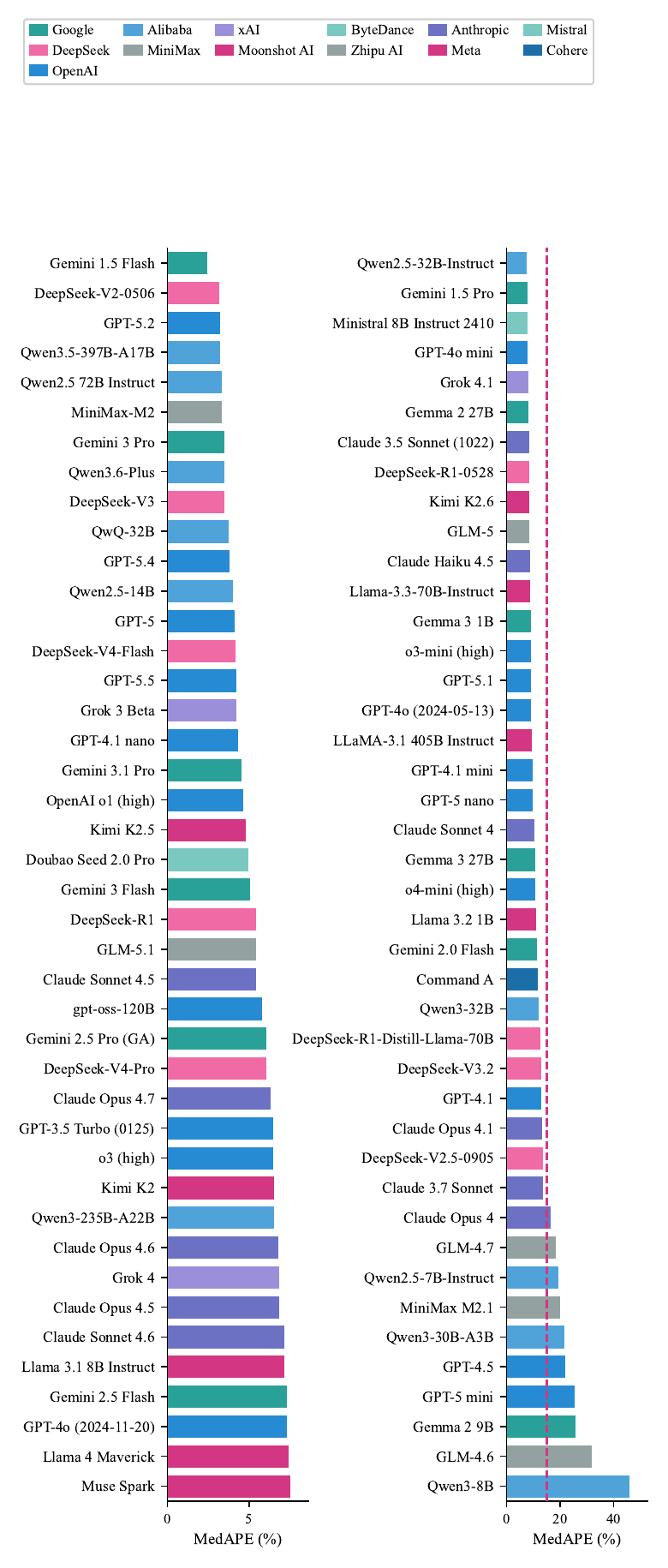}
\caption{\textbf{Per-model predictability.} For each model, half of observed scores are held out and predicted via \benchpress{}; errors are aggregated by model row (10 seeds). Models below the 15\% $\mathsf{MedAPE}$ threshold (dashed red line) are well-predicted by others. Color = provider.}
\label{fig:model_predictability}
\end{figure}

\Cref{fig:model_predictability} shows the per-model $\mathsf{MedAPE}$ for the 84 evaluated models. Roughly 88\% (74/84) fall below the 15\% $\mathsf{MedAPE}$ threshold and the median per-model $\mathsf{MedAPE}$ is $7.6\%$, indicating that most models are inferable from the rest of the matrix at limited additional information loss.

\begin{figure}[!htbp]
\centering
\includegraphics[width=\textwidth]{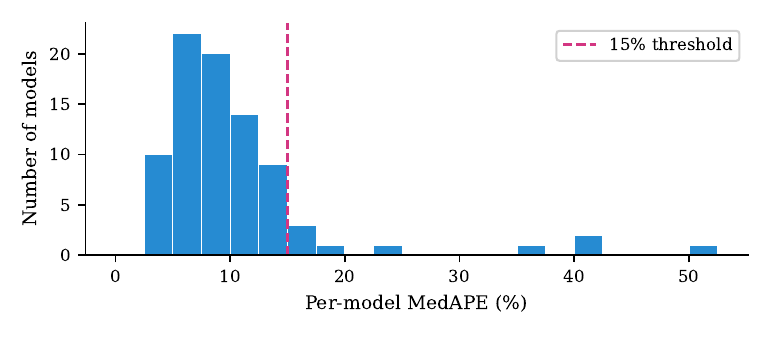}
\caption{\textbf{Distribution of per-model predictability.} The per-model $\mathsf{MedAPE}$ values of \Cref{fig:model_predictability}, binned across the evaluated models. Dashed line: the 15\% threshold.}
\label{fig:model_predictability_hist}
\end{figure}

\Cref{fig:model_predictability_hist} shows the same values as a distribution rather than a ranking: most models sit in a narrow band, and the models past the threshold form a thin right tail rather than a second mode.

\subsection{Estimating Prediction Reliability}
\label{app:confidence_calibration_details}

\Cref{sec:confidence_calibration} adds a reliability estimator to the default \benchpress{} score predictor from \Cref{sec:method_comparison}.
This appendix spells out the three reliability estimators used there.
All three models solve the same task: for a hidden model--benchmark cell, predict how large the absolute error of the fixed point prediction is likely to be.
During training, the target is the held-out absolute error after a $\log(1+x)$ transform.
During evaluation, the reliability estimator may use the training matrix, the fixed Logit Bias ALS prediction, and auxiliary predictions computed from the training fold, but it never sees the hidden score itself.

\paragraph{Ensemble-spread reliability estimator.}
The ensemble-spread model asks whether plausible score predictors agree on the same hidden cell.
It builds two stacks of alternative point predictions.
The first stack measures local sensitivity of the selected score predictor: the three Logit Bias ALS configurations in the \Cref{sec:method_comparison} grid with rank 2 and $\lambda\in\{0.01,0.1,1.0\}$.
The $\lambda=0.1$ member is the selected \benchpress{} score predictor, and the other two show how much the prediction moves under the adjacent regularization strengths in the grid.
The second stack measures disagreement with other strong full-coverage predictors.
We sort transform--method configurations by median percentage error in \Cref{tab:full_grid}, require coverage at least 99.9\%, remove the selected Logit Bias ALS predictor, and keep the first 12 remaining configurations.
In the checked-in run, these are Probit Bias ALS, Quantile Bias ALS, Identity Bias ALS, Quantile Soft-Impute, Logit Soft-Impute, Probit Soft-Impute, Arcsinh Bias ALS, Square root Bias ALS, Identity Soft-Impute, Logit Model-KNN, Probit Model-KNN, and Identity Model-KNN.
For each prediction stack, we record four spread summaries: standard deviation, median absolute deviation, central 80\% span, and the distance between the selected Logit Bias ALS prediction and the stack median.
These eight nonnegative features are transformed with $\log(1+x)$ before split-local standardization.

\paragraph{Matrix-support reliability estimator.}
The matrix-support model ignores alternative predictors and uses only evidence available in the observed score matrix.
For the target model, it records the number of observed benchmark scores and the median observed score.
For the target benchmark, it records the number of observed model scores, the median observed score, and the standard deviation of observed scores.
It also records the strongest peer model for the target model and the strongest neighboring benchmark for the target benchmark, where ``strongest'' means highest absolute Pearson correlation over shared observed scores in the training matrix.
The peer-model features are its absolute correlation with the target model and the number of shared observed benchmarks.
The benchmark-neighbor feature is its absolute correlation with the target benchmark.
We do not include benchmark-neighbor overlap because the stricter H7 ablation in \Cref{sec:bench_predict} does not support it as a joint benchmark-side reliability factor.

\paragraph{Hybrid reliability estimator and calibration.}
The hybrid reliability estimator concatenates the ensemble-spread features and the matrix-support features, then uses the same risk-model selection procedure as the two single-signal models.
Before fitting any of the three learned reliability estimators, each feature column is clipped at zero, transformed with $\log(1+x)$, and standardized using only the training split.
For every evaluated fold, candidate risk models are trained on the other folds only, and the architecture is selected inside those training folds from a linear ridge model with zero hidden layers, one ReLU MLP layer of 16 units, one ReLU MLP layer of 32 units, or two ReLU MLP layers of 64 and 32 units.
Concretely, after holding out the evaluated fold, the remaining folds are split by fold index for architecture selection: cells with fold index divisible by 5 form the inner validation set and the rest form the inner training set, giving roughly a 4:1 split.
If this modulo-5 split leaves too few validation or training cells, we fall back to fold index modulo 3, giving roughly a 2:1 split.
The selected folds chose only MLP configurations: the hybrid estimator selected 16, 32, and 64/32 hidden units in 7, 15, and 8 folds; the ensemble-spread estimator selected them in 12, 5, and 13 folds; the matrix-support estimator selected 32 and 64/32 hidden units in 3 and 27 folds.
After the architecture is selected, the MLP variants use ReLU activations, Adam, $\ell_2$ penalty $10^{-3}$, learning rate $3{\times}10^{-3}$, a separate 15\% early-stopping validation fraction within the fitting routine, 25 no-improvement iterations, at most 500 iterations, and deterministic seeds derived from base seed 42.
Each fitted model outputs a risk score, where larger values mean less reliable point predictions.
For display, we calibrate this ordering into a trust probability: the probability that predictions with similar risk fall within a chosen number of score points of the reported score.
The display calibration bins held-out cells by hybrid risk, estimates the empirical within-tolerance probability in each bin, enforces a monotone nonincreasing map from risk to trust probability, and interpolates this map for displayed cells.
For prediction intervals, we apply the same leave-fold-out conformal wrapper to each reliability estimator: on all folds except the evaluated fold, take the 90th percentile of $|\hat{s}-s|/r$, multiply the evaluated fold's risk score $r$ by that scale, and center the resulting 90\% interval at the fixed Logit Bias ALS point prediction $\hat{s}$.
\Cref{tab:confidence_calibration_widths} reports the resulting interval widths at three coverage levels.

\begin{table}[!htbp]
\centering
\caption{\textbf{Conformal interval widths} (score points) at three coverage levels; lower is sharper. The hybrid row is shaded.}
\label{tab:confidence_calibration_widths}
\small
\setlength{\tabcolsep}{5pt}
\renewcommand{\arraystretch}{1.2}
\begin{tabular}{@{}lccc@{}}
\toprule
Method & 80\% & 90\% & 95\% \\
\midrule
Ensemble spread & 19.40 & 27.43 & 36.36 \\
Matrix support  & 20.39 & 29.18 & 38.75 \\
\cBP\textbf{Hybrid} & \cBP\textbf{19.56} & \cBP\textbf{27.01} & \cBP\textbf{35.43} \\
\bottomrule
\end{tabular}
\end{table}

\paragraph{Interval width versus number of observed scores.}
The widths above condition on each model's full training split.
To isolate how an interval narrows as a single benchmark accumulates observations, we run a controlled variant that varies only the number of revealed scores.
We keep the $23$ benchmarks with at least $30$ observed model scores, hold out the same $10$ scores per benchmark for evaluation, and reveal $k\in\{5,10,15,20\}$ of the remaining scores in that column as conditioning evidence while hiding the rest of the column and keeping all other columns observed.
We predict the $10$ held-out cells with the fixed Logit Bias ALS predictor and form leave-one-benchmark-out conformal $80\%$ intervals at each $k$.
\Cref{tab:interval_width_by_observations} shows that empirical coverage stays near the $80\%$ target while the median interval width falls from $32.3$ to $23.0$ score points and the median absolute error drops from $5.8$ to $4.7$ as more scores are revealed.

\begin{table}[!htbp]
\centering
\caption{\textbf{Prediction intervals sharpen as more scores are observed.} Restricting to benchmarks with at least $30$ observed model scores, holding out the same $10$ evaluation cells, and revealing $k$ conditioning scores per benchmark. Coverage is the empirical fraction of held-out cells inside the $80\%$ interval; width is the median interval width in score points.}
\label{tab:interval_width_by_observations}
\small
\setlength{\tabcolsep}{6pt}
\renewcommand{\arraystretch}{1.2}
\begin{tabular}{@{}cccc@{}}
\toprule
Revealed scores $k$ & MedAE & $80\%$ coverage & Median width \\
\midrule
5  & 5.83 & 0.799 & 32.28 \\
10 & 4.96 & 0.798 & 25.32 \\
15 & 4.93 & 0.800 & 23.32 \\
20 & 4.69 & 0.798 & 23.04 \\
\bottomrule
\end{tabular}
\end{table}

\end{document}